\pdfoutput=1
\documentclass[nohyperref]{article}

\usepackage{microtype}
\usepackage{graphicx}
\usepackage{subfigure}

\usepackage{booktabs} %
\usepackage{natbib}

\usepackage[colorlinks,citecolor=DarkGreen,linkcolor=FireBrick,urlcolor=FireBrick]{hyperref}

\usepackage{graphicx}
\usepackage{tikz}
\usepackage{tikz-cd}
\usepackage{times}
\usepackage{courier}
\usepackage{bm}
\usepackage{physics}
\usepackage{xcolor}
\usepackage{natbib}
\usepackage{mdframed}
\usepackage{nicefrac}
\usepackage{booktabs}
\usepackage{blindtext}
\usepackage{braket}
\usepackage{amsmath}
\usepackage{amssymb}
\usepackage{mathtools}
\usepackage{titletoc}
\usepackage{enumitem}

\usepackage{graphicx}
\usepackage{tikz}
\usepackage{tikz-cd}
\usepackage{times}
\usepackage{courier}
\usepackage{bm}
\usepackage{physics}
\usepackage{xcolor}
\usepackage{natbib}
\usepackage{mdframed}
\usepackage{nicefrac}
\usepackage{booktabs}
\usepackage{lipsum}
\usepackage{titlesec}
\usepackage{wrapfig,lipsum,booktabs}
\usepackage{authblk}
\usepackage{blindtext}
\usepackage{titletoc}
\usepackage{multirow} 
\usepackage{algorithm}
\usepackage{algorithmic}
\usepackage{titletoc}
\usepackage{setspace}

\usepackage{authblk}
\definecolor{DarkGreen}{rgb}{0,0.40,0}
\definecolor{FireBrick}{rgb}{0.698,0.133,0.133}
\usepackage{colortbl}

\usepackage{float}
\usepackage{xcolor}
\definecolor{LightCyan}{rgb}{0.8, 0.9, 1}

\newcolumntype{g}{>{\columncolor{LightCyan}\hspace{0pt}}c}

\newcommand*\rot{\rotatebox{90}}

\usepackage[nameinlink,capitalize,noabbrev]{cleveref}

\usepackage{CJK}

\newenvironment{claim}{  \begin{mdframed}[linecolor=black!0,backgroundcolor=black!10]\noindent%
		\ignorespaces}{\end{mdframed}}

\usepackage{array}
\usepackage{makecell}

\newcommand{\jhcomment}[1]{\textcolor{firebrick}{[JH: #1]}}

\def\half{{\frac{1}{2}}}

\newcommand{\bea}{\begin{eqnarray}}
\newcommand{\eea}{\end{eqnarray}}

\usepackage{slashed}

\def\({\left(}
\def\){\right)}
\def\[{\left[}
\def\]{\right]}

\usepackage{dashbox}
\usepackage{xcolor}
\usepackage{colortbl}
\usepackage{arydshln}
\definecolor{lightyellow}{rgb}{1.0, 0.95, 0.7}
\definecolor{blue}{rgb}{0.0, 0.4, 1.0}
\definecolor{Blue}{rgb}{0,0,1}
\definecolor{darkgreen}{rgb}{0,0.40,0}
\definecolor{firebrick}{rgb}{0.698,0.133,0.133}

\definecolor{colorA}{rgb}{1,0,0}
\definecolor{colorB}{rgb}{0,0.3,1}
\definecolor{colorC}{rgb}{0.9,0.8,0.2}
\definecolor{colorD}{rgb}{0,0.65,0}
\definecolor{lesslightgray}{rgb}{0.5,0.5,0.5}
\definecolor{light-gray}{gray}{0.95}

\let\tilde\widetilde
\let\hat\widehat

\newcommand{\calD}{\mathcal{D}}

\newcommand{\calG}{\mathcal{G}}

\newcommand{\calK}{\mathcal{K}}
\newcommand{\calL}{\mathcal{L}}
\newcommand{\calM}{\mathcal{M}}

\newcommand{\calO}{\mathcal{O}}

\newcommand{\calS}{\mathcal{S}}
\newcommand{\calT}{\mathcal{T}}

\newcommand{\calX}{\mathcal{X}}
\newcommand{\calY}{\mathcal{Y}}

\newcommand{\bA}{\mathbf{A}}

\newcommand{\bD}{\mathbf{D}}

\newcommand{\bK}{\mathbf{K}}

\newcommand{\bP}{\mathbf{P}}
\newcommand{\bQ}{\mathbf{Q}}
\newcommand{\bR}{\mathbf{R}}

\newcommand{\bV}{\mathbf{V}}
\newcommand{\bW}{\mathbf{W}}
\newcommand{\bX}{\mathbf{X}}
\newcommand{\bY}{\mathbf{Y}}
\newcommand{\bZ}{\mathbf{Z}}
\newcommand{\ba}{\mathbf{a}}
\newcommand{\bb}{\mathbf{b}}
\newcommand{\bc}{\mathbf{c}}

\newcommand{\be}{\mathbf{e}}

\newcommand{\bu}{\mathbf{u}}
\newcommand{\bp}{\mathbf{p}}

\newcommand{\br}{\mathbf{r}}
\newcommand{\bv}{\mathbf{v}}

\newcommand{\bx}{\mathbf{x}}
\newcommand{\by}{\mathbf{y}}
\newcommand{\bz}{\mathbf{z}}
\newcommand{\bxi}{{\bm{\xi}}}

\newcommand{\Max}{\mathop{\rm max}}
\newcommand{\Min}{\mathop{\rm min}}
  
\newcommand{\argmax}{\mathop{\mathrm{argmax}}}

\newcommand{\Softmax}{\mathop{\rm{Softmax}}}

\newcommand{\lse}{\mathop{\rm{lse}}}

\newcommand{\sT}{ \mathsf{T} }

\newcommand{\sumM}{\sum_{\mu=1}^M}

\def\R{\mathbb{R}}

\def\E{\mathbb{E}}

\let\cite\citep 

\usepackage{amsthm}
\makeatletter
\def\th@remark{%
  \thm@headfont{\bfseries}%
  \normalfont %
  \thm@preskip\topsep \divide\thm@preskip\tw@
  \thm@postskip\thm@preskip
}
\makeatother
\usepackage[many]{tcolorbox}
\theoremstyle{definition}
\newtheorem{theorem}{Theorem}[section]
\tcolorboxenvironment{theorem}{
  breakable,
  colback=black!10,
  colframe=white,%
  width=\dimexpr\linewidth+10pt\relax,%
  enlarge left by=-5pt,%
  enlarge right by=-5pt,%
  boxsep=5pt,%
  boxrule=0pt,
  left=0pt,right=0pt,top=0pt,bottom=0pt,
  sharp corners,
  before skip=\topsep,
  after skip=\topsep
}
\newtheorem{lemma}{Lemma}[section]
\tcolorboxenvironment{lemma}{
  breakable,
  colback=black!10,
  colframe=white,%
  width=\dimexpr\linewidth+10pt\relax,%
  enlarge left by=-5pt,%
  enlarge right by=-5pt,%
  boxsep=5pt,%
  boxrule=0pt,
  left=0pt,right=0pt,top=0pt,bottom=0pt,
  sharp corners,
  before skip=\topsep,
  after skip=\topsep
}
\newtheorem{corollary}{Corollary}[theorem]
\tcolorboxenvironment{corollary}{
  breakable,
  colback=black!10,
  colframe=white,%
  width=\dimexpr\linewidth+10pt\relax,%
  enlarge left by=-5pt,%
  enlarge right by=-5pt,%
  boxsep=5pt,%
  boxrule=0pt,
  left=0pt,right=0pt,top=0pt,bottom=0pt,
  sharp corners,
  before skip=\topsep,
  after skip=\topsep
}

\theoremstyle{definition}
\newtheorem{definition}{Definition}[section]
\tcolorboxenvironment{definition}{
  breakable,
  colback=black!10,
  colframe=white,%
  width=\dimexpr\linewidth+10pt\relax,%
  enlarge left by=-5pt,%
  enlarge right by=-5pt,%
  boxsep=5pt,%
  boxrule=0pt,
  left=0pt,right=0pt,top=0pt,bottom=0pt,
  sharp corners,
  before skip=\topsep,
  after skip=\topsep
}
\theoremstyle{remark}
\newtheorem{remark}{Remark}[section]
\newtheorem{assumption}{Assumption}[section]
\tcolorboxenvironment{assumption}{
  breakable,
  colback=black!10,
  colframe=white,%
  width=\dimexpr\linewidth+10pt\relax,%
  enlarge left by=-5pt,%
  enlarge right by=-5pt,%
  boxsep=5pt,%
  boxrule=0pt,
  left=0pt,right=0pt,top=0pt,bottom=0pt,
  sharp corners,
  before skip=\topsep,
  after skip=\topsep
}

\tcolorboxenvironment{problem}{
  breakable,
  colback=black!10,
  colframe=white,%
  width=\dimexpr\linewidth+10pt\relax,%
  enlarge left by=-5pt,%
  enlarge right by=-5pt,%
  boxsep=5pt,%
  boxrule=0pt,
  left=0pt,right=0pt,top=0pt,bottom=0pt,
  sharp corners,
  before skip=\topsep,
  after skip=\topsep
}

\crefname{theorem}{Theorem}{Theorems}
\crefname{proposition}{Proposition}{Propositions}
\crefname{lemma}{Lemma}{Lemmas}
\crefname{corollary}{Corollary}{Corollaries}
\crefname{definition}{Definition}{Definitions}
\crefname{assumption}{Assumption}{Assumptions}
\crefname{remark}{Remark}{Remarks}
\crefname{problem}{Problem}{Problems}
\crefname{property}{Property}{property}

\numberwithin{equation}{section}
\numberwithin{theorem}{section}
\numberwithin{proposition}{section}
\numberwithin{definition}{section}
\numberwithin{lemma}{section}
\numberwithin{assumption}{section}
\numberwithin{remark}{section}

\usepackage{lipsum}

\newcommand*{\annot}[1]{\tag*{\footnotesize{\textcolor{black!50}{\big(#1\big)}}}}

\makeatletter
\let\save@mathaccent\mathaccent
\newcommand*\if@single[3]{%
    \setbox0\hbox{${\mathaccent"0362{#1}}^H$}%
    \setbox2\hbox{${\mathaccent"0362{\kern0pt#1}}^H$}%
    \ifdim\ht0=\ht2 #3\else #2\fi
}
\newcommand*\rel@kern[1]{\kern#1\dimexpr\macc@kerna}
\newcommand*\widebar[1]{\@ifnextchar^{{\wide@bar{#1}{0}}}{\wide@bar{#1}{1}}}
\newcommand*\wide@bar[2]{\if@single{#1}{\wide@bar@{#1}{#2}{1}}{\wide@bar@{#1}{#2}{2}}}
\newcommand*\wide@bar@[3]{%
    \begingroup
    \def\mathaccent##1##2{%
        \let\mathaccent\save@mathaccent
        \if#32 \let\macc@nucleus\first@char \fi
        \setbox\z@\hbox{$\macc@style{\macc@nucleus}_{}$}%
        \setbox\tw@\hbox{$\macc@style{\macc@nucleus}{}_{}$}%
        \dimen@\wd\tw@
        \advance\dimen@-\wd\z@
        \divide\dimen@ 3
        \@tempdima\wd\tw@
        \advance\@tempdima-\scriptspace
        \divide\@tempdima 10
        \advance\dimen@-\@tempdima
        \ifdim\dimen@>\z@ \dimen@0pt\fi
        \rel@kern{0.6}\kern-\dimen@
        \if#31
        \overline{\rel@kern{-0.6}\kern\dimen@\macc@nucleus\rel@kern{0.4}\kern\dimen@}%
        \advance\dimen@0.4\dimexpr\macc@kerna
        \let\final@kern#2%
        \ifdim\dimen@<\z@ \let\final@kern1\fi
        \if\final@kern1 \kern-\dimen@\fi
        \else
        \overline{\rel@kern{-0.6}\kern\dimen@#1}%
        \fi
    }%
    \macc@depth\@ne
    \let\math@bgroup\@empty \let\math@egroup\macc@set@skewchar
    \mathsurround\z@ \frozen@everymath{\mathgroup\macc@group\relax}%
    \macc@set@skewchar\relax
    \let\mathaccentV\macc@nested@a
    \if#31
    \macc@nested@a\relax111{#1}%
    \else
    \def\gobble@till@marker##1\endmarker{}%
    \futurelet\first@char\gobble@till@marker#1\endmarker
    \ifcat\noexpand\first@char A\else
    \def\first@char{}%
    \fi
    \macc@nested@a\relax111{\first@char}%
    \fi
    \endgroup
    }
\makeatother
\let\bar\widebar

\makeatletter
\newcommand*{\redefinesymbolwitharg}[1]{%
  \expandafter\let\csname ltx#1\expandafter\endcsname\csname #1\endcsname
  \@namedef{#1}{\@ifnextchar{^}{\@nameuse{#1@}}{\@nameuse{#1@}^{}}}%
  \expandafter\def\csname #1@\endcsname^##1##2{%
     \csname ltx#1\endcsname\ifx!##1!\else^{##1}\fi\mathopen{}\mathclose\bgroup\left(##2\aftergroup\egroup\right)
     }%
}
\makeatother
\redefinesymbolwitharg{sin}
\redefinesymbolwitharg{cos}

\usepackage{url}
\newcommand{\uhop}{{$\mathtt{U\text{-}Hop}$}}

\usepackage{amsmath}
\usepackage{amssymb}
\usepackage{mathtools}
\usepackage{amsthm}
\usepackage{amsmath}
\usepackage{dsfont}

\usepackage[accepted]{icml2024}

\usepackage[textsize=tiny]{todonotes}

\icmltitlerunning{Uniform Memory Retrieval with Larger Capacity for Modern Hopfield Models
}
\usepackage{graphicx} %

\titlespacing\section{0pt}{1pt plus 1pt minus 1pt}{0pt plus 0pt minus 0pt}
\titlespacing\subsection{0pt}{0pt plus 0pt minus 0pt}{0pt plus 0pt minus 0pt}
\titlespacing\subsubsection{0pt}{0pt plus 0pt minus 0pt}{0pt plus 0pt minus 0pt}
\setlist[itemize]{leftmargin=1em, before=\vspace{-0.5em}, after=\vspace{-0.5em}, itemsep=0.1em}
\setlist[enumerate]{leftmargin=1.2em, before=\vspace{-0.5em}, after=\vspace{-0.5em}, itemsep=0.1em}

\usepackage{mathtools}
\newcounter{mysubequations}

\begin{document}

\twocolumn[
\icmltitle{}

\icmlsetsymbol{equal}{*}

\begin{icmlauthorlist}
\icmlauthor{Dennis Wu}{equal,yyy}
\icmlauthor{Jerry Yao-Chieh Hu}{equal,yyy}
\icmlauthor{Teng-Yun Hsiao}{comp}
\icmlauthor{Han Liu}{yyy,stats}
\end{icmlauthorlist}

\icmlaffiliation{yyy}{Department of Computer Science,
University of Northwestern, Evanston, USA}
\icmlaffiliation{stats}{Department of Statistics and Data Science,
University of Northwestern, Evanston, USA}
\icmlaffiliation{comp}{Department of Physics, National Taiwan University, Taipei, Taiwan}

\icmlcorrespondingauthor{Dennis Wu}{\href{mailto:hibb@northwestern.edu}{hibb@northwestern.edu}}
\icmlcorrespondingauthor{Jerry Yao-Chieh Hu}{\href{mailto:jhu@northwestern.edu}{jhu@northwestern.edu}}
\icmlcorrespondingauthor{Teng-Yun Hsiao}{\href{mailto:b10502058@ntu.edu.tw}{b10502058@ntu.edu.tw}}
\icmlcorrespondingauthor{Han Liu}{\href{mailto:hanliu@northwestern.edu}{hanliu@northwestern.edu}}

\icmlkeywords{Machine Learning, ICML}

\vskip 0.3in
]

\printAffiliationsAndNotice{\icmlEqualContribution} %

\titlespacing*{\section}{0pt}{0pt}{0pt}
\titlespacing*{\subsection}{0pt}{0pt}{0pt}
\titlespacing*{\subsubsection}{0pt}{0pt}{0pt}
\begin{abstract}
We propose a two-stage memory retrieval dynamics for modern Hopfield models, termed $\mathtt{U\text{-}Hop}$, with enhanced memory capacity.
Our key contribution is a learnable feature map $\Phi$
which transforms the Hopfield energy function into kernel space. 
This transformation ensures convergence between the local minima of energy and the fixed points of retrieval dynamics within the kernel space.
Consequently, 
the kernel norm induced by $\Phi$
serves as a novel similarity measure. 
It utilizes the stored memory patterns as learning data to enhance memory capacity across all modern Hopfield models.
Specifically, we accomplish this by constructing a separation loss $\calL_\Phi$ that separates the local minima of kernelized energy by separating stored memory patterns in kernel space.
Methodologically,
{\uhop} memory retrieval process consists of:
\textbf{(Stage~I)} minimizing separation loss for a more uniformed memory (local minimum) distribution, followed by \textbf{(Stage~II)} standard Hopfield energy minimization for memory retrieval.
This results in a significant reduction of possible metastable states in the Hopfield energy function, thus enhancing memory capacity by preventing memory confusion.
Empirically, with real-world datasets, we demonstrate that $\mathtt{U\text{-}Hop}$  outperforms all existing modern Hopfield models and SOTA similarity measures, achieving substantial improvements in both associative memory retrieval and deep learning tasks.
Code is available at \href{https://github.com/MAGICS-LAB/UHop}{GitHub};
future updates are on \href{https://arxiv.org/abs/2404.03827}{arXiv}.\footnote{Note Added [November 2024]: The follow-up work \cite{hwl24} presents a provably optimal memory capacity bound for kernelized modern Hopfield models and introduces a sub-linear time algorithm, $\mathtt{U\text{-}Hop}$+, to achieve this optimal capacity.}

\end{abstract}

\section{Introduction}
\label{sec:intro}
\begin{figure*}[t]
    \centering
    \includegraphics[width=1\textwidth]{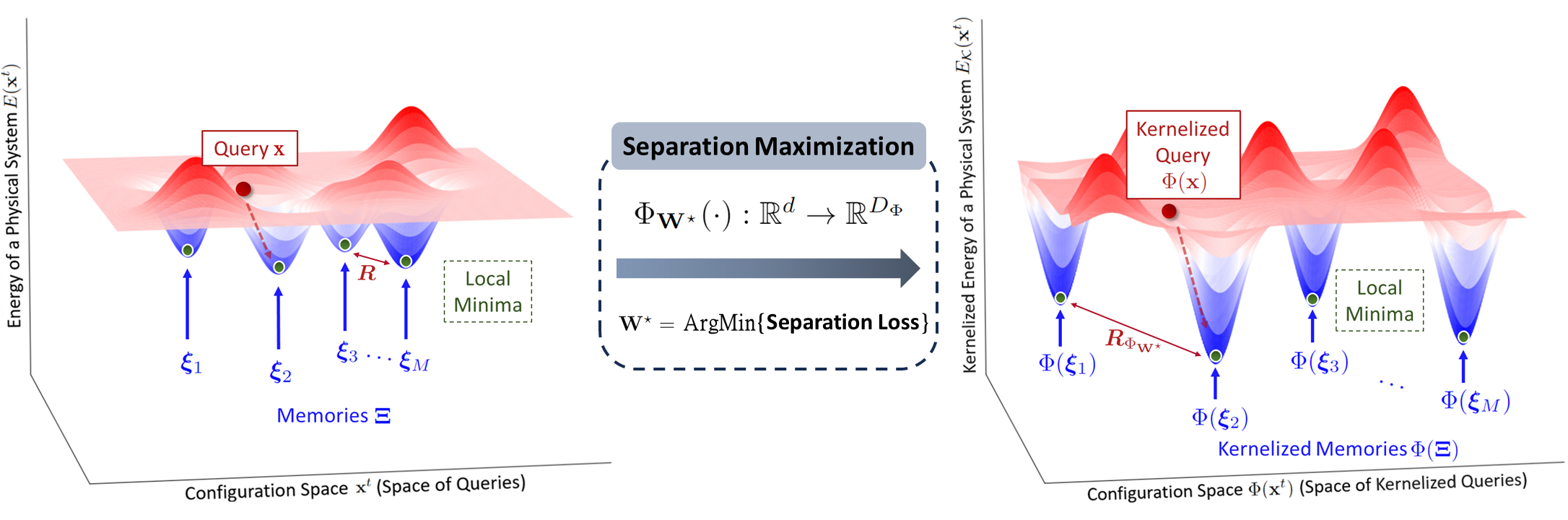}
    \vspace{-1em}
    \caption{\textbf{Visualization of {\uhop}:  Separation Maximization First, then Memory Retrieval Dynamics.}
    The LHS represents the energy landscape in original state space, where the memories stay close to each other.
    With separation loss minimization, we obtain a $\Phi$ parameterized by $\bW^\star$, that is able to relocate memory patterns in the kernel space to more uniform locations, and thus results in the separation between local minima of $E_\calK$.
    }
    \label{fig:visualization}
\end{figure*}

We address the memory confusion problem in the modern Hopfield models by proposing a two-stage optimization formulation,
termed {\uhop}, for the memory retrieval dynamics
of modern Hopfield models.
We construct the similarity measure of modern Hopfield models with a learnable kernel.
The feature map of the kernel is trained by maximizing the separation among the entire stored memory set (\cref{fig:visualization}).
This allows Hopfield models under {\uhop} 
 to distinguish different memory patterns with larger separation and hence achieve larger memory capacity.

\begin{figure}[t]
\centering
    \includegraphics[width=.35\textwidth]{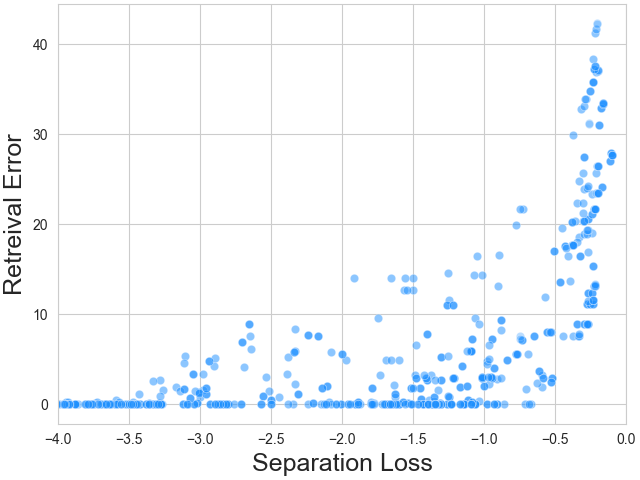}
    \vspace{-1em}
    \caption{\textbf{Separation Loss over Memory Set v.s. Retrieval Error.}
    We perform 200 runs of memory retrieval with {\uhop} on MNIST.
    The result shows a strong correlation between low separation loss and low retrieval error.
    }
    \label{fig:mem-contribution}
\end{figure}

Let $\bx\in\R^d$ be the input query pattern, $\bm{\Xi} \coloneqq  [ \bxi_1,\cdots,\bxi_M ] \in \R^{d\times M}$ be the memory patterns, and 
$\Braket{\ba, \bb} \coloneqq \ba^\sT \bb$ be the inner product for vectors $\ba, \bb \in \R^d$.
Hopfield models are energy-based associative memory models.
They store memory patterns on the local minima of their energy landscapes.
For any input query $\bx$, they retrieve its closest memory pattern through some energy minimization algorithms initialized at $\bx$.
These algorithms are also known as memory retrieval dynamics.
See \cref{fig:visualization} for a visualization.
\citet{ramsauer2020hopfield} proposed a large foundation model  compatible variant, the  
Modern Hopfield Model (MHM).
This model has a specific set of energy function and retrieval dynamics, such that it subsumes transformer attention as its special case (see \cref{sec:connect-attention}) and enjoys superior theoretical properties (see \cite{hu2023SparseHopfield,wu2023stanhop,ramsauer2020hopfield}).
Specifically, they introduce the energy function:
\begin{align}
\label{eqn:MHM}
    E_{\text{MHM}}(\bx) = -\lse(\beta,\bm{\Xi}^\sT \bx) + \frac{1}{2} \langle \bx,\bx \rangle ,
\end{align}
and the retrieval dynamics  
\begin{align}
    \label{eqn:retrival_dyn}
\bx^{\text{new}}=\calT_{\text{MHM}}(\bx) = \bm{\Xi} \cdot \Softmax(\beta \bm{\Xi}^\sT \bx),
\end{align}
where $\lse\(\beta,\bz\)\coloneqq \log\(\sum_{\mu=1}^M e^{\beta z_\mu} \)/\beta$ is the log-sum-exponential function for any given vector $\bz\in\R^M$ and $\beta>0$.
The dot-product $\bm{\Xi}^\sT \bx$ in the lse function is known as the \textit{overlap construction} and serves as the similarity measure between the input query $\bx$ and memory set $\bm{\Xi}$.

One highlighted property of modern Hopfield models is their memory capacity, which is exponential in pattern dimension  \cite{ramsauer2020hopfield, hu2023SparseHopfield, wu2023stanhop}.
However, their memory capacity and retrieval error are dependent on the quality of memory distribution.
To be concrete, 
we define the memory storage and retrieval as\footnote{Recall that,
    Given a function $\calT:\R^d\to \R^d$.
    A generalized fixed point of $\calT$ is a point $\bx\in\R^d$ for which $\bx\in \calT(\bx)$.
}
\begin{definition} [Stored and Retrieved]
\label{def:stored_and_retrieved_org}
For all $\mu\in[M]$, let $R\coloneqq \half \Min_{\mu,\nu\in[M];\mu\neq\nu}\norm{\bxi_\mu-\bxi_\nu}$ be the finite radius 
 of each sphere $\calS_\mu$ centered at memory pattern $\bxi_\mu$.
We say $\bxi_\mu$ is \textit{stored} if all $\bx\in\calS_\mu$ are generalized fixed points of $\calT$, $\bx^\star_\mu \in \calS_\mu$, and $\calS_\mu \cap \calS_\nu=\emptyset$ for $\mu \neq \nu$.
We say $\bxi_\mu$ is $\epsilon$-\textit{retrieved} by $\calT$ with $\bx$ for an error $\epsilon$, if $\norm{\calT(\bx)-\bxi_\mu}\le \epsilon$.
\end{definition}
Let $\Delta_\mu \coloneqq \Braket{\bxi_\mu,\bxi_\mu} - \Max_{\nu\in[M],\nu\neq \mu} \Braket{\bxi_\nu,\bxi_\mu}  $ be the separation between a memory pattern $\bxi_\mu$ from all other memories in $\bm{\Xi}$, and $m$ be the largest norm among memory patterns.
    \citet{ramsauer2020hopfield} gives the retrieval error bound:
    \begin{align}
    \label{eqn:error_bound_exp}
    \norm{\calT_{\text{MHM}}(\bx)-\bxi_\mu} 
    \le  2m(M-1) e^{-\beta(\Delta_\mu -2mR)},
    \end{align}
    for any $\bx\in\calS_\mu$.
    This bound is crucial not only for characterizing retrieval quality but also, in capacity analysis, as a necessary condition for pattern $\bxi_\mu$ to be stored in the model \citep[Theorem~3.1]{hu2023SparseHopfield}.
    Yet, it depends on $\bm{\Xi}$.

$\Delta_\mu$ measures the distance from a given $\bxi_\mu$ to the nearest memory pattern in $\bm{\Xi}$.
$R$ measures the minimal separation among all stored memories $\bm{\Xi}$.
Hence, they are both $\bm{\Xi}$-dependent.
This $\bm{\Xi}$-dependence in \eqref{eqn:error_bound_exp} results in potential \textit{fuzzy retrievals}, namely metastable states caused by multiple nearby local minima in the energy landscape, especially when $\Delta_\mu-2mR$ is small.
When this occurs, 
these fuzzy retrievals deviate the retrieval process from the ground truth, thereby hampering  performance.

This fuzzy memory (memory confusion) issue is well-known in literature.
The dense associative memory model \cite{krotov2016dense} tries to solve this issue by using polynomial energy fucntion.
The modern Hopfield models \cite{demircigil2017model,ramsauer2020hopfield,hu2023SparseHopfield,wu2023stanhop,hu2024nonparametric} try to solve this issue by using exponential energy functions.
However, 
all these attempts still rely on the quality of $\bm{\Xi}$.
In this work, we rethink the use of inner-product similarity measure (i.e. $\bm{\Xi}^\sT \bx$ in \eqref{eqn:retrival_dyn}),
and consider it as primary source of the fuzzy memory problem. 
Specifically,
due to its Euclidean nature,
inner-product assigns equal importance to all dimensions of patterns and yields small $\Delta_\mu-2mR$ if they ($\bxi_\mu$ and some $\bxi_\nu$) share similar direction.
This motivate us to replace the overlap (inner product) construction of the energy function with a similarity measure utilizing this $\bm{\Xi}$-dependence.

To this end, we propose a kernalized similarity measure  for all modern Hopfield models, named \uhop.
This measure is learnable.
We propose to learn it by minimizing the average separation among all possible  stored memory pairs in set $\bm{\Xi}$.
Namely, it is $\bm{\Xi}$-sensitive.
Physically,
it converts the original energy landscape into a \textit{kernalized} landscape with (on average) equally separated minima.
While it does not provably guarantee enlarging $R$ (the minimal separation among $\{\boldsymbol{\xi}_\mu\}_{\mu \in [M]}$) in the kernel space, $\mathtt{U\text{-}Hop}$ addresses the root cause of the fuzzy memory problem with strong empirical evidence.
It delivers a larger memory capacity and a tighter retrieval error bound for modern Hopfield models, surpassing all existing modern Hopfield models \cite{hu2023SparseHopfield,wu2023stanhop,ramsauer2020hopfield,krotov2016dense} and SOTA similarity measures, i.e.  $\ell_2$-distance and Manhattan distance proposed by \citet{millidge2022universal}.

\paragraph{Contributions.} Our contributions are as follows:
\begin{itemize}
    \item 
    We introduce a learnable feature map $\Phi$ that maps energy $E$ to a kernel space with kernel $\calK(\cdot,\cdot)\coloneqq\Braket{\Phi(\cdot),\Phi(\cdot)}$.
    The resulting kernelized energy $E_\calK$ , and its corresponding retrieval dynamics $\calT_\calK$ satisfy the defining properties of modern Hopfield models: convergence between local minima of $E$ and fixed points of retrieval dynamics $\calT$. 
    This allows us to construct a separation loss $\calL_\Phi$ that distinguishes the local minima of $E_\calK$ by separating stored memory patterns in kernel space.

    \item 
    Methodologically,
    we introduce Uniform Hopfield Memory Retrieval ({\uhop}).
    It is a two-stage optimization formulation. %
    The first stage is separation loss $\calL_\Phi$ minimization, distancing stored memory patterns in kernel space.
    The second stage performs energy minimization with the kernel-induced $\calT_\calK$.
    The first stage enhanced $E_\calK$, making it able to relocate its local minima to a more separated coordinate.
    As a result, modern Hopfield models under {\uhop} is able to obtain improved memory capacity.

    \item 
    Empirically, $\mathtt{U\text{-}Hop}$ improves memory retrieval outcomes by a large margin comparing to other baselines.
    When applied to deep learning scenarios, $\mathtt{U\text{-}Hop}$ significantly improves model's memorization capacity, generalization and convergence speed.
    We show that {\uhop} improves memory retrieval tasks by an average 30\% margin even under a single iteration of separation minimization, and learning tasks by an average 3\% margin.
    
\end{itemize}
\paragraph{Organization.}
\cref{sec:method} presents {\uhop}.
\cref{sec:DL} connects {\uhop} to deep learning.
\cref{sec:exp} conducts extensive numerical experiments to support {\uhop}. 
Appendix includes proofs, experimental details, and additional experimental studies.

\paragraph{Notations.}
Bold lower case letters denote vectors and bold upper case letters denote matrices.
We write $\Braket{\ba, \bb} \coloneqq \ba^\sT \bb$ as the inner product for vectors $\ba, \bb \in \R^d$.
The index set $\{ 1, ..., I \}$ is denoted by $
\[ I \]$, where $I \in \mathbb{N}^+$. 
The spectral norm is denoted by $\norm{\cdot}_2$ which is equivalent to the $\ell_2$-norm when applied to a vector.
Throughout this paper, we denote the memory patterns (keys) by $\bxi \in \R^d$ and the state/configuration/query pattern by $\bx \in \R^d$, and $\bm{\Xi} \coloneqq \( \bxi_1, ... ,\bxi_M \) \in \R^{d \times M}$ as shorthand for stored memory (key) patterns $\{ \bxi_\mu \}_{\mu \in \[ M \]}$.
We set norm $n := \norm{\bx}$ to be the norm of the query pattern, and $m \coloneqq \max_{\mu \in \[M\]} \norm{\bxi_\mu} $ be the largest norm of memory patterns. We also provide a nomenclature table (\cref{sec:tab_notation}) in the appendix.

\section{$\mathtt{U\text{-}Hop}$: Retrieval as Two-Stage Optimization}
\label{sec:method}
In this section, 
\cref{sec:kernel} introduces a learnable feature map that maps patterns and the energy function into a kernel space, and demonstrate the fixed-point convergence property of kernelized modern Hopfield models.
\cref{sec:UHop_alg} presents {\uhop} (\cref{algorithm1}), a two-stage algorithm for the kernel learning with optimal theoretical guarantees.
It maximizes pattern separation by minimizing a novel Separation Loss.

\subsection{Kernelized Memory Hopfield Energy}
\label{sec:kernel}

In this section,
we first parameterize the similarity measure(s) 
in modern Hopfield model(s)  with a learnable kernel (via feature map \eqref{eqn:linear_kernel}), and then show the induced  models (with energy \eqref{eqn:eng-fn}) satisfying the defining properties of modern Hopfield models (\cref{lemma:retrieval_dyn}, \cref{lemma:convergence}).

Let $\calK(\cdot,\cdot)\coloneqq\Braket{\Phi(\cdot),\Phi(\cdot)}:\R^{d}\times \R^{d}\to \R_+$  be the kernel  for the given feature mapping $\Phi:\R^d \to \R^{D_\Phi}$ with $D_\Phi\gg d$.
In this work, we consider the linear affine feature map: for any $\bu,\bv\in \R^d$, 
\bea
\label{eqn:linear_kernel}
\Phi(\bu) \coloneqq \bW \bu,\quad\text{with}\quad\bW\in\R^{D_\Phi \times d},
\eea
such that $\calK(\bu,\bv)=\bu^\sT\bW^\sT\bW\bv$.
Moreover, we shorthand $\{ \calK(\bxi_\mu, \bx) \}_{\mu=1}^M \in \R^M$ with $\calK(\bm{\Xi}, \bx)\in\R^M$.
With \eqref{eqn:linear_kernel},
we introduce the Kernelized Memory Hopfield Energy
\begin{align}
\label{eqn:eng-fn}
E_\calK(\bx) = 
\calK(\bx, \bx)/2  - \Psi_\alpha^\star\(\beta,\calK(\bm{\Xi}^\sT\bx)\),
\end{align}
where $\Psi_\alpha^\star$ is the convex conjugate of the Tsallis entropic regularizer introduced in \cite{wu2023stanhop,hu2023SparseHopfield}:
for any $\bz =\calK(\bm{\Xi}^\sT\bx)\in\R^M$,
\begin{align}
    \begin{cases}
        \Psi^\star_{\alpha=1}\(\beta,\bz\)=\text{lse} \(\beta, \bz \),\\
        \Psi^\star_{\alpha=2}\(\beta,\bz \)= \frac{1}{2} \norm{\beta \bz}^2 - \frac{1}{2} \norm{ {\bz^\star}  - \beta \bz }^2 + \frac{1}{2},\\
        \Psi^\star_{\alpha\in \[1, 2\]}\(\beta,\bz \)= \int \dd\bz \; \alpha\text{-EntMax}\(\beta \bz\)  ,
    \end{cases}
\end{align}
with $\bz^\star = \text{Sparsemax} \(\beta, \bz\)$.
\cref{app:entmax} includes the definitions of $\alpha\text{-EntMax}$ and $\text{Sparsemax}$.
\begin{assumption}
\label{asum:non-sing}
$\bW\in\R^{D_\Phi\times d}$ with $D_\Phi\gg d$ is full-rank.
\end{assumption}
\begin{remark}
    This assumption is practical and implies that $\bA = \bW^\sT \bW\in\R^{d\times d}$ is non-singular.
    In practice, initializing the weights randomly and independently with a continuous distribution (e.g., Gaussian) makes it almost impossible for $\bW$ to be non-full rank, especially if  $D_\Phi\gg d$.
\end{remark}
\begin{theorem}[Retrieval Dynamics]
\label{lemma:retrieval_dyn}
    With \cref{asum:non-sing}, the energy function $E(\bx)$  was monotonically decreased by the following retrieval dynamics:
\begin{align}
\label{eqn:ret-dyn}
    \calT_\calK\(\bx\) = \bm{\Xi} \cdot \text{Sep}_\alpha \( \beta, \calK\(\bm{\Xi}, \bx \) \),
\end{align}
where $\text{Sep}_{\alpha=1} \(\cdot\)=\text{Softmax} \(\cdot\) $, $\text{Sep}_{\alpha=2} \(\cdot\) = \text{Sparsemax} \(\cdot\)$ and 
$\text{Sep}_{\alpha\in[1,2]} \(\cdot\) = \alpha\text{-EntMax} \(\cdot\)$.
\end{theorem}

\begin{proof}[Proof Sketch]
By \cref{asum:non-sing} and the convexity of $\calK$, there exists an inverse map that transforms the CCCP results in kernel space back to the state space, where $\bx$ and $\bxi_\mu$ are located. 
We then complete the proof using the Concave-Convex Procedure (CCCP) and the convex conjugate construction following \cite{hu2023SparseHopfield,wu2023stanhop}.
See \cref{proof:retrieval_dynamics} for a detailed proof.
\end{proof}
The introduction of $\calK$ releases similarity measure from Euclidean inner-product to a learnable form via the weight $\bW$ of the features map $\Phi$.
Moreover, 
the new Hopfield model (\eqref{eqn:eng-fn} and \eqref{eqn:ret-dyn}) includes all deep learning compatible existing modern Hopfield models \cite{hu2023SparseHopfield,wu2023stanhop,ramsauer2020hopfield}.
If we replace the kernel $\calK(\cdot,\cdot)$ with inner-product $\Braket{\cdot,\cdot}$, then \eqref{eqn:eng-fn} reduces back to the general sparse model Hopfield model \cite{wu2023stanhop}\footnote{Recall that the general sparse Hopfield model encompasses both dense \cite{ramsauer2020hopfield} and sparse \cite{hu2023SparseHopfield} models as its special cases.}.

While \cref{lemma:retrieval_dyn} guarantees the 
monotonic minimization of energy using $\calT$,
the fixed point of $\calT$ might not be the local minima of $E(\bx)$ according to \citet{sriperumbudur2009convergence}.
Therefore, we provide the next lemma to ensure their alignment, following \cite{hu2023SparseHopfield, wu2023stanhop,ramsauer2020hopfield,sriperumbudur2009convergence}.
\begin{lemma}[Convergence on retrieval dynamics $\calT_\calK$]
\label{lemma:convergence}
    Given the energy function $E(\bx)$ \cref{eqn:eng-fn} and retrieval dynamics $\calT_\calK(\bx)$ \cref{eqn:ret-dyn}, respectively. 
    For any sequence $\{\mathbf{x}_t\}_{t=0}^{\infty}$ generated by the iteration $\mathbf{x}_{t'+1} = \calT_\calK(\mathbf{x}_{t'})$, all limit points of this sequence are stationary points of $E$.
\end{lemma}
\begin{proof}[Proof Sketch]
By the monotonic energy minimization property of $\calT$ (\cref{lemma:retrieval_dyn}) along with \citep[Lemma~2.2]{hu2023SparseHopfield}, 
we prove this through  
Zangwill’s global convergence theory \cite{zangwill1969nonlinear,sriperumbudur2009convergence}.
See \cref{proof:ret-conv} for a detailed proof.
\end{proof}
In summary, 
with $\Phi$, the parameterized similarity measure $\calK$ introduces an additional degree of freedom for us to relocate the minima of energy landscape $E_\calK$. 
We show that the Uniform Memory Hopfield  Energy \eqref{eqn:eng-fn} and its induced retrieval dynamics \eqref{eqn:ret-dyn} satisfies the defining properties of modern Hopfield models (\cref{lemma:retrieval_dyn} and \cref{lemma:convergence}). 
Importantly, 
\cref{lemma:convergence} states that minimizing the energy $E$ with $\mathcal{T}$ also leads to convergence to the fixed point of $\mathcal{T}$. 

This is pivotal in motivating our next step: constructing a separation loss $\mathcal{L}_\Phi$.
This loss distinguishes the local minima of $E_{\mathcal{K}}$ by separating stored memory patterns in the kernel space. 
With $\mathcal{L}_\Phi$, 
we then formulate the memory retrieval dynamics of the modern Hopfield associative memory model as a two-stage optimization, termed $\mathtt{U\text{-}Hop}$. 
This includes an additional stage of separation maximization (by learning the kernel), significantly enhancing memory capacity.

We first extend the standard notion of storage and retrieval (\cref{def:stored_and_retrieved_org})  literature using kernelized 
features ($\Phi(\bx)\in\R^{D_\Phi}$) to replace states of the model ($\bx\in\R^d$.)
\begin{definition}[Pattern Stored and Retrieved]
\label{def:stored_and_retrieved}
For all $\mu\in[M]$, let $R_\Phi\coloneqq \half \Min_{\nu\neq \mu;\nu,\mu\in[M]}\norm{\Phi(\bxi_\mu)-\Phi(\bxi_\nu)}$ be the finite radius 
 of each (kernelized) sphere $\calS_{\Phi,\mu}$ centered at (kernelized) memory pattern $\Phi(\bxi_\mu)$.
We say $\bxi_\mu$ is \textit{stored} if there exists a generalized fixed point of $\calT_\calK$, such that $\Phi(\bx^\star_\mu) \in \calS_{\Phi,\mu}$, to which all limit points $\Phi(\bx) \in \calS_{\Phi,\mu}$ converge to, and $\calS_{\Phi,\mu} \cap \calS_{\Phi,\nu}=\emptyset$ for $\nu \neq \mu$. 
We say $\bxi_\mu$ is $\epsilon$-\textit{retrieved} by $\calT_\calK$ with $\bx$ for an error $\epsilon$.
\end{definition}

\subsection{Separation Loss and {\uhop}}
\label{sec:UHop_alg}
\begin{algorithm}[t]
\caption{{\uhop}: Two-Stage Memory Retrieval}\label{algorithm1}
\textbf{Input:}  Separation (Stage I) iterations $N$, Energy (Stage II) iteration $T$, feature map $\Phi(\bx) \coloneqq \bW \bx$, memory set $\bm{\Xi}$, query $\bx$, retrieval dynamics $\calT$, learning rate $\gamma \leq 1/G$ where $G$ is the Lipschitz constant of $\calL_\Phi(\bm{\Xi})$  \\
\textbf{Output:} $\bx$  
\begin{algorithmic}[1]
\FOR{$i = 1, ... N$}  
    \STATE $\bW \leftarrow  \bW - \gamma \cdot \nabla_\bW \calL_\Phi( \bm{\Xi})$.\hfill {{// Stage I}}
\ENDFOR
\STATE Normalize the rows of $\bW$
\STATE $\bx^0 \leftarrow \bx$
\FOR{$t = 1, ... T$}
    \STATE $\bx\leftarrow \calT_\calK\(\bx\)$ using \cref{lemma:retrieval_dyn}
    \hfill {{// Stage II}}
\ENDFOR
\STATE return $\bx$
\end{algorithmic}
\end{algorithm}
In this section, 
we first introduce a separation loss $\mathcal{L}_\Phi$ (\cref{def:max-loss}) over the stored memory set $\bm{\Xi}$. 
Minimizing $\mathcal{L}_\Phi$ results in the separation of stored patterns within any given $\bm{\Xi}$. 
Consequently, we incorporate this separation-maximization step into the standard memory retrieval process (\eqref{eqn:ret-dyn}), leading to a novel two-stage formulation/algorithm, $\mathtt{U\text{-}Hop}$, for memory retrieval (\cref{algorithm1}).

For any $ \Phi(\bu), \Phi(\bv) \in\R^{D_\Phi}$ and some $t > 0$,
let 
\begin{align*}
    \calG_t(\Phi(\bu), \Phi(\bv)) 
    \coloneqq \exp{-t \norm{\Phi(\bu) -\Phi(\bv)}_2^2} ,
\end{align*}
be the Radial Basis Function ($\Phi$-RBF) kernel $\calG_t : \R^{D_\Phi} \times \R^{D_\Phi} \to \R_+$.
We introduce the objective for learning the feature map $\Phi:\R^d\to \R^{D_\Phi}$ (defined in \eqref{eqn:linear_kernel}) over the memory set 
$\bm{\Xi}=\{\bxi_\mu\}_{\mu\in[M]}$.
\begin{definition}[Average Separation Loss]
\label{def:max-loss}
Given 
a stored memory set $\bm{\Xi}$, and a feature map $\Phi : \R^d \to \R^{D_{\Phi}}$, 
the separation loss of the function $\Phi$ is 
\begin{align*}
        \mathcal{L}_{\Phi} \(\bm{\Xi};t \)
        \coloneqq
        \log  \underset{\bu,\bv \sim \bm{\Xi}}{\E} \[ 
        \calG_t\( \Phi(\bu),\Phi(\bv)\) \] , \; t > 0.
\end{align*}
\end{definition}
$\calL_\Phi$ indicates the logarithm of average Gaussian separation of $\Phi$ vector pairs over $\bm{\Xi}$.
Naturally, minimization of $\calL_\Phi$ leads to an on-average dissimilarity  among kernelized memory patterns, i.e., $\{\Phi(\bxi_\mu)\}_{\mu\in[M]}$.
Notably, $\calL_\Phi$ is convex by design and hence exists an optimizer $\bW^\star$ the maximizes the average distance between all possible memory pattern pairs.

For later convenience,
we also denote the logarithm of $\Phi$-RBF distance of two vectors $\bu, \bv \in \R^d$ as
\bea
\ell_\Phi(\bu, \bv) = \log \calG_t \(  \Phi(\bu), \Phi(\bv)\), \; t > 0.
\eea
It has a naive upper bound: $ \ell_\Phi\( \bu,\bv \) \leq 0$.
The upper bound $0$ happens only when $\bu = \bv$ or $\Phi$ outputs a fixed feature vector.

Now we introduce \cref{algorithm1}, the Uniform Memory Retrieval {{\uhop}}, for learning a suitable kernel and then retrieving stored memory from the learned kernel space.

\cref{algorithm1} is a 2-stage optimization process.
For the first stage, we run $N$ iterations of kernel learning to minimize the separation loss, thus resulting in larger $\Delta_\mu$ for $\mu\in[M]$.
Next, we rescale each row of the affine matrix $\bW$ to ensure the magnitude remains the same for memory patterns.
For the second stage, we run $T$ update steps for the retrieval dynamics, thus resulting in Hopfield energy minimization.
Note that the learned kernel results in a new energy landscape of $E$, and is expected to encode memory patterns into local minima that separates from all other memory patterns. 


\subsection{Exact Memory Retrieval}
\label{sec:exact_retrieval}
Let $\bx^\star$ be fixed points of $\calT$.
By \cref{def:stored_and_retrieved} and \citet[Definition~2.2]{hu2023SparseHopfield},
the retrieval error exhibits a naive bound
\begin{align*}
\norm{\calT_\calK(\bx)-\bxi_\mu}\le \Max\{\norm{\bx-\bxi_\mu},\norm{\bx^\star-\bxi_\mu}\}.
\end{align*}
The $\norm{\bx^\star-\bxi_\mu}$ term forbids the exact memory retrieval. 
Explicitly, exact memory retrieval requires the memory pattern to be the fixed point of $\calT$, namely $\norm{\bx^\star-\bxi_\mu}=0$.
With this observation, we deduce the condition of exact retrieval
\begin{align}\label{eqn:exact}
    \text{Sep} \( \beta, \calK \( \bm{\Xi}, \bxi_\mu \) \) = \be_\mu,
\end{align}
where $\be_\mu$ is the one-hot vector with the $\mu$-th element as $1$.
By plugging $\bxi_\mu$ into $\calT(\cdot)$, we see it is a fixed point $\calT(\bxi_\mu) = \bxi_\mu$ and retrieves the target memory $\bxi_\mu$ only when \eqref{eqn:exact} holds.
In the standard modern Hopfield model (utilizing the $\Softmax$ Sep function), 
the inability of $\Softmax$ to satisfy \eqref{eqn:exact} results in a lack of exact retrieval \cite{martins2023sparse}, 
thereby preventing the modern Hopfield network from converging to a single memory pattern.

To combat this,
we show {\uhop} achieves exact memory retrieval when $\alpha > 1$, based on the sparse extensions of modern Hopfield model \cite{wu2023stanhop,hu2023SparseHopfield, martins2023sparse}.
Specifically, we study the application of $\mathtt{U\text{-}Hop}$ with $\alpha$-EntMax as separation when $\alpha > 1$.
\begin{theorem}\label{thm:exact}
Let $\calT_{\text{sparse}}$ be $\calT_\calK$ from \cref{lemma:retrieval_dyn} with $\alpha>1$.
    Let $\calT_{\text{sparse}}$ a real-valued kernel $\calK$ with feature map $\Phi$.
    Let $t>0, \beta > 0$.
    Supposed the query $\bx \in \calS_{\Phi, \mu}$, $\Phi\( \bxi_\mu \)$ is the fixed point of $\calT_{\text{sparse}}$ if the following condition is satisfied:
    \begin{align}\label{prop:eq1}
         \ell_\Phi \(\bxi_\mu, \bxi_\mu \) -  \underset{\nu, \nu \neq \mu}{\max}\ell_\Phi \(\bxi_\nu, \bxi_\mu \) 
        \leq -\frac{2t}{\beta (\alpha - 1) }.
    \end{align}
\end{theorem}
\begin{proof}
See \cref{proof:optimal} for a detailed proof.
\end{proof}
From \eqref{prop:eq1}, minimizing the separation loss gives the benefit of having the memory pattern to be the fixed point of $\calT_{\text{sparse}}$.
As a result, Sparse and Generalized Sparse Hopfield models \cite{hu2023SparseHopfield, martins2023sparse, wu2023stanhop} under \uhop $\;$further improves the retrieval accuracy.
The next corollary is an extension of the above theorem where we observe the condition with respect to the lipschitzness of $\Phi$.
\begin{corollary}
\label{coro:exact_coro}
    Let $L>0$ be the Lipschitz constant of $\Phi$.
    Following  \cref{thm:exact}, $\calT_{\calK}$ achieves exact memory retrieval if 
    \begin{align*}
        \Min_{\nu\in[M],\nu\neq\mu}{\norm{\bxi_\mu - \bxi_\nu}} \geq \sqrt{\frac{2}{L^2 \beta (\alpha-1) }}.
    \end{align*}
\end{corollary}

\begin{proof}
    See \cref{proof:optimal} for a detailed proof.
    Note that with $\Phi$ defined in \eqref{eqn:linear_kernel}, $\Phi$ is always $L$-Lipschitz.
\end{proof}

\section{Connecting to Modern Deep Learning}
\label{sec:DL}

\begin{figure*}[t]
\centering
\begin{minipage}[b]{0.48\linewidth}
\centering
\includegraphics[width=1.025\linewidth]{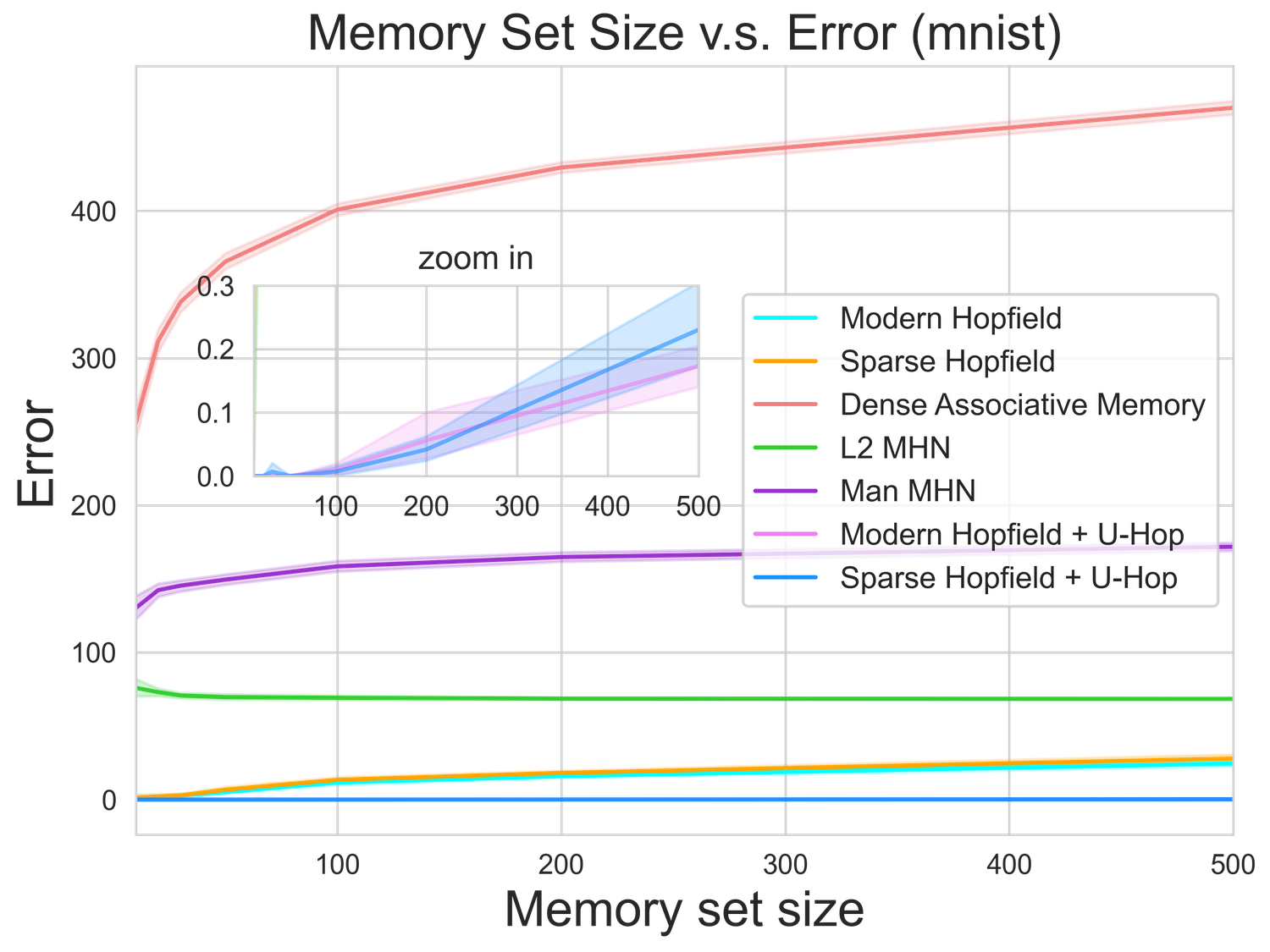}
\end{minipage}
\begin{minipage}[b]{0.48\linewidth}
\centering
\includegraphics[width=1.025\linewidth]{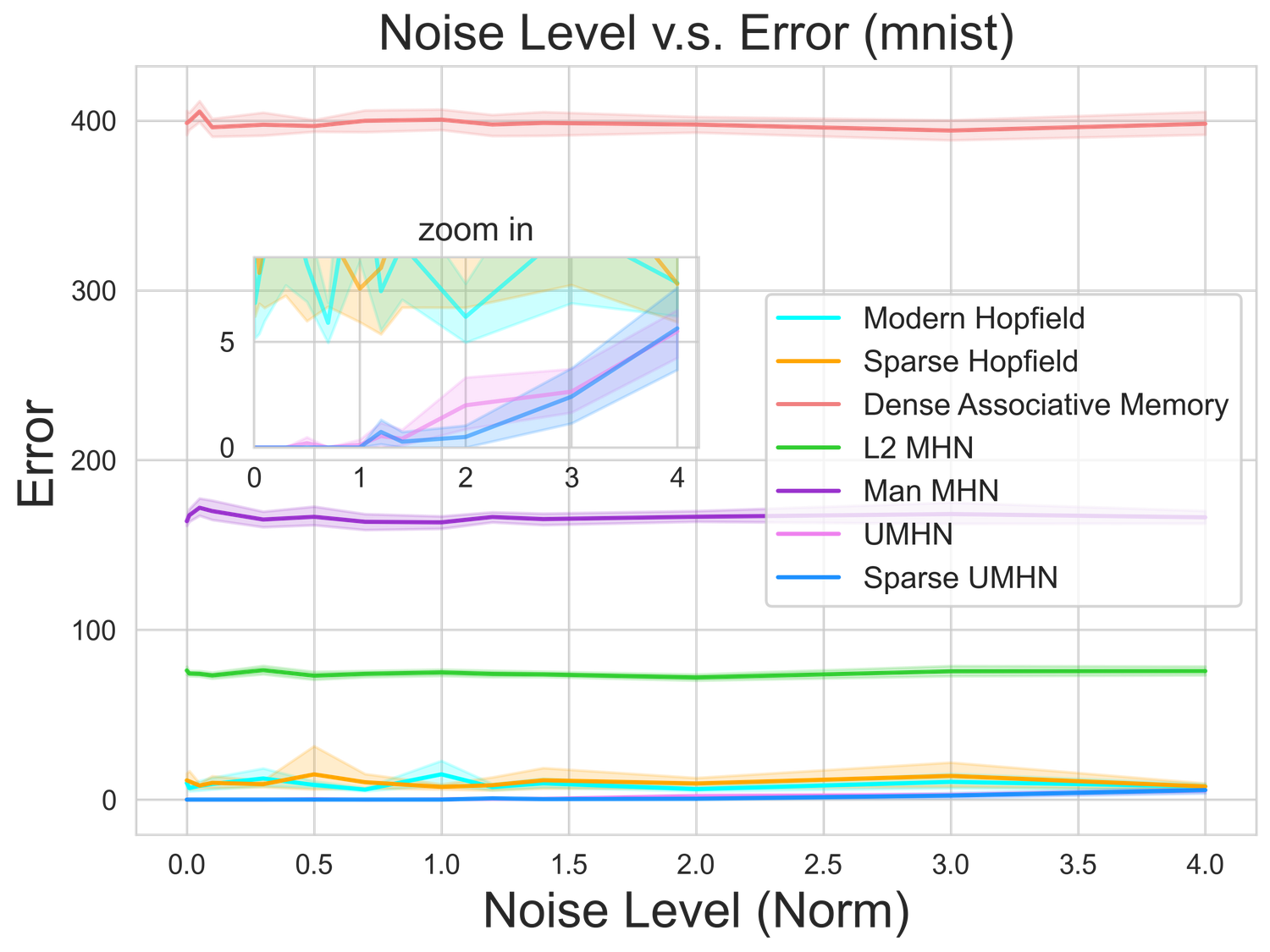}
\end{minipage}
\\ %
\begin{minipage}[b]{0.48\linewidth}
\centering
\includegraphics[width=1.025\linewidth]{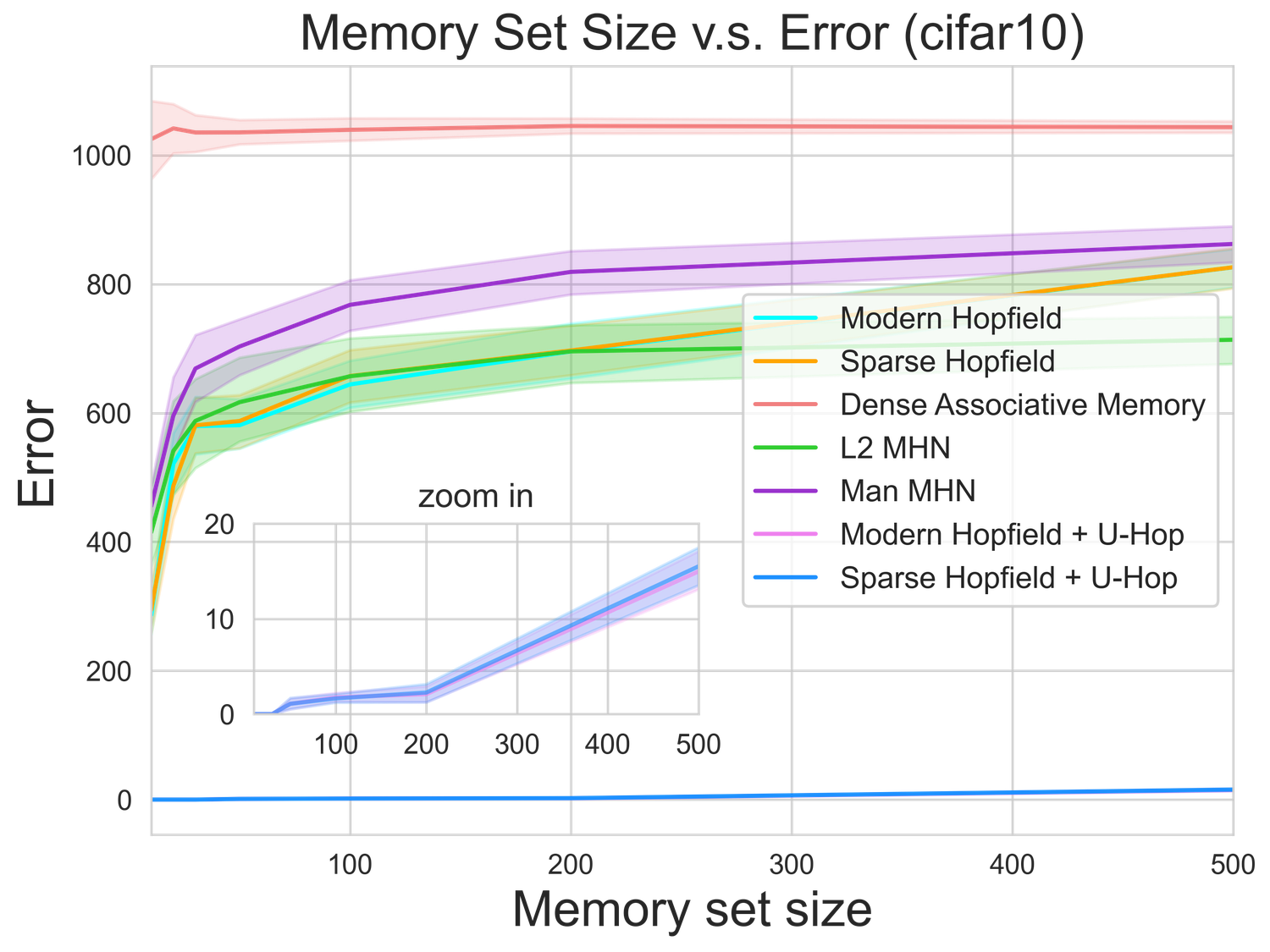}
\end{minipage}
\begin{minipage}[b]{0.48\linewidth}
\centering
\includegraphics[width=1.025\linewidth]{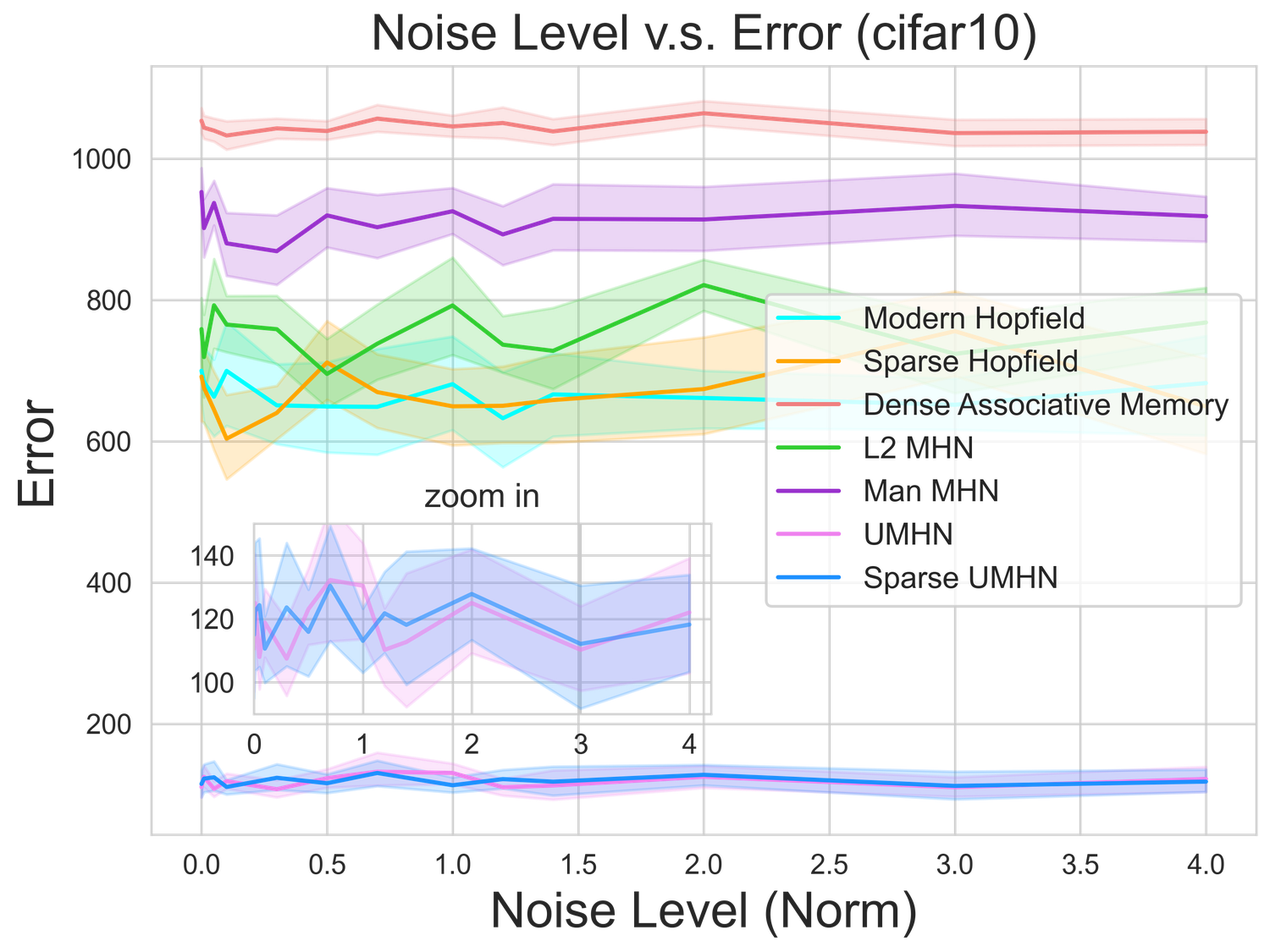} %
\end{minipage}
\vspace{-1em}
\caption{\textbf{Memory Retrieval Error Comparison (\cref{seec:exp_memory}: Memory Capacity \& Noise Robustness)}. We conduct memory retrieval experiments on the MNIST and CIFAR10 datasets. 
For the ``Memory Set Size v.s. Error'' plots, we vary the memory set size for retrieval. 
For the ``Noise Level v.s. Error'' plots, we randomly sample Gaussian noise and rescale the norm of the noise w.r.t. different noise levels. All four plots show U-Hop retrieved patterns with significantly less error compared to all existing Hopfield models across all sizes of memory and noise levels.}
\label{fig:ret-error}
\end{figure*}

To incorporate {\uhop} into deep learning, we first introduce a kernelized Hopfield layer. 
Here we propose a deep learning compatible layer based on {\uhop} as
\begin{align*}
    &\mathtt{U\text{-}Hop}(\bm{\Xi}, \bm{X}) 
    = \text{Sep} \( \beta \bW_K \Phi(\bm{\Xi}) \bW_Q \Phi(\bx) \) \bW_V \bW_K \bm{\Xi}.
\end{align*}
Note that this is a kernelized version of $\mathtt{Hopfield}$ \cite{ramsauer2020hopfield}, $\mathtt{SparseHopfield}$ \cite{hu2023SparseHopfield} and $\mathtt{GSH}$ \cite{wu2023stanhop} layers, which serve as an alternative to attention mechanism variants.

Next, we introduce the average separation loss for deep learning compatible {\uhop}.
\begin{definition}[Separation Loss for DL]\label{dfn:separation-loss}
Given 
a stored memory set $\bm{\Xi}$, and a feature map $\Phi: \R^d \to \R^{D_{\Phi}}$, 
the deep learning compatible separation loss of the function $\Phi$ is defined as
\begin{align*}
        \bar{\mathcal{L}}_{\Phi} \(\bm{\Xi};t \)
        \coloneqq
        \log  \underset{\bu,\bv \sim \bm{\Xi}}{\mathbb{E}} \[ 
        \exp{2t  \[\calK(\bu,\bv)^2 - 1\]}\] \; t > 0.
\end{align*}
\end{definition}
Let the pairwise distance of $\bar{\calL}_\Phi$ for any given $\bu, \bv \in \R^d$ be
\begin{align*}
    \bar{\ell}_{\bar{\Phi}}(\bu, \bv)
    =
    2t \[ \bar{\calK}_{\bar{\Phi}}(\bu, \bv) - 1 \], t > 0,
\end{align*}
with $\Bar{\calK}(\cdot,\cdot)\coloneqq \Braket{\Bar{\Phi}(\cdot),\Bar{\Phi}(\cdot)}$ for some $\Bar{\Phi}:\R^d\to\R^{D_{\Bar{\Phi}}}$.
We present next the theorem for the expressiveness of $\mathtt{U\text{-}Hop}$.
\begin{theorem}[Kernelized Representation Theorem]\label{thm:perfect_kernel}
Let $\bar{\Phi}$ be a  feature map such that $\bar{\Phi} \coloneqq \R^d \rightarrow \R^{D_{\bar{\Phi}}}$, and $\bar{\calK}$ be $\bar{\Phi}$-induced kernel.
    Assuming $\bar{\calK}$ satisfies: $\bar{\ell}_{\bar{\Phi}}(\bu, \bv) = -2t$ for any given $\bu, \bv \in \bm{\Xi}$.
    With $\beta > 0$, input $\bX \in \R^{d \times M}$, $M \leq d$, an arbitrary positive column stochastic matrix $\bP \in \R^{M \times M}$, there always exists matrices $\bW_Q, \bW_K$ such that 
    \begin{align*}
        \text{Softmax} \(  \beta \; (\bW_K \bar{\Phi}(\bX))^\top \bW_Q \bar{\Phi}(\bX)\) = \bP.
    \end{align*}
Specifically, $\bar{\calK}$ is the loss minimizer of $ \bar{\mathcal{L}}_{\Phi} \(\bm{\Xi};t \)$.
\end{theorem}
\begin{proof}
   See \cref{proof:representation} for a detailed proof. 
\end{proof}
The empirical validation is in \cref{sec:additional-exp}. 
This theorem shows that with a suitable kernel, the expressiveness of $\mathtt{Hopfield}$ layers under {\uhop} reaches its full potential. 
The main difference between this new loss function and the separation loss is the square on $\calK(\bu, \bv)$. 
Note that this theorem requires $\bar{\calK}(\bu, \bv) = 0$ for any given $\bu, \bv \in \bm{\Xi}$, $\bu \neq \bv$, which implies it is only possible when $d \geq M$.
In the context of deep learning, the patch size must not be larger than the hidden dimension to realize this result. 
This theorem extends the representation theorem in \cite{bhojanapalli2020low} to a practical setting, showing that $\bar{\calK}$ overcomes the low-rank bottleneck of the attention mechanism and Hopfield layer as well.

The next algorithm is the realization of searching for $\bar{\calK}$ under supervised learning schema.
Consider a supervised learning problem with input data $\calX = \{\bX_1, ..., \bX_n\}$, label $\calY = \{\bY_1, ..., \bY_n\}$, model $F \coloneqq \calX \rightarrow \calY$, 
where $F$ consists of one layer of ``$\mathtt{U\text{-}Hop}$ + Hopfield layer".
The stage-I of {\uhop} is parameterized by $\theta$, and $F$ is parameterized be $\theta_F$.

\begin{algorithm}[H]
\caption{$\mathtt{U\text{-}Hop}$ for Learning}\label{algo:bi-level}
\textbf{Input: } Data $\calD=(\calX,\calY)$, Iteration number $N_i, N_o$, model: $F: \calX \rightarrow \calY$, step sizes $(\bar{\gamma}, \bar{\alpha})$, training objective $\mathcal{L}$, Stage I SGD batch size $B_1$, Stage II SGD batch size $B_2$
\begin{algorithmic}[1]
\FOR{$i = 1$ to $N_o$}
    \FOR{$j = 1$ to $N_i$}
        \STATE Sample mini-batch $\bX \sim \calX$
        \STATE $\theta^i = \theta^{i-1} - \bar{\gamma} \hat{\grad}
        \bar{\mathcal{L}}_\Phi(x)$. \hfill {{// SGD with $B_1$}}
    \ENDFOR
    \STATE Sample mini-batch $\bX,\bY \sim \calX, \calY$
    \STATE $\theta_F^j = \theta_F^{j-1} - \bar{\alpha} \hat{\grad} \mathcal{L}(x,y)$ \hfill {{// SGD with $B_2$}}
\ENDFOR
\STATE return $F$
\end{algorithmic}
\end{algorithm}

\section{Experimental Studies}
\label{sec:exp}\begin{table*}[t]
    \centering
        \caption{\textbf{Model maximal training accuracy and test accuracy with and without {\uhop} on CIFAR10, CIFAR100 and Tiny ImageNet (\cref{sec:exp_sl}: Supervised Learning Tasks).} 
    MHM denotes Modern Hopfield Model \cite{ramsauer2020hopfield}.
    We omit variance as all variance are $\leq 0.03\%$.
    The result demonstrates with {\uhop}, models are able to consistently memorize more samples in the training data, and further obtain generalization improvement.
    Note that the improvement on Max. Training accuracy is a validation of \cref{thm:perfect_kernel}.
    In \cref{sec:additional-exp}, we also show {\uhop} allows modern Hopfield models to converge faster.
    }
    \vspace{0.5em}
    \resizebox{ \textwidth}{!}{%
    \begin{tabular}{lcccccc}
    \toprule 
        Models & \multicolumn{2}{c}{\textbf{CIFAR10}}  & \multicolumn{2}{c}{\textbf{CIFAR100}} &\multicolumn{2}{c}{\textbf{Tiny ImageNet}  }  \\
        \midrule 
        & Max Train Acc. & Test Acc.  & Max Train Acc. & Test Acc.  &Max Train Acc. & Test Acc. \\
        \hline
        MHM &  $56.0\%$ & $52.2\%$ & $32.3\%$ & $26.3\%$ & $48.9\%$ & $12.2\%$ \\
        MHM + {\uhop} & \cellcolor{LightCyan} $\bm{64.6\%}$ & \cellcolor{LightCyan} $\bm{55.2\%}$ & \cellcolor{LightCyan} $ 
        \bm{44.1\%}$ & \cellcolor{LightCyan} $\bm{28.7\%}$ & \cellcolor{LightCyan} $\bm{61.4\%}$ & \cellcolor{LightCyan} $\bm{12.7\%}$ \\
        \hline
        Sparse MHM & $55.9\%$ & $52.0\%$ & $49.6\%$ & $26.0\%$ & $17.2\%$ & $12.3\%$ \\
        Sparse MHM + {\uhop} & \cellcolor{LightCyan} $ \bm{66.4\%}$ & \cellcolor{LightCyan} $\bm{55.4\%}$ & \cellcolor{LightCyan} $\bm{45.4\%}$ & \cellcolor{LightCyan} $\bm{29.0\%}$ & \cellcolor{LightCyan} $\bm{60.6\%}$ & \cellcolor{LightCyan} $\bm{12.5\%}$  \\
        \bottomrule
    \end{tabular}
    }
    \label{tab:img-cls}
\end{table*}
\begin{figure*}[!h]
    \centering
    \includegraphics[width=\textwidth]{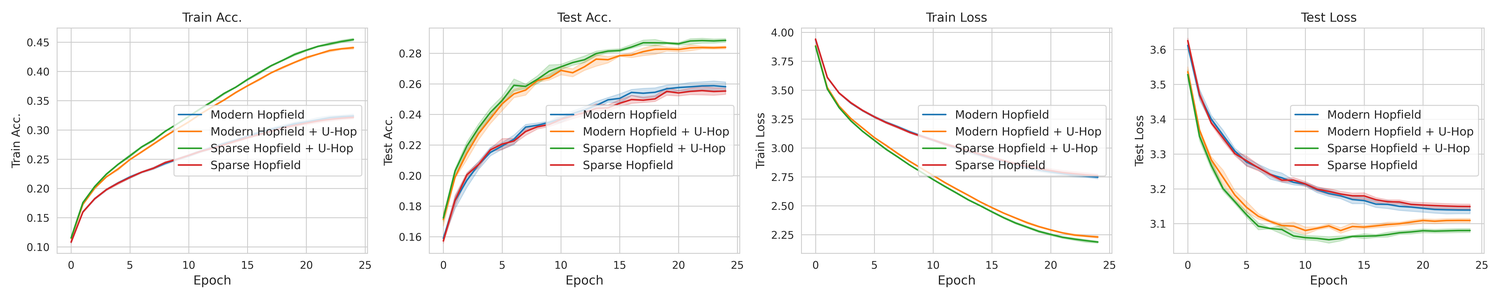}
    \vspace{-2em}
    \caption{
    \textbf{Model Convergence Comparison with  and without {\uhop} on CIFAR100 (\cref{sec:exp_sl}: Image Classification Task).}    
    Left to right: Training Accuracy, Test Accuracy, Training Loss and Test Loss.
    Yellow and green curves represent modern Hopfield + {\uhop} and Sparse modern Hopfield + {\uhop}.
    Blue and red curves represent modern Hopfield and Sparse modern Hopfield.
    The result demonstrates without {\uhop}, Hopfield layers fall into the low-rank bottleneck \cite{bhojanapalli2020low} despite of high embedding dimension.
    On the other hand, {\uhop} successfully avoid such issue and thus have better training accuracy.
    For generalization power and convergence speed, {\uhop} also outperforms other baselines by a large margin.
    For other datasets and sample size, we leave the results in \cref{sec:additional-exp}.
    }
    \label{fig:cifar100-main}
\end{figure*}

To validate the efficacy of \uhop,
we test it on both associative memory retrieval task 
and deep learning task (image classification) with multiple real world datasets.

\subsection{Memory Retrieval}
\label{seec:exp_memory}
\paragraph{Memory Capacity.}
The memory retrieval task involves retrieving a memory pattern from a stored memory set.
In particular, this experiment aim to reconstruct memories based on a query.
The query is generated by randomly masking 50\% of pixels in the target image.
We compare our method against several modern Hopfield models \cite{hu2023SparseHopfield,ramsauer2020hopfield,krotov2016dense}.
We also vary the iteration number $N$ for the first stage in \cref{algorithm1}.
We use MNIST, CIFAR10 datasets for this task.
Please see \cref{app:exp-detail} for  experimental details.

\paragraph{Noise Robustness.}
This experiment follows the same procedure as the memory capacity tasks, but with multiple levels of injected noise on the target image instead of masking out pixels to generate queries.
We use Gaussian noise to contaminate the queries and vary the noise level by altering the mean of the Gaussian vectors.
As the noise level increases, it becomes more difficult to retrieve the memory with low error. A higher noise level results in greater difficulty in achieving low-error retrieval.
We use MNIST, CIFAR10 for this task.
Please see \cref{sec:additional-exp} for experimental details.

\paragraph{Baselines.}
We compare our method with  Modern Hopfield Model \cite{ramsauer2020hopfield}, Sparse Modern Hopfield network \cite{hu2023SparseHopfield}, Dense Associative Memory (Polynomial Hopfield) \cite{krotov2016dense} (using $10$-th order polynomial energy function).
We also compare {\uhop} with existing similarity measures: L2 distance ($\ell_2$ MHM) and Manhattan distance (Man. MHM) \cite{millidge2022universal}.

\paragraph{Setting and Metrics.}
We set $\beta = 1, t = 2$ across all the memory retrieval experiments.
For the evaluation metric, we follow \cite{hu2023SparseHopfield,millidge2022universal} to use the Sum-of-Square pixel differences between the ground truth image and the retrieved image.

\begin{figure*}[!t]
\centering
\includegraphics[width=\textwidth]{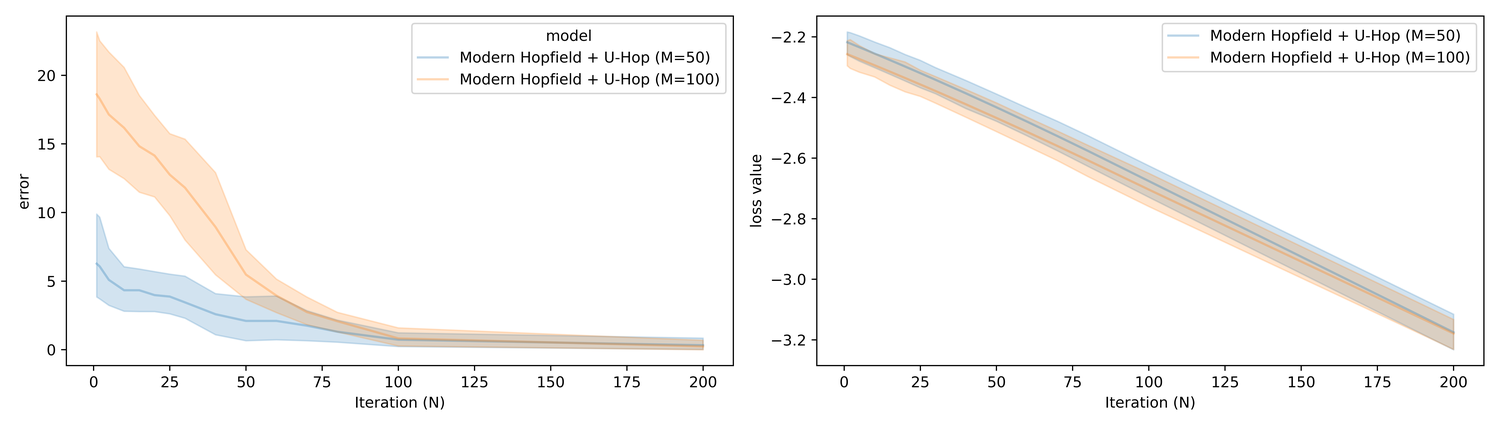} 
\vspace{-2.5em}
\caption{
\textbf{Retrieval Error v.s. Separation-Maximization (Stage I of \cref{algorithm1}) Iteration $N$ (\cref{seec:exp_memory}).}
We vary the iteration number $N$ and perform memory retrieval on \uhop$\;$ with modern Hopfield.
We set $\beta=1, t=2$ and report the sum-of-square pixel differences.
The result shows the retrieval error decays fast with respect to the increase of $N$.
}
\label{fig:kernel-epoch}
\end{figure*}

\paragraph{Results.} See \cref{fig:ret-error} for results of memory capacity and noise robustness, \cref{fig:kernel-epoch} for results of ``Stage I iteration improve retrieval error'' and \cref{sec:sep-vs-error-scatter} for the relationship between separation loss and retrieval error.
\begin{itemize}
    \item 
    For \textbf{memory capacity}, {\uhop} outperforms all other baselines by a large margin.
    This result shows the retrieval dynamics under {\uhop} is near optimal across all memory set sizes.
    Next, we vary the iteration $N$ to observe how fast the retrieval error decreases as the $N$ goes up.
    In \cref{fig:kernel-epoch}, we show a strong correlation between $N$ and retrieval error.
    \item 
    For \textbf{noise robustness}, {\uhop} shows strong performance against all baselines as well as showed in \cref{fig:ret-error}.
\end{itemize}

\subsection{Supervised Learning Tasks}
\label{sec:exp_sl}

\begin{figure}[!ht]
\begin{minipage}[h]{0.49\linewidth}
\begin{center}
\includegraphics[width=1.1\linewidth]{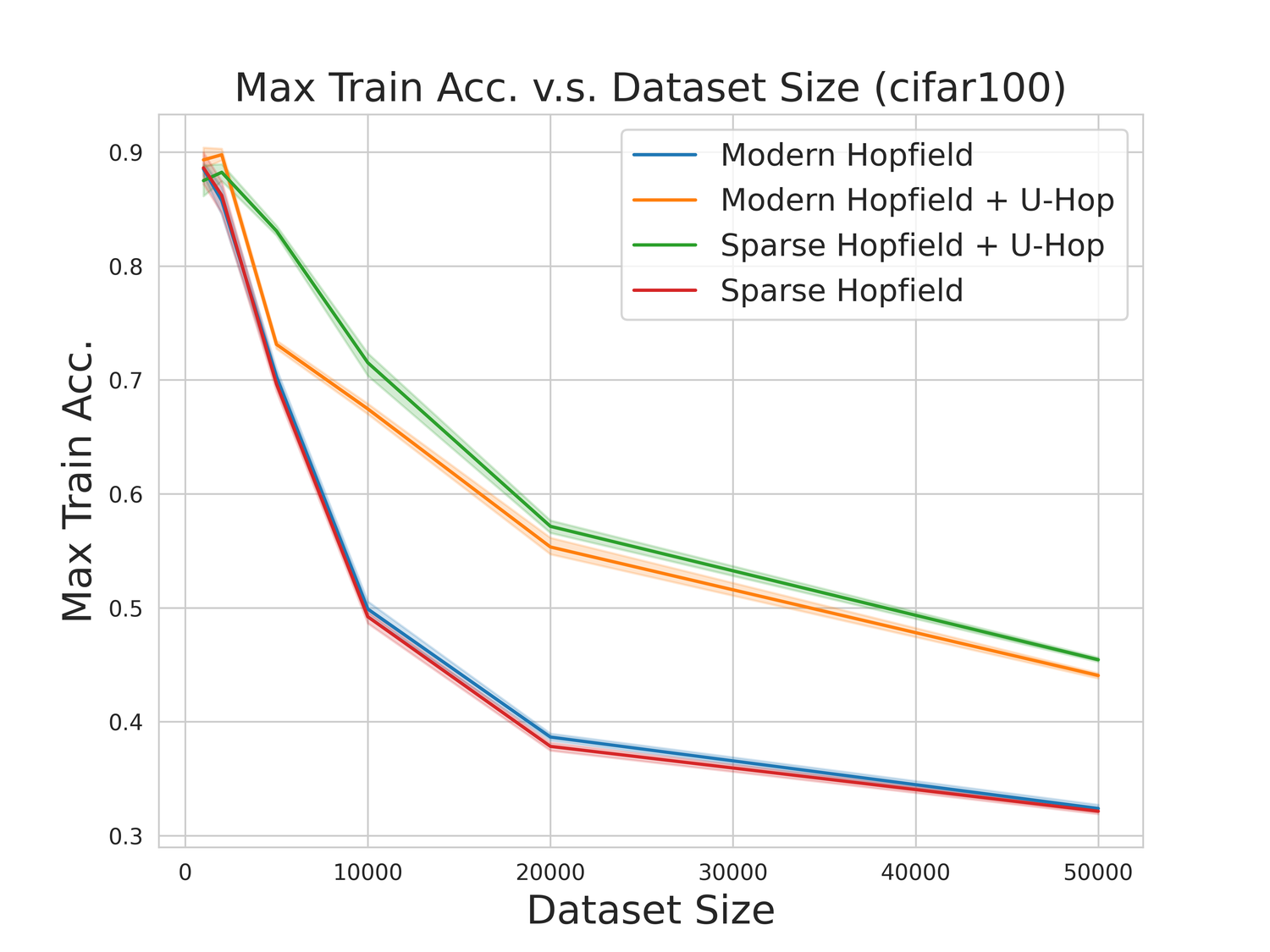} 
\end{center} 
\end{minipage}
\hfill
\begin{minipage}[h]{0.49\linewidth}
\begin{center}
\includegraphics[width=1.1\linewidth]{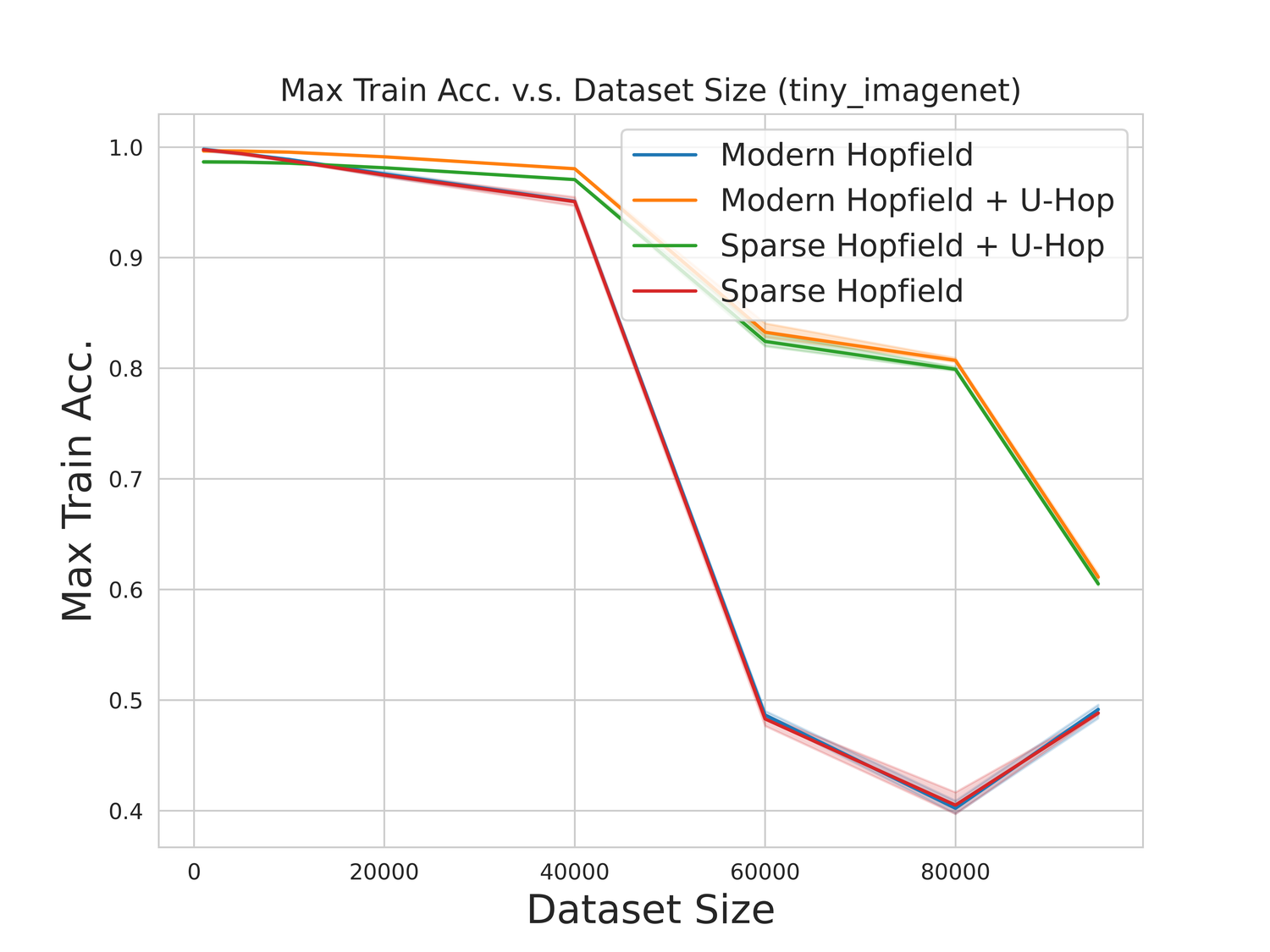} 
\end{center}
\end{minipage}
\caption{\textbf{Maximum Training Accuracy v.s. Training Sample Size (\cref{sec:exp_sl}:  Expressiveness).} 
Here we report the train accuracy comparison between modern Hopfield models with and without {\uhop}.
The maximum training accuracy represents how many percentages of samples a model memorizes, which is highly related to model expressiveness and complexity.
Note that using {\uhop} does not increase model complexity, which shows {\uhop} improves model expressiveness by a large margin.
}
\label{fig:expresivenss}
\end{figure}

\paragraph{Image Classification.}
For classification tasks, we compare our method against $\mathtt{Hopfield}$ \cite{ramsauer2020hopfield} and $\mathtt{SparseHopfield}$ \cite{hu2023SparseHopfield}.
We test two settings: 
\begin{itemize}
    \item {\uhop} + Dense Modern Hopfield Model \cite{ramsauer2020hopfield}, and
    \item {\uhop} + Sparse Modern Hopfield Model \cite{hu2023SparseHopfield}.
\end{itemize}\vspace{-1em}
We vary the training sample size and observe model performance.
We focus on (i) convergence speed (speed of loss decay), (ii) generalization power (test accuracy).
We use CIFAR10, CIFAR100 and TinyImageNet for this task.
Please see \cref{app:exp-detail} for more experimental details.

We use the following $\mathtt{Hopfield}$ layer \cite{ramsauer2020hopfield} to replace the self-attention mechanism in Vision Transformer:
\begin{align*}
    &~ \mathtt{Hopfield}\left(\bm{X}\right) \\
    = &~
    \bW_V \bW_K \cdot \Softmax \left( \beta \bW_K \Phi( \bX ) \bW_Q \Phi(\bX) \right),
\end{align*}
where $\bW_K, \bW_Q$ are the same as in self-attention, and $\bW_V \in \R^{h \times h}$, where $h$ is the hidden dimension.
\paragraph{Expressiveness.}
To verify \cref{thm:perfect_kernel}, we evaluate how many samples a model can memorize in supervised learning task.
We follow the image classification settings, and see how Hopfield models with and without {\uhop} react to sample size growth.

\paragraph{Time Series Prediction.}
We also use the $\mathtt{STanHop}\text{-}\mathtt{Net}$  \cite{wu2023stanhop} as our test-bed and observe the performance change with and without \uhop.
For this task, we use ETTh1, ETTm1 and WTH datasets.
We use the prediction horizon of $\{96, 192, 336, 720\}$ for all datasets.
Please see \cref{app:exp-detail} for experimental and hyperparameter details.

\paragraph{Baselines, Setting and Metrics.}
We compare the performance of Modern Hopfield and Sparse Hopfield with and without \uhop.
For image classification, we use Vision Transformer \cite{dosovitskiy2020image} as test-bed and replace the attention mechanism with $\mathtt{Hopfield}$ \cite{ramsauer2020hopfield} and $\mathtt{SparseHopfield}$ layer \cite{hu2023SparseHopfield}.
For time series prediction,
we compare the performance of $\mathtt{STanHop\text{-}Net}$ \cite{hu2023SparseHopfield} with and without \uhop.

\paragraph{Results.} See \cref{tab:img-cls} for convergence results of image classification task, \cref{fig:expresivenss} for expressiveness results (\cref{thm:perfect_kernel})  and \cref{table:time-series} for time series prediction.
\begin{itemize}
    \item
    For \textbf{image classification}, we observe that modern Hopfield models under {\uhop} consistently outperform other baselines, and the performance gap increases with the sample size growth.
    Additionally, $\mathtt{U\text{-}Hop}$ models shows superior convergence speed comparing to other baselines on both training and test set.
    For model generalization, see \cref{tab:img-cls}, for convergence results, see \cref{app:cls-extra}.
    \item 
    For \textbf{model expressiveness}, we observe that when the dataset size is small, $\mathtt{U\text{-}Hop}$ has similar memorization capability as $\mathtt{Hopfield}$.
    However, as the dataset size increases, $\mathtt{Hopfield}$ without {\uhop} shows a sharp degeneration on training accuracy and struggles to converge well, as evidenced in \cref{fig:expresivenss}.
    For more detailed results, see \cref{sec:expressive}.
    \item 
    For \textbf{time series prediction}, 
    our results (in \cref{sec:time-series}) demonstrate that even on SOTA Hopfield-based time series model, {\uhop}  delivers  performance improvement across different datasets and prediction horizons.
\end{itemize}

\subsection{More Discussions on Experimental Results}
For memory retrieval tasks,
{\uhop} delivers significant improvements on retrieval error by lowering separation loss over the memory set.
For the epochs $N$ required for kernel learning, 
we demonstrate that low retrieval error has strong correlation with large size of $N$.
This is expected as our separation loss is convex and guaranteed to obtain global optima with a rate of $\calO(1/N)$.
As showed in \cref{fig:kernel-epoch}, separation loss consistently decreased as $N$ goes up.

For classification tasks, $\mathtt{U\text{-}Hop}$ delivers significant improvements in predictive power of the underlying models.
Comparing to contrastive self-supervised learning \cite{wang2020understanding, chen2020simple}, where they maximize pairwise distance between samples, {\uhop} maximizes the pairwise distance between patches.
As \citet{saunshi2022understanding} show maximizing the distance over samples improves class generalization and is beneficial to downstream tasks.

Our experiment results indicate 2 new insights that the separation on the patch/token level also leads to better generalization.
Firstly, 
we hypothesize that the Stage~I of {\uhop} serves as a pre-training step for a better representation with more separated data geometry.
Namely, tokens/patches projected to kernel space have higher quality of representation as {\uhop}'s first iteration leads to better patch separation.
Secondly, 
though `` $\mathtt{Hopfield}$ layers with and without {\uhop}'' and ``expressiveness'' experiments (\cref{tab:img-cls} and \cref{fig:expresivenss}), 
we observe that solely increasing embedding dimension do not guarantee to escape from the low-rank bottleneck in attention- and Hopfield-based models \cite{bhojanapalli2020low}.
We conclude that this is because these models do not utilize their full expressive power (as in \cref{thm:perfect_kernel}), despite of high embedding dimension.
This observation supplements the existing ``high-dimensional embedding improves low-rank bottleneck'' conjecture \cite{bhojanapalli2020low} with an intuitive yet effective learning scheme.

\section{Concluding Remarks}
\label{sec:conclusion}We present a two-stage  formulation for memory retrieval of modern Hopfield models, {\uhop}.
Our key contribution is a learnable similarity measure utilizing the stored memory patterns as learning data.
Through our analyses, {\uhop} is theoretically grounded and empirically strong.
Experimentally, 
it improves memory retrieval tasks by an average 30\% margin even with only a single separation-maximization iteration and learning tasks by an average 3\% margin.
These results are benchmarked against STOA similarity measures ($\ell_2$- and Manhattan- distance \cite{millidge2022universal}) and existing modern Hopfield models \cite{wu2023stanhop,hu2023SparseHopfield,ramsauer2020hopfield,krotov2016dense}.

\paragraph{Complexity Analysis.}
\cref{algorithm1} has a time complexity of $\mathcal{O}(N + T)$. 
\cref{algo:bi-level} has a time complexity of $\mathcal{O}(N_o N_i)$. 
Although this increases the standard supervised learning training time by a factor of $N_i$, our experimental results demonstrate that models under $\mathtt{U\text{-}Hop}$ mitigate this issue with a faster convergence speed, requiring fewer epochs to converge.
See \cref{app:cls-extra} for related empirical results.

\paragraph{Limitation and Future.} 
One notable limitation is that the optimality of  separation loss (\cref{def:max-loss}) does not guarantee maximal separation for $R \coloneqq \frac{1}{2} \min_{\mu,\nu \neq \mu \in [M]} \|\boldsymbol{\xi}_\mu - \boldsymbol{\xi}_\nu\|$ for any given $\boldsymbol{\Xi}$.
This problem (maximizing $R$) is inherently a max-min (non-convex) problem and is less straightforward to analyze (Comparison between max and avg. loss is in \cref{sec:max-vs-avg}). To achieve provably optimal memory capacity, we plan to explore different loss functions or learning schemes in the future.

\paragraph{Note Added [November 2024].}
To address the above limitation, the follow-up work \cite{hwl24} presents a provably optimal memory capacity bound for kernelized modern Hopfield models and introduces a sub-linear time algorithm, $\mathtt{U\text{-}Hop}$+, to achieve this optimal capacity.

\section*{Impact Statement}
This research is theoretical and is not expected to have negative social impacts. As outlined in the introduction and related works, the primary goal of this study is to enhance our understanding of the underlying principles of large Hopfield-based and transformer-based foundation models from an associative memory perspective.

\section*{Acknowledgments}
JH would like to thank Stephen Cheng, Shang Wu, Dino Feng and Andrew Chen for enlightening discussions, the Red Maple Family for support, and Jiayi Wang for facilitating experimental deployments.
The authors would also like to thank the anonymous reviewers and program chairs for their constructive comments.

JH is partially supported by the Walter P. Murphy Fellowship.
HL is partially supported by NIH R01LM1372201, NSF CAREER1841569, DOE DE-AC02-07CH11359, DOE LAB 20-2261 and a NSF TRIPODS1740735.
This research was supported in part through the computational resources and staff contributions provided for the Quest high performance computing facility at Northwestern University which is jointly supported by the Office of the Provost, the Office for Research, and Northwestern University Information Technology.
The content is solely the responsibility of the authors and does not necessarily represent the official
views of the funding agencies.

\clearpage
\newpage
\normalsize
\titlespacing*{\section}{0pt}{*1}{*1}
\titlespacing*{\subsection}{0pt}{*1.25}{*1.25}
\titlespacing*{\subsubsection}{0pt}{*1.5}{*1.5}

\setlength{\abovedisplayskip}{10pt}
\setlength{\abovedisplayshortskip}{10pt}
\setlength{\belowdisplayskip}{10pt}
\setlength{\belowdisplayshortskip}{10pt}

\setlist[itemize]{ before=\vspace{-0.5em}, after=\vspace{-0.5em}, itemsep=0.1em}
\setlist[enumerate]{before=\vspace{-0.5em}, after=\vspace{-0.5em}, itemsep=0.1em}

\onecolumn
\appendix
\part*{Supplementary Material}
\label{sec:append}

{
\setlength{\parskip}{-0em}
\startcontents[sections]
\printcontents[sections]{ }{1}{}
}

\begin{itemize}
    \item \hyperref[sec:tab_notation]{\textbf{Section~A.}} \textbf{Table of Notations}
    \item \hyperref[sec:related_work]{\textbf{Section~B.}} \textbf{Related Work}
    \item \hyperref[sec:connect-attention]{\textbf{Section~C.}} \textbf{Connection to Attention}
    \item \hyperref[sec:sup-theory]{\textbf{Section~D.}} \textbf{Supplementary Theoretical Backgrounds}
    \item \hyperref[sec:proof-main]{\textbf{Section~E.}} \textbf{Proofs of Main Text}
    \item \hyperref[app:exp-detail]{\textbf{Section~F.}} \textbf{Implementation Details}
    \item \hyperref[sec:additional-exp]{\textbf{Section~G.}} \textbf{Additional Experiments}
\end{itemize}

\section{Table of Notations}

\begin{table}[h]\label{sec:tab_notation}

    \caption{Mathematical Notations and Symbols}
    \centering
    \resizebox{ \textwidth}{!}{ %
    \begin{tabular}{cl}
    \toprule
        Symbol & Description \\
    \midrule
        $\ba[i]$ & The $i$-th component of vector $\ba$ \\
        $\Braket{\ba,\bb}$ & Inner product for vectors $\ba,\bb\in \R^d$ \\
        $[I]$ & Index set $\{1,\cdots,I\}$, where $I\in\mathbb{N}^+$ \\
        $\norm{\cdot}$ & Spectral norm, equivalent to the $l_2$-norm when applied to a vector \\
    \midrule
        $d$ & Dimension of patterns \\
        $M$ & Number of stored memory patterns \\
        $\beta$ & Scaling factor of the energy function controlling the learning dynamics. 
        We set $\beta=1/\sqrt{d}$ in practice
        \\
    \midrule
        $\bx$ & State/configuration/query pattern in $\R^d$ \\
        $\bx^\star$ & Stationary points of the Hopfield energy function \\
        $\bxi$ & Memory patterns (keys) in $\R^d$ \\
        $\delta$ & Noises in memory patterns in $\R^d$\\
        $\bm{\Xi}$ & Shorthand for $M$ stored memory (key) patterns $\{\bxi_\mu\}_{\mu\in[M]}$ in $\R^{d\times M}$ \\
        $\bm{\Xi}^\sT \bx$ & $M$-dimensional overlap vector  $\(\Braket{\bxi_1,\bx},\cdots,\Braket{\bxi_\mu,\bx},\cdots,\Braket{\bxi_M,\bx}\)$ in $\R^{M}$ \\
    \midrule
        $\Phi(\cdot)$ & Kernelized feature mapping $\Phi(\cdot):\R^d\to D_\Phi$\\
        $D_\Phi$ & Dimension of the kernel space, i.e., dimension of output of $\Phi(\cdot)$ \\
        $\bW$ & Weighted matrix of the linear affine feature map defined in \eqref{eqn:linear_kernel} in $\R^{d\times D_f}$ \\
        $\calK(\cdot,\cdot)$ & Kernel function takes the inner product form $\calK(\cdot,\cdot)=\Braket{\Phi(\cdot),\Phi(\cdot)}$: $\R^{d}\times\R^{d}\to\R_+$ \\
    \midrule
        $\calM$ & Reduced support set for $\calT_{\text{SVR}}$ $\calM\coloneqq\{\calM(1),\ldots,\calM(k)\}\subseteq\{1,\ldots,M\}$ \\
        $\mathds{1}_{\calM(\mu)}$ & Indicator function corresponding to $\calM$, where $\mathds{1}_{\calM(\mu)}=1$ for $\mu\in \calM$ and $\mathds{1}_{\calM(\mu)}=0$ for $\mu\not\in \calM$ \\
        $k$ & Size of the support set $\calM$, defined as $k\coloneqq\abs{\calM}$ \\
    \midrule
        $n$ & Norm of $\bx$, denoted as $n\coloneqq\norm{\bx}$ \\
        $m$ & Largest norm of memory patterns, denoted as $m\coloneqq \Max_{\mu\in[M]}\norm{\bxi_\mu}$ \\
    \midrule
        $R$ & Minimal Euclidean distance across all possible pairs of memory patterns, denoted as $R\coloneqq \half \Min_{\mu,\nu\in[M]}\norm{\bxi_\mu-\bxi_\nu}$ \\
        $\calS_\mu$ & Sphere centered at memory pattern $\bxi_\mu$ with finite radius $R$ \\
        $\bx^\star_\mu$ & Fixed point of $\calT$ covered by $S_\mu$, i.e., $\bx_\mu^\star \in S_\mu$ \\
    \bottomrule
    \end{tabular}
    }
    \label{tab:nomenclature}
\end{table}

\clearpage
\section{Related Work}
\label{sec:related_work}

\paragraph{Hopfield Networks.}
Associative memory models \cite{willshaw1969non, kanerva1988sparse} have been widely discussed in both the neuroscience and machine learning fields.
The main goal of these models are to store a set of memory patterns where those patterns can be retrieved with respect to a given query.
Hopfield models represent a primary category within the class of computational associative memory models \cite{hopfield1982neural}.
Starting from the classical Hopfield models \cite{hopfield1982neural, hopfield1984neurons, krotov2020large}, these models are able to store and retrieve binary patterns with guaranteed memorization capacity.
Their biologically plausible designs provides significant insights to understand both human brains \cite{yampolskaya2023controlling,krotov2020large} and modern deep learning paradigms \cite{burns2024semantically,cabannes2024learning,cabannes2023scaling,kozachkov2023building, negri2023storage,ramsauer2020hopfield}.
Recently, these Hopfield models regain interest in the deep learning field due to its connection to the attention mechanism in transformers.
Notably, \citet{ramsauer2020hopfield} propose the Modern Hopfield models (MHMs) whose single-step update is equivalent to the attention mechanism \cite{vaswani2017attention}.
As a result, this connection (starting from the dense associative memory model \cite{krotov2016dense}) facilitates the integration of associative memory models into modern deep learning \cite{hofmann2024energy,hu2024nonparametric,xu2024bishop,wu2023stanhop,burns2023simplicial,auer2023conformal,widrich2020modern} and large foundation models \cite{hu2024outlier,pan2024conv,furst2022cloob}.

\paragraph{Theory of Modern Hopfield Models.} 
Beside empirical success, Modern Hopfield Models (MHM) offer a low-assumption theoretical framework for analyzing transformer-based deep learning architectures. 
Toward their fundamental theory, \citet{hu2023SparseHopfield} and \citet{wu2023stanhop} point out that the energy function of MHM and its sparse variants are actually tied to the convex conjugates of different entropic regularizers. 
This has led to the Sparse and Generalized Sparse HMHs, which are connected to attention mechanisms with various degrees of sparsity \cite{correia2019adaptively, vaswani2017attention, martins2016softmax}. 
Extending this foundation, \citet{hu2024nonparametric} further complement this understanding with the principled construction of possible efficient variants from a nonparametric perspective.
Furthermore, \citet{hu2024computational} provide a detailed theoretical analysis of all possible efficient variants, through the lens of fine-grained complexity theory.

We would like to comment further on the results of \cite{hu2024computational}. 
First, it observes that the magnitude of the patterns (i.e., the norms of queries and memories) not only affects retrieval accuracy (as seen in the linear 
$m$ scaling in \eqref{eqn:error_bound_exp}), but also determines the efficiency of a variant of the modern Hopfield model. 
This norm-based efficiency criterion, with precision guarantees, echoes the \textit{outlier effect} in the attention heads of transformer models \cite{hu2024outlier}. 
This outlier effect is well-known in pretraining large transformer-based models for its negative impact on model quantization performance \cite{sun2024massive, bondarenko2023quantizable, bondarenko2021understanding}. To address this, \citet{hu2024outlier} interpret the outlier effect as inefficient \textit{rare} memory retrieval and propose the outlier-efficient Hopfield layer for transformer-based large models, demonstrating strong empirical performance and theoretical guarantees. 
The benefits of removing outliers in the attention heads of transformer-based large foundation models are also highlighted in \cite{gu2024conv,gu2024tensor, alman2024fine, as23_tensor,alman2023fast, gao2023fast} from various theoretical perspectives. 
In this work, the removal of outliers is achieved by the row-wise normalization in {\uhop} (see line 4 of \cref{algorithm1}).

\begin{table*}[t]
\centering
\caption{Comparison between uniform memory Hopfield and other existing works.}
\resizebox{ \textwidth}{!}{ %
\begin{tabular}{lccc}
\toprule
   Model & Overlap Construction & Separation & Adaptivity \\
   \midrule
   Dense Associative Memory \cite{krotov2016dense} & Dot Product & Polynomial & No \\
   Modern Hopfield Nework \cite{ramsauer2020hopfield} & Dot Product & Softmax & No \\
   Sparse Modern Hopfield Network \cite{hu2023SparseHopfield} & Dot Product & Sparsemax & No \\
   $\mathtt{U\text{-}Hop}$ + \cite{wu2023stanhop,hu2023SparseHopfield,ramsauer2020hopfield} & Kernel Function & Not Restricted & Yes \\
   \bottomrule
\end{tabular}
}\label{table:1}
\end{table*}

\paragraph{Learning Associative Memory Models.}
Another line of research focuses on learning an associative memory model \cite{tyulmankov2021biological, salvatori2021associative} that has the ability to ``read" 
 (retrieve) and ``write" (store) memories.
Particularly, this type of method contains a "readout" network to retrieve/generate memories with a given query.
\citet{bartunov2019meta} propose a meta learning framework to learn a generative network that treats the retrieval error as their energy function.
\citet{yoo2022bayespcn} propose a hierarchical associative memory model that relaxes the requirement of meta learning.
\citet{salvatori2021associative} propose a hierarchical generative network trained with predictive coding.
Instead of deriving the retrieval dynamics from the energy function, these methods normally use a generative model for memory retrieval.
With the expressiveness of deep neural networks, such method showed great empirical performances.
However, since the structure of the readout network does not connect or dependent on the energy function, they are not able to preserve appealing theoretical guarantees like Hopfield models.

\paragraph{Kernel Memory Networks.}
\citet{iatropoulos2022kernel}  propose a kernelized memory network\footnote{In a similar vein, \citet{schaeffer2024bridging} bridge associative memory models and probabilistic modeling.
}.
They formulate the modern Hopfield models with a recurrent SVM model.
In particular, their kernel is a single layer feed forward network that is trained to memorize patterns.
However, their framework consists of several high assumptions. 
In comparison, our proposed framework has mild assumption on parameters and pattern distributions.
In addition, {\uhop} has significant practical usage and was validated through extensive experiments in both memory retrieval and supervised learning tasks.

This work bridges two paradigms of associative memory models via a non-singular kernel, such that the kernelized energy function \eqref{eqn:eng-fn} still satisfies the defining properties of modern Hopfield models, i.e. attention-included retrieval dynamics (\cref{lemma:retrieval_dyn}). 
A comparison between Uniform Memory Hopfield and similar models are shown in \cref{table:1}.

\paragraph{Kernel Approach in Transformer Attention.}

The usage of kernels and feature expansions in transformers has been extensively discussed in previous literature. 
One primary objective of these studies is to reduce the computational complexity associated with attention mechanisms. 
For instance, \citet{chen2021skyformer, kitaev2020reformer, chen2021scatterbrain} demonstrate empirically and theoretically that these efficient algorithms can effectively approximate SoftMax attention. 
\citet{song2021implicit} provides a generalized framework for attention mechanism by decomposing it into two parts, RBF kernel as similarity measure and $L_2$ norm weighting on tokens.
In our paper, we offer a distinct perspective aimed at enhancing memory capacity, drawing inspiration from the construction of the modern Hopfield model. 
Therefore, our approach differs from attempting to approximate the standard modern Hopfield association. Instead, we focus on relocating memory patterns to facilitate easier retrieval.

\clearpage
\section{Connection to Transformer Attentions}\label{sec:connect-attention}
Suppose that $\bX$ and $\bm{\Xi}$ are embedded from the \textit{raw} query ${\bR}$ and $\bY$ memory patterns, respectively, via $\bX^\sT={\bR}\bW_Q\coloneqq \bQ$,
and $\bm{\Xi}^\sT=\bY \bW_K \coloneqq \bK$,
with some projection matrices $\bW_Q$ and $\bW_K$.
Then, taking the transport of $\calT$ in \eqref{eqn:retrival_dyn} and multiplying with $\bW_V$ such that $\bV\coloneqq \bK\bW_V$, we obtain
\begin{align*}
    \bZ\coloneqq \bQ^{\text{new}} \bW_V
    =\Softmax\(\beta  \bQ\bK^\sT\)\bV.
\end{align*} 
This result enables that the modern Hopfield models are able to serve as powerful alternatives to the attention mechanism equipped with additional functionalities.

\begin{claim}
This connection provides a insightful theoretical foundation for the attention mechanism. 
Specifically, the update step in a Transformer's attention mechanism functions as an inner-loop optimization, minimizing an underlying energy function defined by the queries, keys, and values.
Please see \cite{ramsauer2020hopfield,hu2023SparseHopfield,wu2023stanhop}.
\end{claim}

\clearpage
\section{Supplementary Theoretical Backgrounds}\label{sec:sup-theory}

\subsection{Sparsemax and $\alpha$-EntMax}\label{app:entmax}
Here we quote some known results from \cite{hu2023SparseHopfield,martins2016softmax}.

Let $\bz,\bp\in\R^M$, and $\Delta^{M}\coloneqq\{\bp\in\R^M_+ \mid \sum_\mu^M p_\mu=1\}$ be the $(M-1)$-dimensional unit simplex.
 where $\alpha\text{-EntMax}(\cdot):\R^M\to \Delta^M$ is a finite-domain distribution map defined as follows.
\begin{definition}[$\alpha$-EntMax] The variational form of $\alpha\text{-EntMax}$ is defined by the optimization problem
\begin{align}
  \label{eqn:entmax}
\alpha\text{-EntMax}(\bz) \coloneqq \argmax_{\bp \in \Delta^M} [\Braket{\bp,\bz}-\Psi_\alpha(\bp)],  
\end{align}
where $\Psi_\alpha(\cdot)$ is the Tsallis entropic regularizer given by \eqref{eqn:T_entropy}.
\begin{align}
\label{eqn:T_entropy}
\Psi_\alpha(\bp)\coloneqq
\begin{cases}
    \frac{1}{\alpha(\alpha-1)}\sum^M_{\mu=1}\(p_\mu-p_\mu^\alpha\), \quad & \alpha\neq 1,\\
    -\sumM p_\mu\ln p_\mu,\quad &\alpha=1,
\end{cases},\quad\text{for }\alpha\ge 1,
\end{align}
\end{definition}

Let $\bz\in\R^M$.
Denote $[a]_+\coloneqq \Max\{0,a\}$, $z_{(\nu)}$ the $\nu$'th element in a sorted descending $z$-sequence $\bz_{\text{sorted}}\coloneqq z_{(1)}\ge z_{(2)}\ge\ldots\ge z_{(M)}$, and 
$
\kappa(\bz)\coloneqq\Max\big\{k\in [M]\;\big\vert \; 1+kz_{(k)}>\sum_{\nu\le k}z_{(\nu)}\big\}.
$
Sparsemax$\( \cdot \)$ is defined as (Proposition~1 of \cite{martins2016softmax}) :
\begin{align}
  \label{eqn:sparsemax_exact}
\text{Sparsemax}(\bz)=\[\bz-\tau(\bz)\mathbf{1}
_M\]_+,  
\end{align}
where $\tau:\R^M \to \R$ is the threshold function 
$\tau(\bz)=\[\(\sum_{\nu\le \kappa(\bz)}z_{(\nu)}\)-1\]/\kappa(\bz)
$,
satisfying $\sumM \[z_\mu-\tau(\bz)\]_+=1$ for all $\bz$.
Notably, $\kappa(\bz)=\abs{S(\bz)}$ where $S(\bz)=\{\mu\in[M]\;|\;\text{Sparsemax}\mu(\bz)>0\}$ is the support set of $\text{Sparsemax}(\bz)$.

\subsection{Separation Loss}

With the output vector of $\Phi$ is normalized, we have
\begin{align}
\label{eqn:L_normalize}
    \ell_{\Phi} \( \bu, \bv\) &= 2t \cdot \langle \Phi(\bu), \Phi(\bv) \rangle - 2t \nonumber\\
    &= -t \cdot \| \Phi(\bu) - \Phi(\bv) \|^2.
\end{align}

Given the query $\bx = \bxi_\mu + \br$, with $\br \in \R^d$, the uniformity loss satisfies the following:
\begin{align*}
    \ell_\Phi \( \bx, \bxi_\mu \)
    =
     \ell_\Phi \( \bxi_\nu, \bxi_\mu \) + \ell_\Phi \( \br, \bxi_\mu \) + 2t.
\end{align*}

\subsection{Convergence Rate of Gradient Descent}

\begin{lemma}[\jhcomment{citation}]\label{lem:converge-gd}
    Suppose a function $f \coloneqq \R^d \rightarrow \R$ is convex and differentiable, and that its gradient is Lipschitz continuous with constant $G > 0$, i.e. $\norm{ \nabla f(x) - \nabla f(y) }_2 \leq G \norm{ x-y }$ for any $x, y \in \R^d$. Then if we run gradient descent for $k$ iterations with a fixed step size $t \leq 1/G$, it returns a solution which satisfies 
    \begin{align*}
        f(x^{k}) - f(x^\star) \leq \frac{ \norm{ x^0 - x^\star}^2_2 }{2tk},
    \end{align*}
    where $f(x^\star)$ is the optimal value of $f$. Intuitively, this means that gradient descent is guaranteed to converge and that it converges with rate $\calO(1/k)$.
\end{lemma}

\begin{remark}
    From \cref{lem:converge-gd}, to achieve a bound of:
    \begin{align*}
        f(x^k) - f(x^\star) \leq \epsilon,
    \end{align*}
    we must run $k = \calO(1/\epsilon)$ iterations of gradient descent, which gives us a sub-linear 
\end{remark}

\clearpage
\section{Proofs of Main Text}\label{sec:proof-main}

\subsection{Proof of \cref{lemma:convergence}}\label{proof:ret-conv}

\begin{proof}

Here we introduce a helper lemma from \citep[Lemma~5]{sriperumbudur2009convergence}.
\begin{lemma}[Lemma~5 of \cite{sriperumbudur2009convergence}]\label{lem:help-conv}
    Following \cref{lemma:retrieval_dyn}, $\bx$ is called the fixed point of iteration $\calT$ w.r.t. $E$ if $\bx = \calT(\bx)$ and is considered as a generalized fixed point of $\calT$ if $\bx \in \calT(\bx)$.
    If $\bx^\star$ is a generalized fixed point of $\calT$, then, $\bx^\star$ is a stationary point of the energy minimization problem in \cref{eqn:eng-fn}.
\end{lemma}

Based on Zangwill's global convergence theory \cite{zangwill1969nonlinear}, a set of limit points of $\{ x_t \}^\infty_{t=0}$ are all generalized fixed points if the retrieval dynamics and energy function satisfies the following conditions:
\begin{enumerate}
    \item For any sequence $\{x_t\}^\infty_{t=0}$ with $x_0 \in S_\mu$ as starting point, all points in $\{x_t\}^\infty_{t=0}$ are in the compact set $S_\mu$.
    \item $E(\bx)$ is monotonically decreased by $\calT(\bx)$, where $E(\bx_{t+1}) \leq E(\bx_t), \forall \bx_{t+1} = \calT(\bx_t)$.
    \item For all $t$, if $E(\bx_{t+1}) \leq E(\bx_t)$, $\calT$ is closed at $\bx_t$.
\end{enumerate}

From \cref{def:stored_and_retrieved}, since radius $R$ is bounded and closed, $S_\mu$ is a compact set. Thus satisfies the first condition.
CCCP \cite{yuille2001concave} studied the monotonic decreasing property.
With our definition of $E_{\text{convex}}, E_{\text{concave}}$, we have
$E_U( \bx, \by ) = E_{\text{convex}}(\bx) + E_{\text{concave}}(\by) + \braket{(\bx - \by), \nabla_\bx E_{\text{concave}}(\by) }$ is continuous in $\bx, \by$.
As a result, by \citep[Lemma~E.1]{hu2023SparseHopfield}, condition (iii) holds due to the non-empty assumption on the point-to-set map $\calT$.
Thus, by Zangwill's global convergence theory, all limit points are also the stationary points of the energy minimization problem in \eqref{eqn:eng-fn}.
By the results in \cref{lem:help-conv}, these fixed points are also the stationary points of the minimization problem.
Thus, \eqref{eqn:eng-fn} is guaranteed to converge to local minimum.
\end{proof}

\subsection{Proof of \cref{lemma:retrieval_dyn}}\label{proof:retrieval_dynamics}

\begin{proof}

Since the $\lse$ function is non-decreasing and convex, and $\calK$ is convex, the composited function $\lse \( \calK(\bm{\Xi}, \bx) \)$ is convex.
Thus, the energy function is the sum of a convex function: $   \Braket{\bW \bx, \bW \bx} /2$ and a concave function: $- \lse\( \calK(\bm{\Xi}, \bx) \)$.

Therefore, we have $E(\bx) = E_{\text{convex}}(\bx) + E_{\text{concave}}(\bx)$.
With the Concave-Convex Procedure (CCCP) \cite{yuille2001concave}, the energy function $E(\bx)$ is guaranteed to monotonically decrease the energy $E$ as a function of time with the following update rule:
\begin{align*}
    \underbrace{\nabla_\bx E_{ \text{convex} }(\bx^{t+1})}_{=\bA \bx^{t+1}} = \underbrace{- \nabla_\bx E_{\text{concave}}(\bx^\top) }_{=\nabla_x \int \calT(\bx) \dd \bx},
    \end{align*}
    such that
    \begin{align*}
    \bA \bx^{t+1} = \underbrace{\bA}_{d \times d} \cdot \underbrace{\bm{\Xi} \cdot \text{Sep} \( \calK \(\bm{\Xi}, \bx \) \)}_{d\times 1} + \bc_1,
\end{align*}
where $\bc_1\in\R^{d}$ is a constant vector.

Since the matrix $\bA$ is non-singular by \cref{asum:non-sing}, the solution of the update rule $\bA \bx^{t+1} = \bA \bm{\Xi} \cdot \text{Sep} \(\calK(\bm{\Xi}, \bx)\)\in\R^{d}$ is the solution of $\bx^{t+1}=\bm{\Xi} \cdot \text{Sep} \(\calK ( 
\bm{\Xi}, \bx)\)\in\R^d$ minimizes the energy function $E(\bx)$.
This completes the proof.
\end{proof}

\clearpage

\subsection{Proofs of \cref{thm:exact} and \cref{coro:exact_coro}}\label{proof:optimal}
\begin{proof}
Here we use the same proof strategy in \citep[Proposition 2]{martins2023sparse}.

     The Fenchel-Young loss \cite{scholkopf2018learning} indicates that if a vector $\bm{\theta} \in \mathbb{R}^d$ for any $d > 1$ satisfies
    \begin{align*}
       \alpha\text{-EntMax} ( \bm{\theta} ) = \bm{e}_\mu,
    \end{align*}
    then $\bm{\theta}$ must satisfy
    \begin{align}\label{eqn:fy-loss-onehot}
        \theta_\mu - \underset{\nu \neq \mu}{\max}\; \theta_\nu \geq \frac{1}{\alpha - 1}.
    \end{align}

    If we have exact memory retrieval of pattern $\bxi_\mu$, the following equation holds:
    \begin{align}\label{eqn:update-onehot}
        \be_\mu = \alpha\text{-EntMax}
        \(
        \beta
        \calK(\bm{\Xi}, \bxi_\mu)
        \).
    \end{align}
    This is also equivalent to $\bxi_\mu$ itself being the fixed point.

    By combining \eqref{eqn:fy-loss-onehot} and \eqref{eqn:update-onehot}, we have
    \begin{align*}
        \calK( \bxi_\mu, \bxi_\mu ) - \underset{\nu \neq \mu}{\max}  \calK ( \bxi_\nu, \bxi_\mu ) &\geq \frac{1}{\beta  \(\alpha-1\)}. 
    \end{align*}

    With $\calK( \bxi_\mu, \bx ) = \frac{1}{2t} \cdot \ell_\Phi \(\bxi_\mu, \bx\)$, we have
    \begin{align*}
     \ell_\Phi \(\bxi_\mu, \bxi_\mu \) -  \underset{\nu \neq \mu}{\max}\ell_\Phi \(\bxi_\nu, \bxi_\mu \) & \geq \frac{2t}{\beta  \(\alpha-1\)}.
    \annot{By \cref{thm:exact}}
    \end{align*}

    With the assumption of normalized patterns, we are able to reduce the above condition to
    \begin{align}\label{eqn:exact-condition-1}
        \underset{\nu \neq \mu}{\max}\ell_\Phi \(\bxi_\nu, \bxi_\mu \) & \leq \frac{2t}{\beta  \(\alpha-1\)}.
    \end{align}

    Assuming $\Phi$ is $L$-lipschitz\footnote{$\Phi$ is always $L$-lipschitz for some $L \in \mathbb{N}$ in our construction since a linear affine function always has $L$-Lipschitzness.}, we derive another upper bound from \eqref{eqn:exact-condition-1}:
    \begin{align}\label{eqn:lipschitz-kernel}
         \underset{\nu \neq \mu}{\max}\ell_\Phi\(\bxi_\nu, \bxi_\mu \)  
         & =
         \underset{\nu \neq \mu}{\max} -t \cdot \norm{ \Phi(\bxi_\nu) - \Phi(\bxi_\mu) }^2 \nonumber\\
         & \leq 
         \underset{\nu \neq \mu}{\max}
         -t\cdot  L^2 \norm{\bxi_\nu - \bxi_\mu}^2
         .
    \end{align}
    By combining \eqref{eqn:lipschitz-kernel} and \eqref{eqn:exact-condition-1}, we obtain
    \begin{align*}
         \underset{\nu \neq \mu}{\min} \norm{\bxi_\nu - \bxi_\mu}^2
         \geq \frac{2}{\beta  \(\alpha-1\) \cdot L^2}.
         \annot{By \cref{coro:exact_coro}}
    \end{align*}
This completes the proof.
\end{proof}

\clearpage
\subsection{Proof of \cref{thm:perfect_kernel}}\label{proof:representation}

\setcounter{theorem}{0} %
\renewcommand{\thetheorem}{3.1} %
\begin{theorem}[Kernelized Representation Theorem]
Let $\bar{\Phi}$ be a feature map such that $\bar{\Phi} \coloneqq \R^d \rightarrow \R^{D_{\bar{\Phi}}}$, and $\bar{\calK}$ be a $\bar{\Phi}$-induced kernel.
    Assuming $\bar{\calK}$ satisfies: $\bar{\ell}_{\bar{\Phi}}(\bu, \bv) = -2t$ for any given $\bu, \bv \in \bm{\Xi}$.
    With $\beta > 0$, input $\bX \in \R^{d \times M}$, $M \leq d$, and an arbitrary positive column stochastic matrix $\bP \in \R^{M \times M}$, there always exist matrices $\bW_Q, \bW_K$ such that 
    \begin{align*}
        \text{Softmax} \left( \beta \;  
        \(
        \bW_K \bar{\Phi}(\bX)
        \)^\sT
        \bW_Q \bar{\Phi}(\bX) \right) = \bP.
    \end{align*}
\end{theorem}

\begin{proof}
    Let $\bX^\prime = \bar{\Phi}(\bX) \in \R^{d \times M}$.

    By construction, any two columns in $\bX^\prime$ satisfies:
    \begin{align*}
    \Braket{ \bar{\Phi}(\bu), \bar{\Phi}(\bv)}
    = 0,
    \annot{By $\bar{\ell}_{\bar{\Phi}}(\bu, \bv) = -2t$}
    \end{align*}
    for all $\bu, \bv \in \bX^\prime$, $\bu\neq\bv$.
    Thus, any two columns in $\bX^\prime$ are orthogonal to each other, and hence $\bX^\prime$ has left inverse $\bX^\dagger \in \R^{M \times d}$.
    
    Let $\bW_K \coloneqq \Tilde{\bW}_K \bX^\dagger$, and  $\bW_Q\coloneqq \Tilde{\bW}_Q \bX^\dagger$ for some $\Tilde{\bW}_K,\Tilde{\bW}_Q\in\R^{d\times M}$.

This gives us
    \begin{align*}
        \(\bW_K \bX^\prime\)^\sT \bW_Q \bX^\prime
        =
        \(\tilde{\bW}_K \bX^\dagger \bX^\prime\)^\sT \Tilde{\bW}_Q \bX^\dagger \bX^\prime 
        =
        \Tilde{\bW}_K^\sT \Tilde{\bW}_Q = \tilde{\bW}_{KQ}.
        \annot{$\tilde{\bW}_{KQ} \in \R^{M \times M}$}
    \end{align*}
    
    With softmax, we have
    \begin{align}\label{eqn:softmax-representation}
        \Softmax \( \beta \; \(\bW_K \bX^\prime\)^\sT \bW_Q \bX^\prime \)
        &=
        \Softmax \(
\beta \; \Tilde{\bW}_{KQ}    \) \nonumber\\
&=
\exp{\beta \Tilde{\bW}_{KQ} } \cdot \bD^{-1}_{\tilde{\bW}_{KQ}},
    \end{align}
where $\bD^{-1}_{\tilde{\bW}_{KQ}} \in \R^{M \times M}$ is a diagonal matrix which
\begin{align}\label{eqn:construction1}
    \(\bD_{\Tilde{\bW}_{KQ}} \)_{ii} = \sum_{j=1}^M \exp{\beta \(\Tilde{\bW}_{KQ}\)_{ji}} =
    \( \mathbf{1}^\top \) \sum_{j=1}^M \exp{\beta \(\Tilde{\bW}_{KQ}\)}.
\end{align}
Now with $\bP$, we are able to construct $\Tilde{\bW}_{KQ}$ by picking an arbitrary positive diagonal matrix such that
    \begin{align}\label{eqn:construction2}
        \Tilde{\bW}_{KQ} = \beta^{-1} \cdot \log \( \bP \cdot \bD_0 \).
    \end{align}

    By combining \eqref{eqn:construction1} and \eqref{eqn:construction2}, we have
    \begin{align*}
        \bD_{\tilde{\bW}_{KQ}} = 
        \text{Diag}  \( \mathbf{1}^\top \exp{ \beta \beta^{-1} \cdot \log \( \bP \cdot \bD_0 \) } \) =
        \text{Diag} \( \mathbf{1}^\top \bP \cdot \bD_0 \)
        = \bD_0.
    \end{align*}

    With $\bD_{\tilde{\bW}_{KQ}} = \bD_0$, and using \eqref{eqn:softmax-representation}, we obtain
    \begin{align*}
        \Softmax \(\beta \; \Tilde{\bW}_{KQ} \) = \exp{\beta \Tilde{\bW}_{KQ} } \cdot \bD^{-1}_{\tilde{\bW}_{KQ}} = 
        \exp{ \log \(\bP \cdot \bD_{\tilde{\bW}_{KQ}} \)}
        \cdot \bD^{-1}_{\tilde{\bW}_{KQ}} = \bP.
    \end{align*}
    As a result, to construct $\bW_K$ and $\bW_Q$ such that they satisfy \cref{thm:perfect_kernel}, any two matrices must satisfy
    \begin{align*}
    \bW_K^\sT \bW_Q = \beta^{-1} \cdot \log( \bP \cdot \bD_{\tilde{\bW}_{KQ}} )
    \end{align*}
This completes the proof.
\end{proof}

\clearpage
\section{Implementation Details}\label{app:exp-detail}

\subsection{Data}

\begin{itemize}
    \item 
    \textbf{MNIST.}
It is a hand written digits image recognition dataset \cite{lecun1998gradient} consists of 60000 training samples and 10000 test samples.
Each image has the size of $28\times28$.
The label contains digits from 0 to 9.
    \item 
    \textbf{CIFAR10.}
It is an image recognition dataset \cite{krizhevsky2009learning} consists of 50000 training samples and 10000 test samples.
Each image has the size of $32\times32$.
The dataset contains 10 categories with 6000 samples for each.

    \item 
\textbf{CIFAR100.}
It is an image recognition dataset \cite{krizhevsky2009learning} consists of 50000 training samples and 10000 test samples.
Rach image has the size of $32\times32$.
The dataset contains 100 categories with 600 samples for each.

    \item 
\textbf{TinyImageNet.}
It is an image recognition dataset \cite{le2015tiny} contains 100000 images of 200 classes.
Each image is downsized to 64×64 colored images. 
Each class has 500 training images, 50 validation and test images.

    \item \textbf{ETT (Electricity Transformer Temperature).}
    ETT \cite{zhou2021informer} records 2 years of data from two counties in China. 
    We use two sub-datasets: ETTh1 (hourly) and ETTm1 (every 15 minutes). 
    Each entry includes the ``oil temperature" target and six power load features.

    \item 
\textbf{WTH (Weather).}
    WTH records climatological data from approximately 1,600 U.S. sites between 2010 and 2013, measured hourly. 
    Entries include the ``wet bulb" target and 11 climate features.

\end{itemize}

\subsection{Memory Capacity}
For memory capacity experiment, we follow the settings in \cite{hu2023SparseHopfield, wu2023stanhop}.

\begin{figure*}[h]
\begin{minipage}[h]{0.47\linewidth}
\begin{center}
\includegraphics[width=1\linewidth]{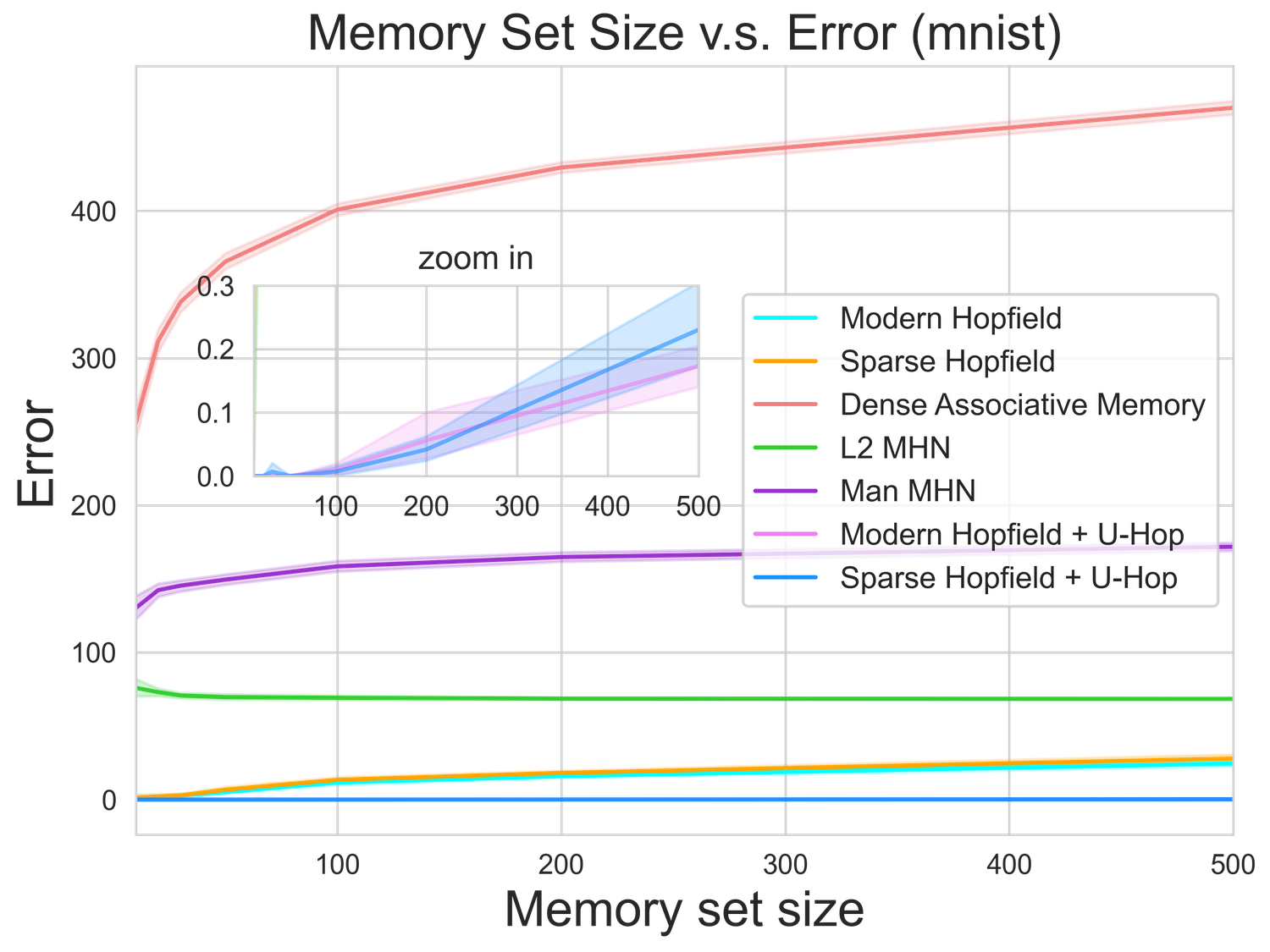} 
\end{center} 
\end{minipage}
\hfill
\begin{minipage}[h]{0.47\linewidth}
\begin{center}
\includegraphics[width=1\linewidth]{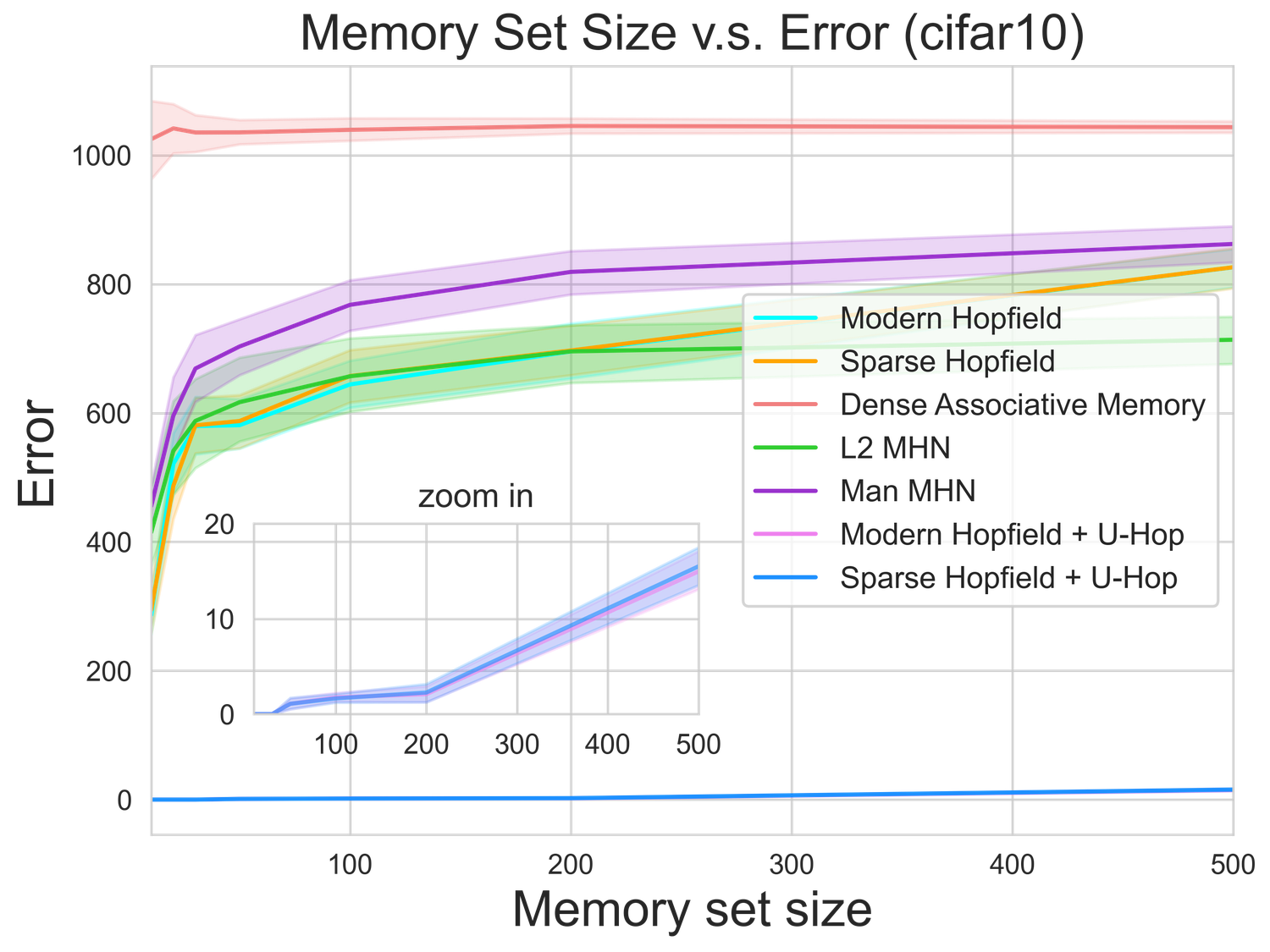} 
\end{center}
\end{minipage}
\caption{\textbf{Memory Retrieval Error v.s. Memory Set Size ($M$).} \textbf{Left: MNIST, Right: CIFAR10}.
We conduct memory retrieval experiment on MNIST and CIFAR10 dataset.
We vary the memory size to adjust the difficulty of retrieval process.
}
\label{fig:mem-ret-big}
\end{figure*}

We randomly mask $50\%$ of the pixels in the image, using the masked image as a query for a single-step update with various Hopfield models. 
In the case of {\uhop}, we trained the kernel with different numbers of epochs on the memory set and then used it for memory retrieval. 
We reported the sum-of-square pixel difference between the retrieved image and the ground truth. 
In each run, we repeated this process for every image in the memory set, conducting the experiment 20 times for each baseline.
The range of kernel learning epochs, memory set size can be found in \cref{table:hyper-mem-ret}.
For \cref{fig:mem-ret-big}, we use $N=100$ for MNIST and $N=500$ for CIFAR10.

\begin{table}[h]
        \centering
        \caption{Hyperparameter used in the Memory Retrieval Task.
        }
        \vspace*{0.05truein} 
        \begin{tabular}{l*{1}{c}}
        \toprule
            \bf{parameter} & \multicolumn{1}{c}{\bf{values}}\\ 
            \midrule
            Kernel optimizer & SGD \\
            Kernel learning rate & 1 \\
            Kernel epoch  & $\{ 1, 10, 20, 50, 100, 200, 500, 1000 \}$ \\
            Memory set size & $\{ 10, 20, 30, 50, 100, 200, 500 \}$ \\
            Noise level & $\{ 0.0, 0.01, 0.05, 0.1, 0.3, 0.5, 0.7, 1.0, 1.2, 1.4, 2.0 \}$ \\
            \bottomrule
        \end{tabular}
        \label{table:hyper-mem-ret}
    \end{table} 

\subsection{Noise Robustness}
For  noise robustness experiment, we follow the settings in \cite{hu2023SparseHopfield, wu2023stanhop}.

For the noise robustness experiment, we randomly sampled a Gaussian noise vector for each image, varying the norm of the sampled noise to adjust the noise level. We then added the noise to the query image and performed a single-step update with different Hopfield models. 
For {\uhop}, we trained the kernel with N iterations on the memory set and then used it for memory retrieval. 
We set $N = 100$ for MNIST and $N = 200$ for CIFAR10. 
We reported the sum-of-square pixel difference between the retrieved image and the ground truth. 
In each run, we repeated this process for every image in the memory set, conducting the experiment 20 times for each baseline.

\subsection{Classification}

\textbf{CIFAR10 and CIFAR100.} For these two datasets, we consider four different Hopfield layers as encoder:
\begin{itemize}
    \item $\mathtt{Hopfield}$ \cite{ramsauer2020hopfield}
    \item $\mathtt{SparseHopfield}$ \cite{hu2023SparseHopfield}
    \item $\mathtt{Hopfield}$ +{\uhop} 
    \item $\mathtt{SparseHopfield}$ + {\uhop}.
\end{itemize}

We use a fully connected layer right after the encoder for classification.
For each image, we follow the same process as introduced in \cite{dosovitskiy2020image}.
We split an image into patches and add an additional $\mathtt{CLS}$ patch for classification.
We send patches into the $\mathtt{Hopfield}$ layer with the $\mathtt{CLS}$ patch as query and other patches as memory.
We then send the output to a fully connected layer for prediction.
We use the CrossEntropy loss and Adam optimizer for training.
Hyperparameters are in \cref{table:hyper-cifar-cls}.

\begin{table}[h]
        \centering
        \caption{Hyperparameter used in the classifciation task on CIFAR10 and CIFAR100.
        }
        \vspace*{0.05truein} 
        \begin{tabular}{l*{1}{c}}
        \toprule
            \bf{parameter} & \multicolumn{1}{c}{\bf{values}}\\ 
            \midrule
            learning rate & $1e-3$\\
            embedding dimension & 512 \\
            Epoch & 25 \\
            Batch size & 128 \\
            Model optimizer & Adam \\
            Kernel optimizer & SGD \\
            Kernel learning rate & 0.1 \\
            Patch size & 32 \\
            \bottomrule
        \end{tabular}
        \label{table:hyper-cifar-cls}
    \end{table}

\begin{table}[h]
        \centering
        \caption{Hyperparameter used in the classifciation task on Tiny ImageNet.
        }
        \vspace*{0.05truein} 
        \begin{tabular}{l*{1}{c}}
        \toprule
            \bf{parameter} & \multicolumn{1}{c}{\bf{values}}\\ 
            \midrule
            learning rate & $1e-4$\\
            embedding dimension & 512 \\
            Epoch & 25 \\
            Batch size & 128 \\
            Model optimizer & Adam \\
            Kernel optimizer & SGD \\
            Kernel learning rate & 0.1 \\
            Patch size & 64 \\
            \bottomrule
        \end{tabular}
        \label{table:hyper-tiny-cls}
    \end{table}

\textbf{Tiny ImageNet.} For this dataset, we use a 3 layered Vision Transformer as backbone \cite{dosovitskiy2020image}, and use $\mathtt{Hopfield}$ variations to replace attention mechanism in .
Other processes are the same as introduced in the above paragraph.
For kernel learning, we learn all kernels in different layers by passing a full forward pass.
We then send the output to a fully connected layer for prediction.
Hyperparameters are in \cref{table:hyper-tiny-cls}.

\subsection{Hopfield-Based Time Series Prediction with STanHop-Net \cite{wu2023stanhop} (\cref{table:time-series})}

For this task, we use STanHop-Net \cite{wu2023stanhop} as backbone, and equip it with {\uhop}.
In addition, we use \cref{algo:bi-level} to minimize both separation loss and the MAE loss.
For prediction horizon of $96, 192$, we use one layered STanHop-Net, for $336, 720$, we use a two layered STanHop-Net.
We use the same settings for with and without {\uhop}
\footnote{We thank the authors of \cite{reneau2023feature} for their helpful comments on this part.}.

\begin{table}[h]
        \centering
        \caption{Hyper-parameter used in the time series prediction.
        }
        \vspace*{0.05truein} 
        \begin{tabular}{l*{1}{c}}
        \toprule
            \bf{parameter} & \multicolumn{1}{c}{\bf{values}}\\ 
            \midrule
            learning rate & $1e-4$\\
            embedding dimension & 256 \\
            Epoch & 50 \\
            Patience & 25 \\
            Batch size & 16 \\
            Model optimizer & Adam \\
            Kernel optimizer & SGD \\
            Kernel learning rate & 0.1 \\
            Patch Sequence Length & 12 \\
            Window Size & 2 \\
            \bottomrule
        \end{tabular}
        \label{table:hyper-time-series}
    \end{table}

\clearpage

\section{Additional Numerical Experiments}\label{sec:additional-exp}

\subsection{Relationship between Separation Loss and Retrieval Error}\label{sec:sep-vs-error-scatter}
This section is a visualization of relationship between separation loss and retrieval error on MNIST and CIFAR10.
The result shows that the retrieval error is highly correlated with respect to the value separation loss.

\begin{figure*}[h]
\begin{minipage}[h]{0.47\linewidth}
\begin{center}
\includegraphics[width=1\linewidth]{imgs/mnist_scatter.png} 
\end{center} 
\end{minipage}
\hfill
\begin{minipage}[h]{0.47\linewidth}
\begin{center}
\includegraphics[width=1\linewidth]{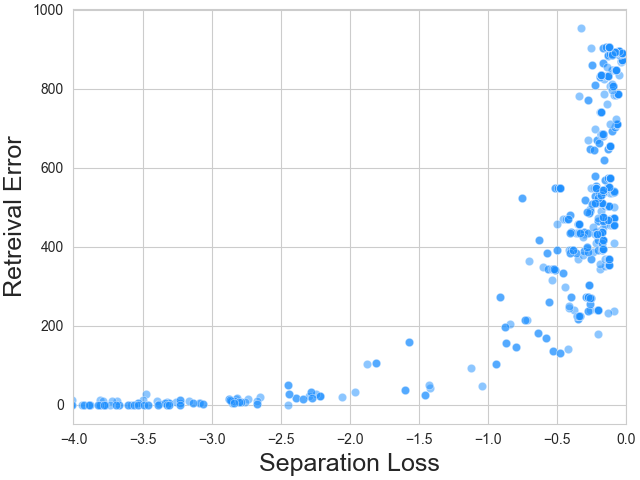} 
\end{center}
\end{minipage}
\caption{\textbf{Memory Retrieval Error v.s. Separation Loss:} \textbf{Left: MNIST, Right: CIFAR10}.
We conduct memory retrieval experiment on MNIST and CIFAR10 dataset.
We use randomly sampled kernel learning rate in $N(0, 1)$, and uniformly random sampled $N \in \[ 1, 200 \]$ to obtain diverse separation loss.
}
\label{fig:ret-error-sep-loss}
\end{figure*}

\subsection{Max. Loss v.s. Avg. Loss}\label{sec:max-vs-avg}
In the main paper, we discuss the differences between minimizing the maximum separation loss and the average separation loss.

Ideally, minimizing the maximum separation loss directly contributes to $R_\Phi$. However, as stated in the main text, such a loss is a max-min problem, which is challenging to optimize. Moreover, it entails an additional quadratic time complexity due to the max operation. On the other hand, the average loss is more time-efficient. It is also convex, thereby guaranteeing convergence to the global optimum at a rate of $\mathcal{O}(1/N)$ under gradient descent, where $N$ is the number of iterations. However, the average loss does not guarantee maximizing $R_\Phi$, nor does it ensure an optimal $\Delta_{\Phi, \mu}$ for any $\mu \in [M]$. 
Therefore, its theoretical impact on memory capacity and retrieval error bound  is difficult to quantify.

As a result, we conduct an analysis comparing the performance of each loss function. We vary the memory size and the kernel learning iteration $N$ and perform memory retrieval on MNIST and CIFAR10 datasets.

The results demonstrate that minimizing the average loss yields a lower retrieval error, and this advantage grows with the size of the memory set. Additionally, the retrieval error decreases more rapidly with respect to $N$ under average loss than under maximum loss. This outcome is anticipated, as minimizing the maximum loss is a non-convex problem and does not guarantee reaching global optima. This empirical finding indicates that in practice, minimizing the average loss leads to better efficiency and retrieval outcomes.

\begin{figure*}[h]\label{mnist-avg-max-epoch}
    \centering
    \includegraphics[width=\textwidth]{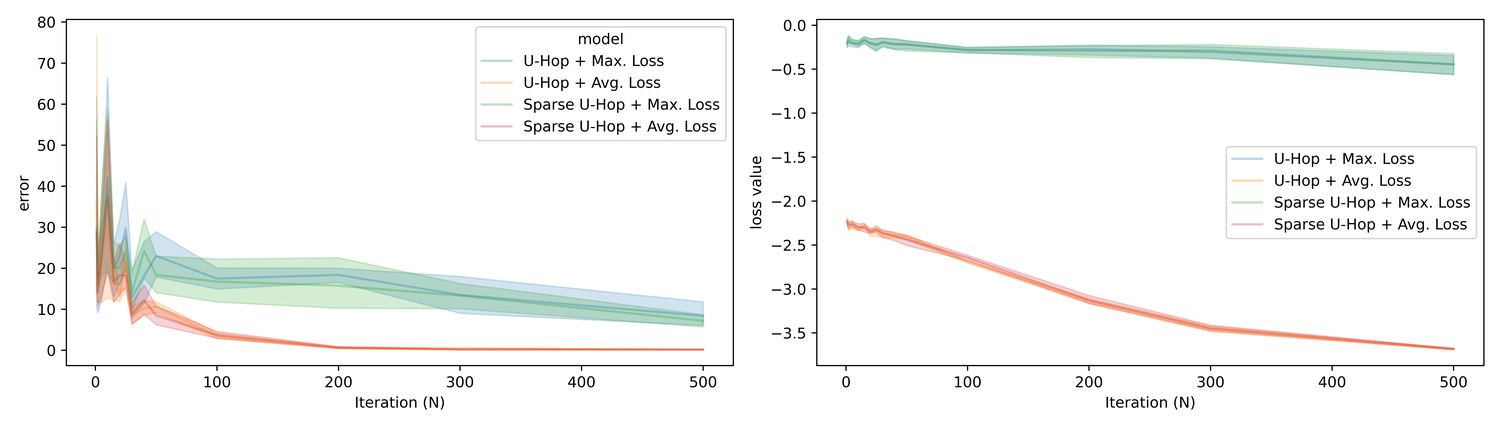}
    \caption{
    \textbf{Loss Value v.s. $N$ and Retrieval Error v.s. $N$.}
    }
\end{figure*}

\paragraph{Max. \&  Avg. Loss Comparison on MNIST.}
We observe that with the average loss, $\mathtt{U\text{-}Hop}$ achieves almost perfect retrieval outcomes with $N=200$.
However, $\mathtt{U\text{-}Hop}$ using the maximum loss struggles to reach its global minimum during optimization, thus hindering its ability to achieve optimal memory capacity. 
This is observed in \cref{mnist-avg-max-epoch}.

\begin{figure*}[h]
    \centering
    \includegraphics[width=\textwidth]{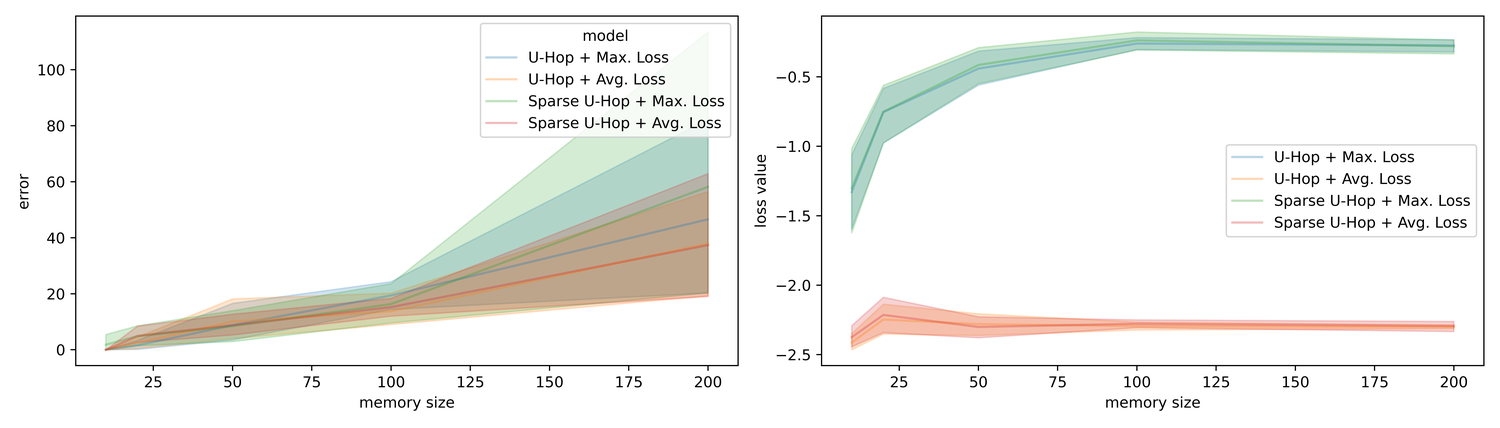}
    \caption{
    \textbf{Max. vs Avg. Loss on MNIST with $N = 10$.}
    }
\end{figure*}

\begin{figure*}[h]
    \centering
    \includegraphics[width=\textwidth]{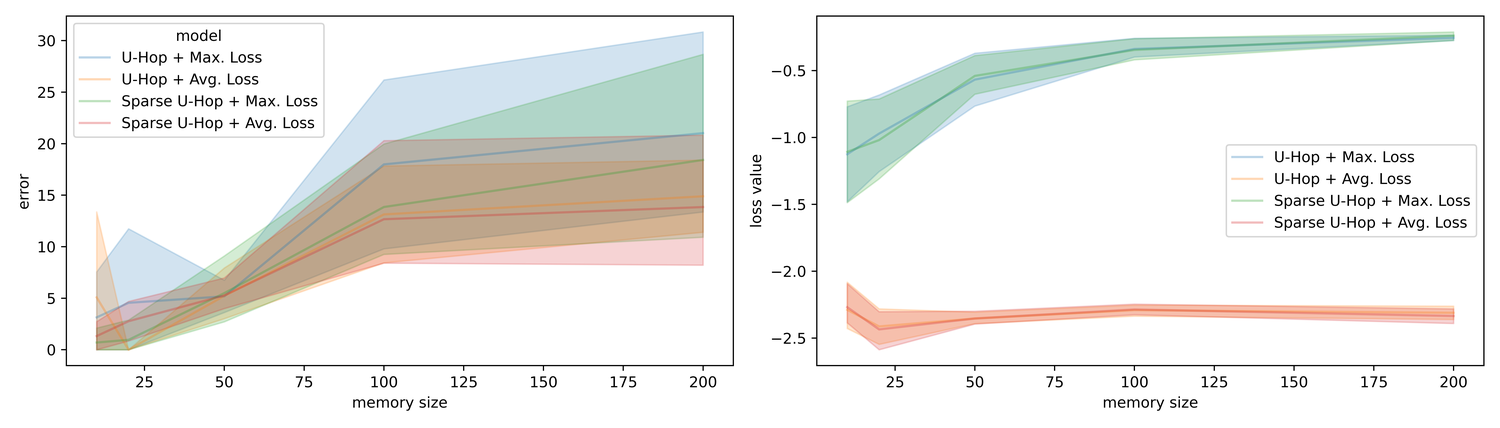}
    \caption{
    \textbf{Max. vs Avg. Loss on MNIST with $N = 20$.}
    }
\end{figure*}

 \begin{figure*}[h]
    \centering
    \includegraphics[width=\textwidth]{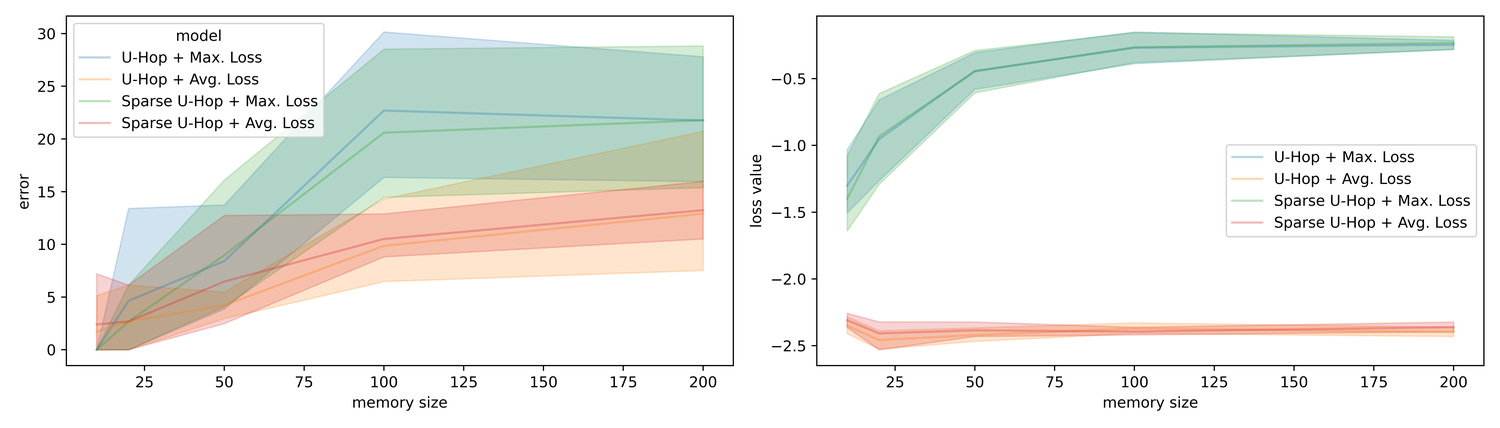}
    \caption{
    \textbf{Max. vs Avg. Loss on MNIST with $N = 30$.}
    }
\end{figure*}

\begin{figure*}[h]
    \centering
    \includegraphics[width=\textwidth]{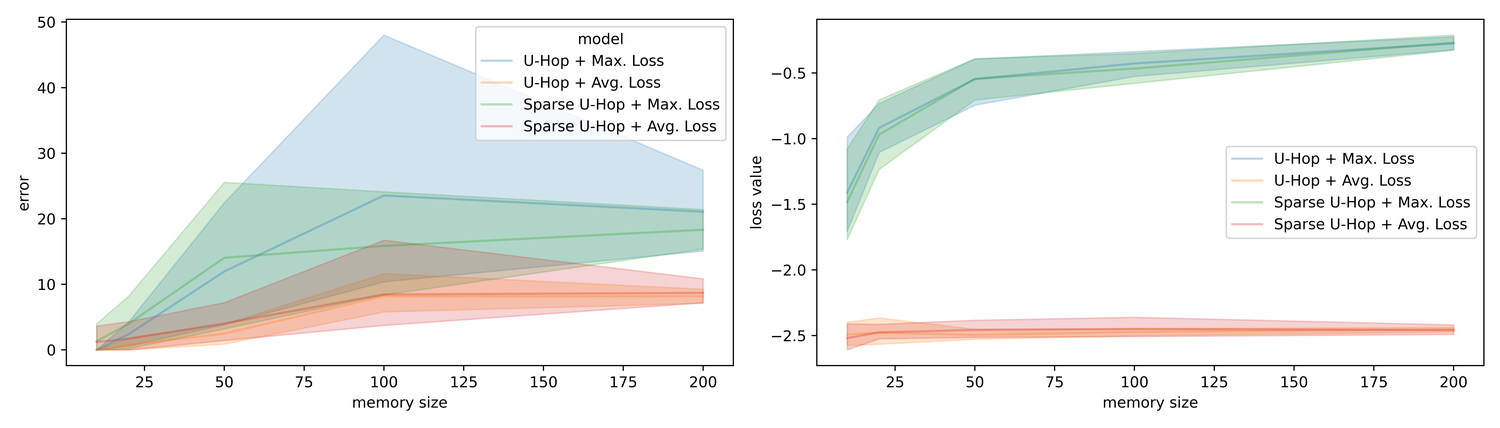}
    \caption{
    \textbf{Max. vs Avg. Loss on MNIST with $N = 50$.}
    }
\end{figure*}

\begin{figure*}[h]
    \centering
    \includegraphics[width=\textwidth]{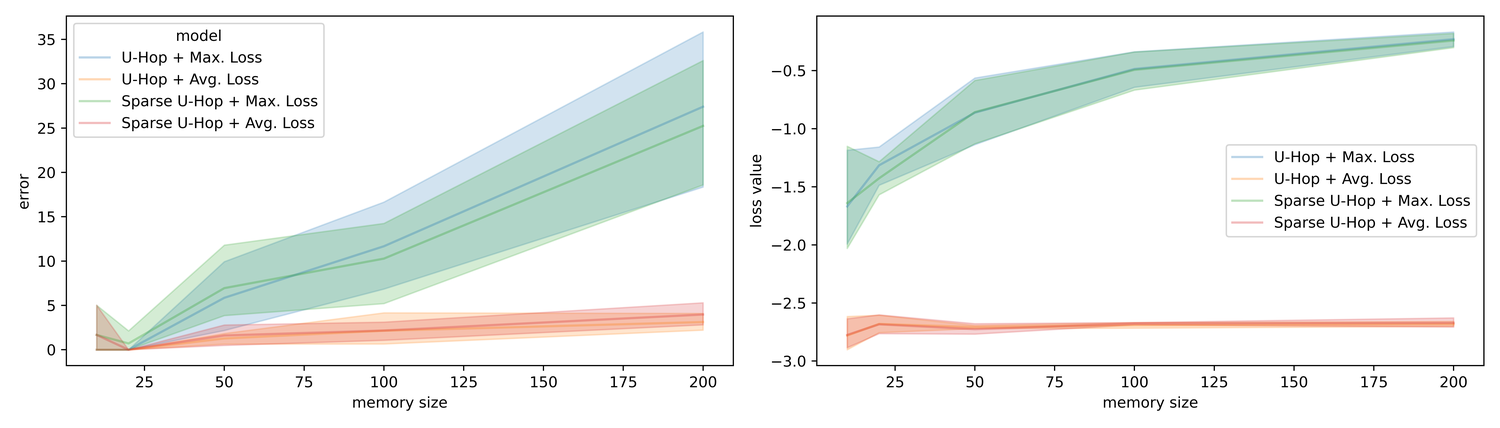}
    \caption{
    \textbf{Max. vs Avg. Loss on MNIST with $N = 100$.}
    }
\end{figure*}
\begin{figure*}[h]
    \centering
    \includegraphics[width=\textwidth]{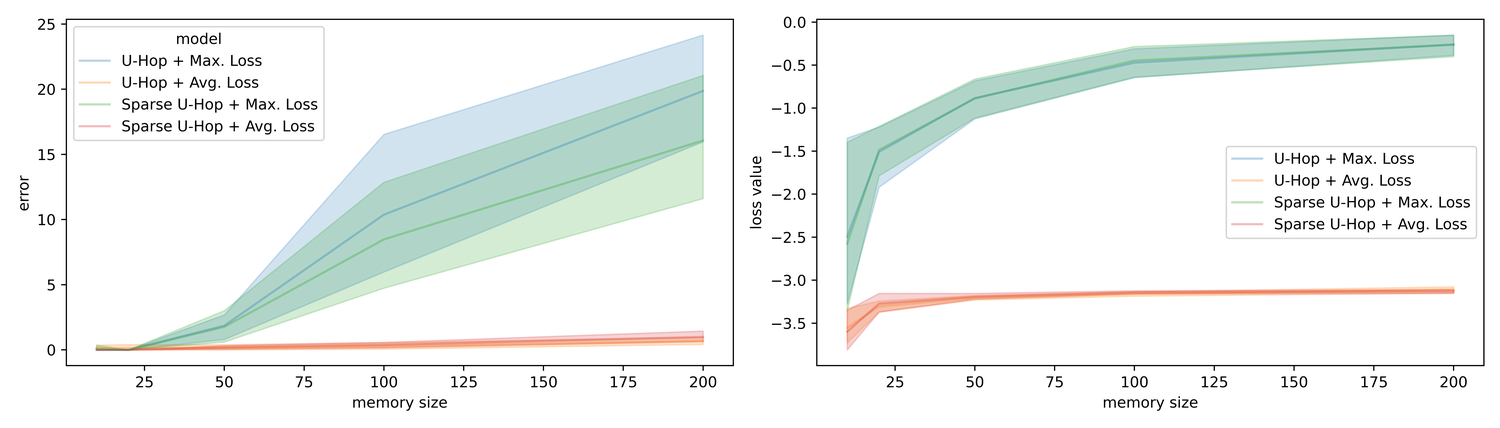}
    \caption{
    \textbf{Max. vs Avg. Loss on MNIST with $N = 200$.}
    }
\end{figure*}

\begin{figure*}[h]
    \centering
    \includegraphics[width=\textwidth]{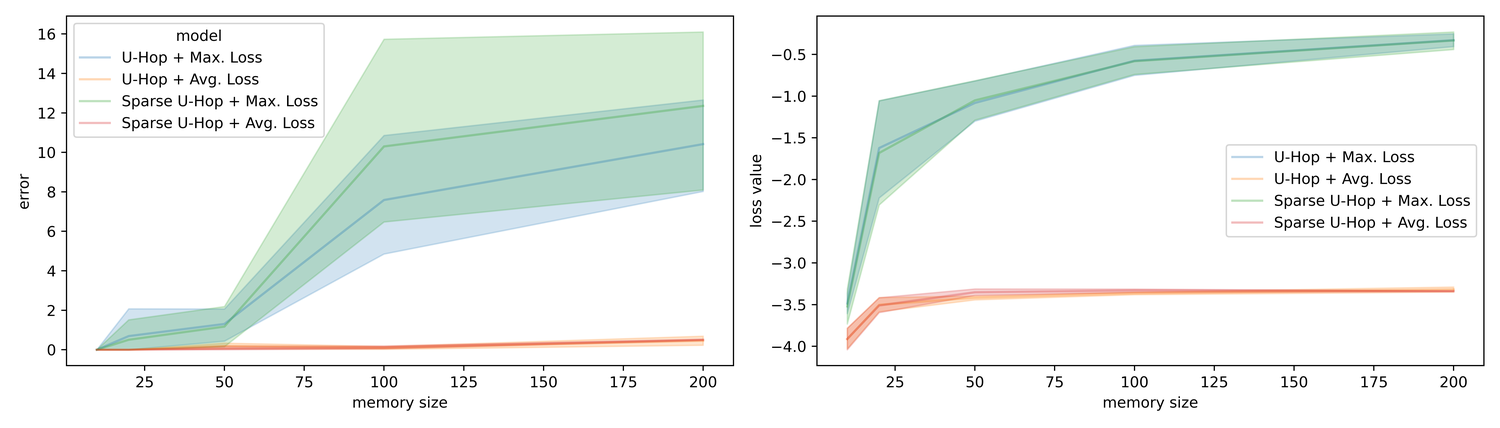}
    \caption{
    \textbf{Max. vs Avg. Loss on MNIST with $N = 10$.}
    }
\end{figure*}

\begin{figure*}[h]
    \centering
    \includegraphics[width=\textwidth]{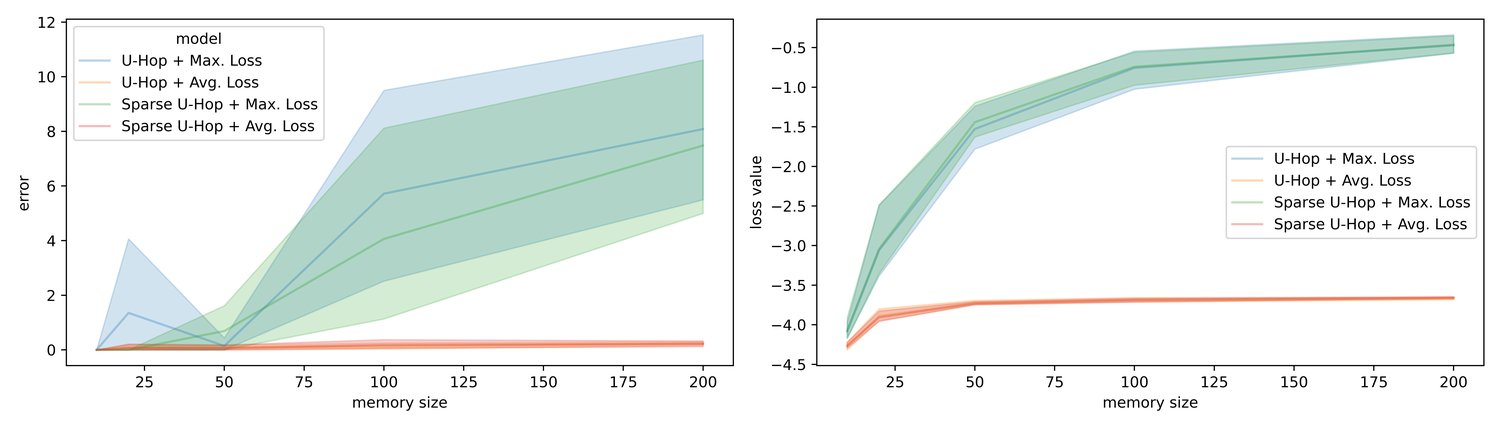}
    \caption{
    \textbf{Max. vs Avg. Loss on MNIST with $N = 500$.}
    }
\end{figure*}

\clearpage

\paragraph{Max. Avg. Loss Comparison on CIFAR10.}
With CIFAR10 being more difficult comparing to MNIST, {\uhop} under average loss still outperforms {\uhop} under Max. loss.

\begin{figure*}[h]\label{cifar-avg-max-epoch}
    \centering
    \includegraphics[width=\textwidth]{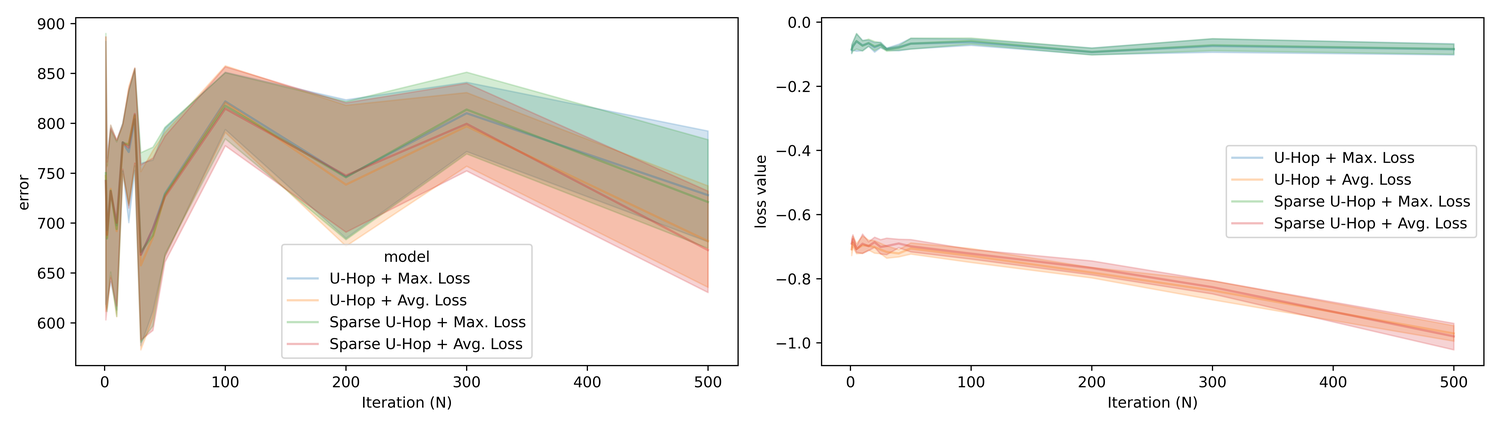}
    \caption{
    \textbf{Loss Value v.s. $N$ and Retrieval Error v.s. $N$ on CIFAR10.}
    }
\end{figure*}

\begin{figure*}[h]
    \centering
    \includegraphics[width=\textwidth]{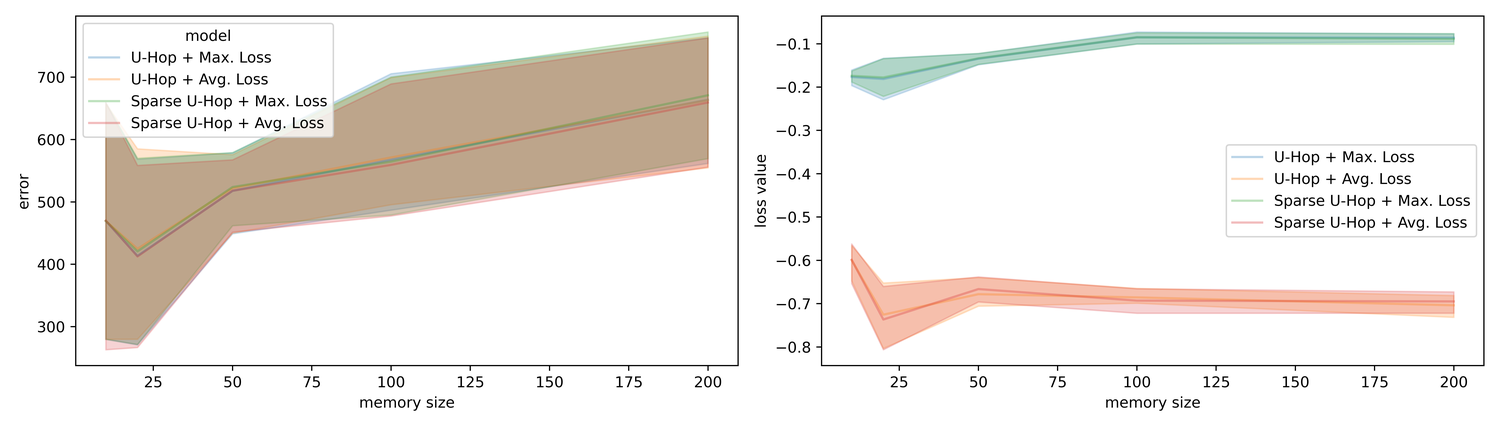}
    \caption{
    \textbf{Max. vs Avg. Loss on CIFAR10 with $N = 10$.}
    }
\end{figure*}

\begin{figure*}[h]
    \centering
    \includegraphics[width=\textwidth]{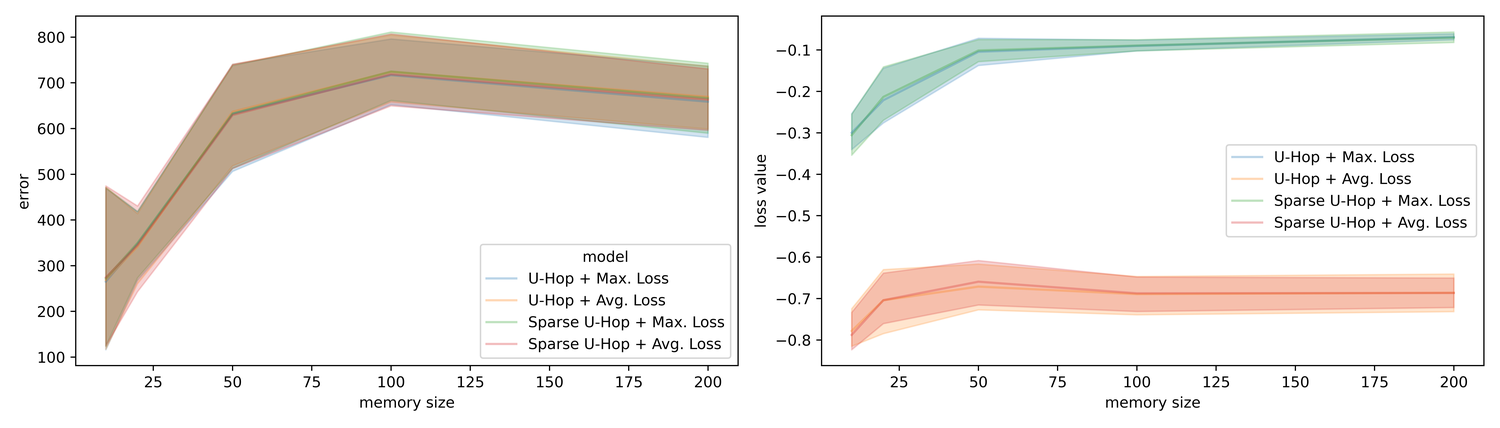}
    \caption{
    \textbf{Max. vs Avg. Loss on CIFAR10 with $N = 20$.}
    }
\end{figure*}

 \begin{figure*}[h]
    \centering
    \includegraphics[width=\textwidth]{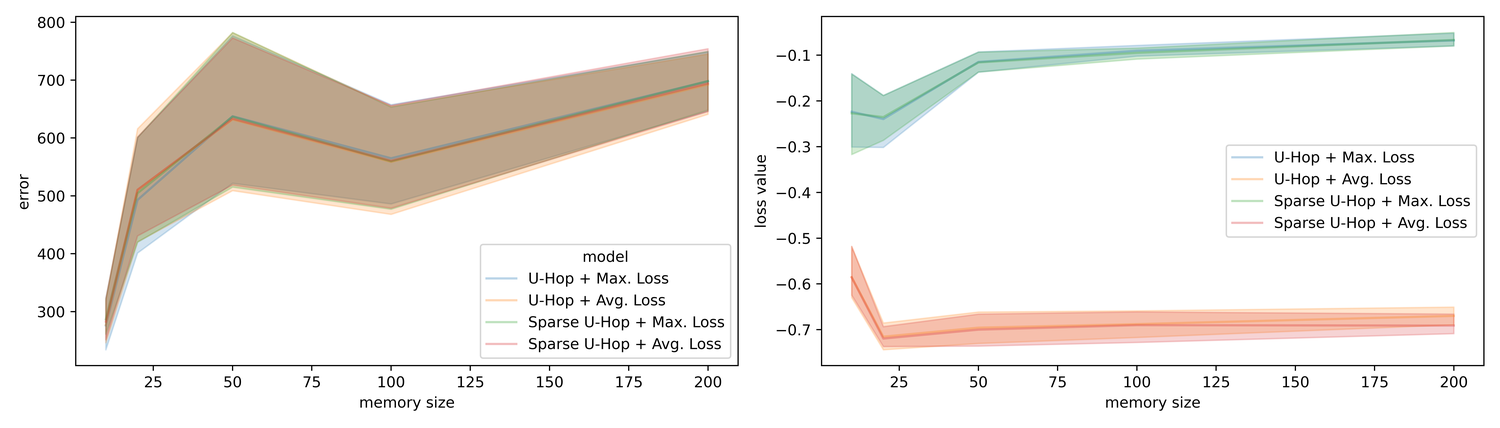}
    \caption{
    \textbf{Max. vs Avg. Loss on CIFAR10 with $N = 30$.}
    }
\end{figure*}

\begin{figure*}[h]
    \centering
    \includegraphics[width=\textwidth]{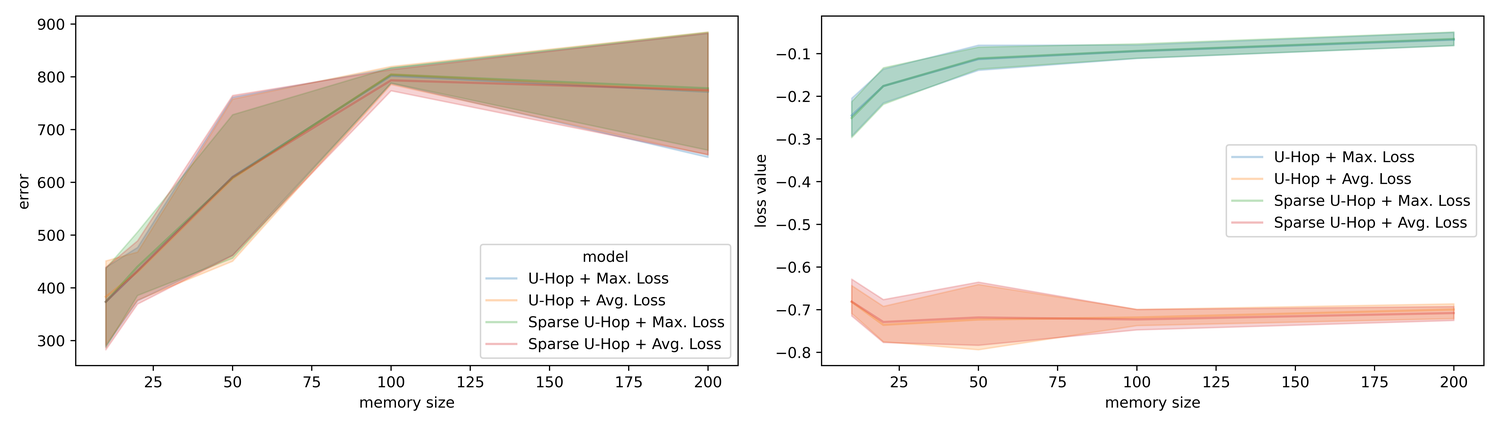}
    \caption{
    \textbf{Max. vs Avg. Loss on CIFAR10 with $N = 50$.}
    }
\end{figure*}

\begin{figure*}[h]
    \centering
    \includegraphics[width=\textwidth]{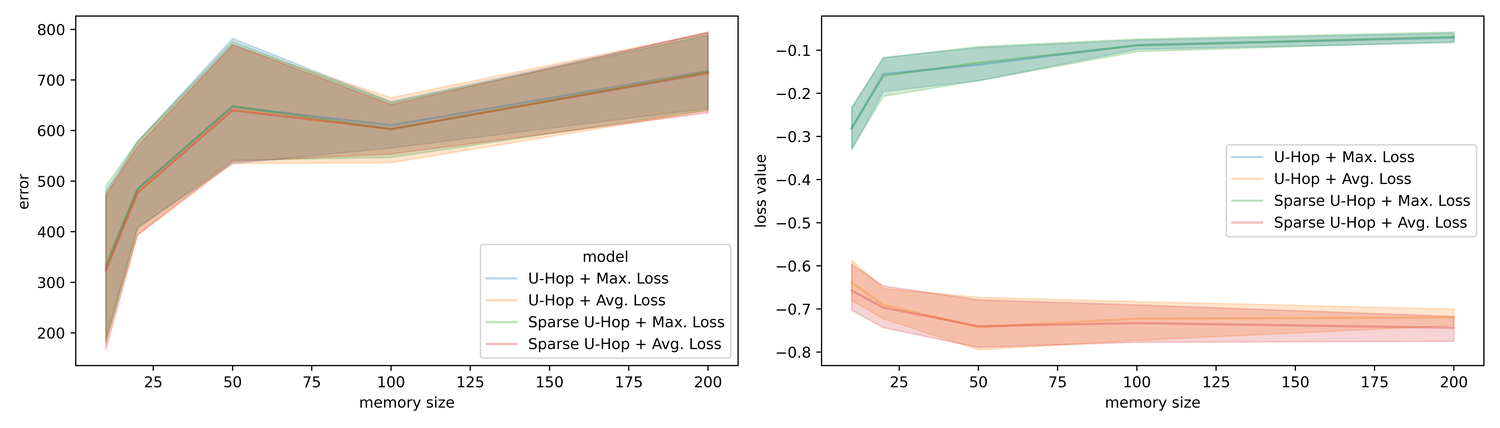}
    \caption{
    \textbf{Max. vs Avg. Loss on CIFAR10 with $N = 100$.}
    }
\end{figure*}
\begin{figure*}[h]
    \centering
    \includegraphics[width=\textwidth]{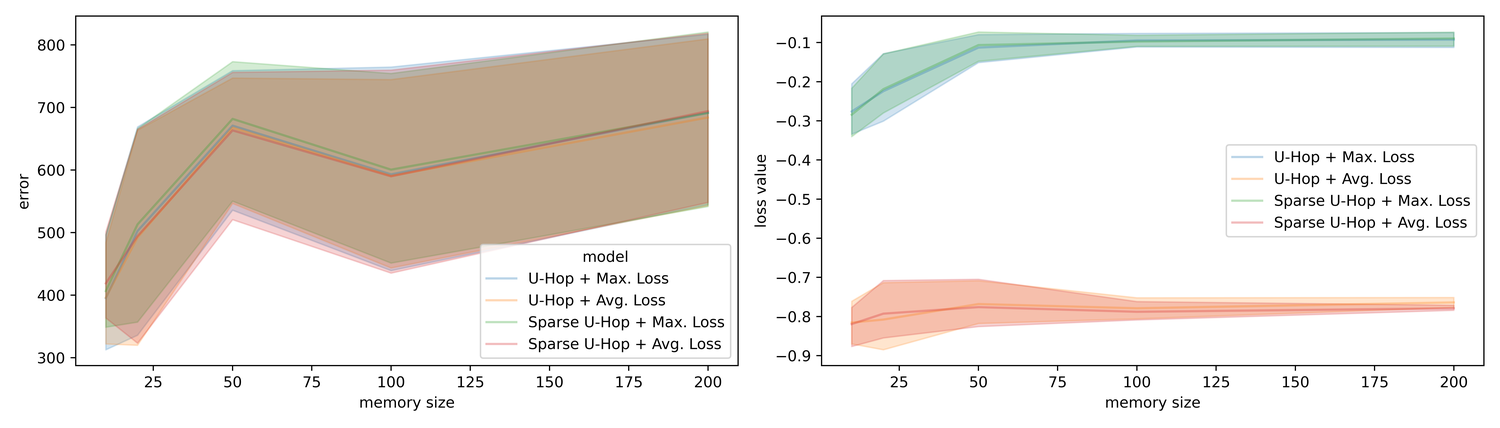}
    \caption{
    \textbf{Max. vs Avg. Loss on CIFAR10 with $N = 200$.}
    }
\end{figure*}

\begin{figure*}[h]
    \centering
    \includegraphics[width=\textwidth]{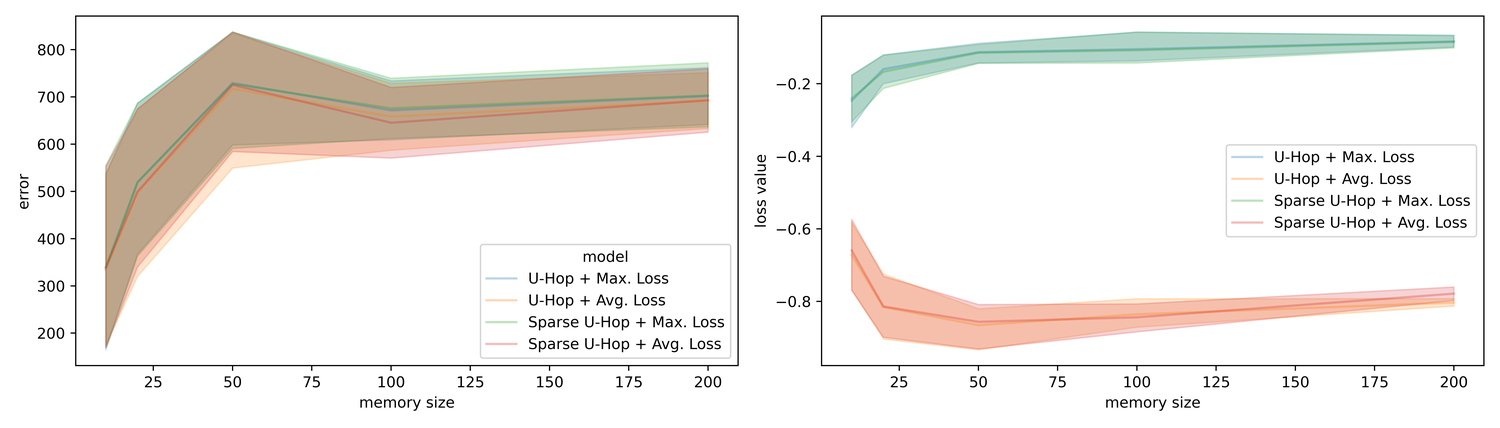}
    \caption{
    \textbf{Max. vs Avg. Loss on CIFAR10 with $N = 250$.}
    }
\end{figure*}

\begin{figure*}[h]
    \centering
    \includegraphics[width=\textwidth]{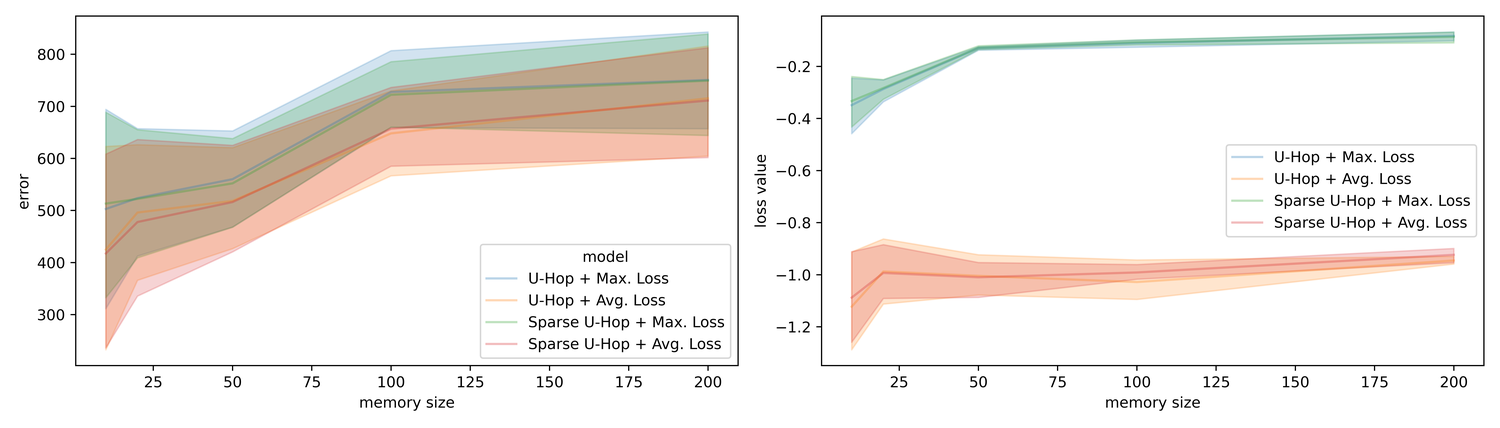}
    \caption{
    \textbf{Max. vs Avg. Loss on CIFAR10 with $N = 500$.}
    }
\end{figure*}

\clearpage

\subsection{Memory Retrieval}\label{mem-ret-clear}

Here we again show the memory retrieval results with higher resolution.
The result demonstrates with {\uhop}, modern Hopfield models obtain significant improvement on both datasets.

\clearpage

\subsection{Model Expressiveness}\label{sec:expressive}
Here we present the model expressiveness on training data.
This is a empirical validation for \cref{thm:perfect_kernel}.
We observe that without {\uhop}, baseline models suffer from sharp performance drop, which also lead to generalization degradation as show in \cref{app:extra-exp-cifar10}.
In contrast, with {\uhop}, models show better robustness against sample size increase.

\begin{figure*}[h]
    \centering
    \includegraphics[width=\textwidth]{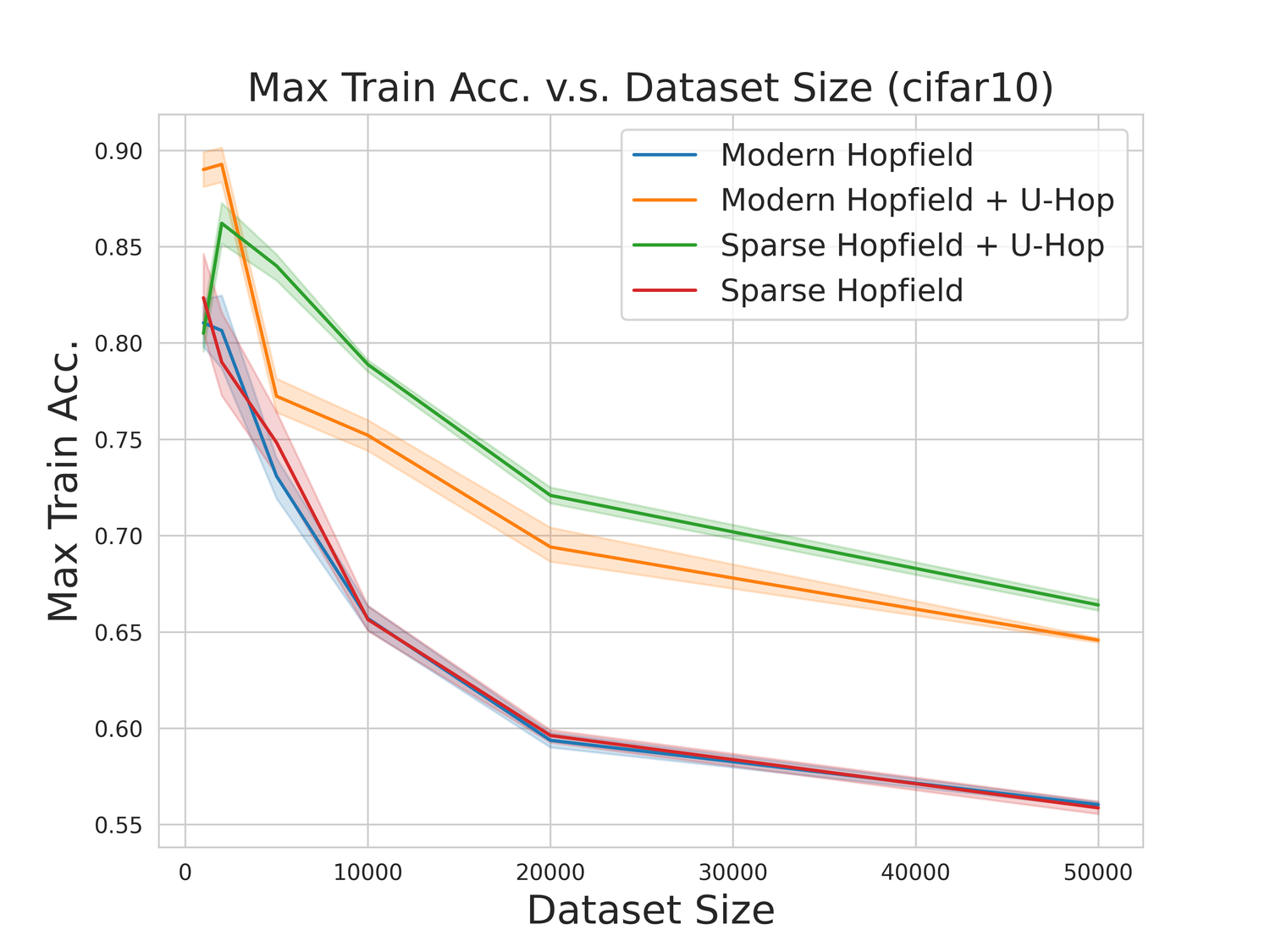}
    \caption{
    \textbf{Max Training Accuracy with respect to Sample Size Increase on CIFAR10}
    }
    \label{fig:cifar10-max}
\end{figure*}

\begin{figure*}[h]
    \centering
    \includegraphics[width=\textwidth]{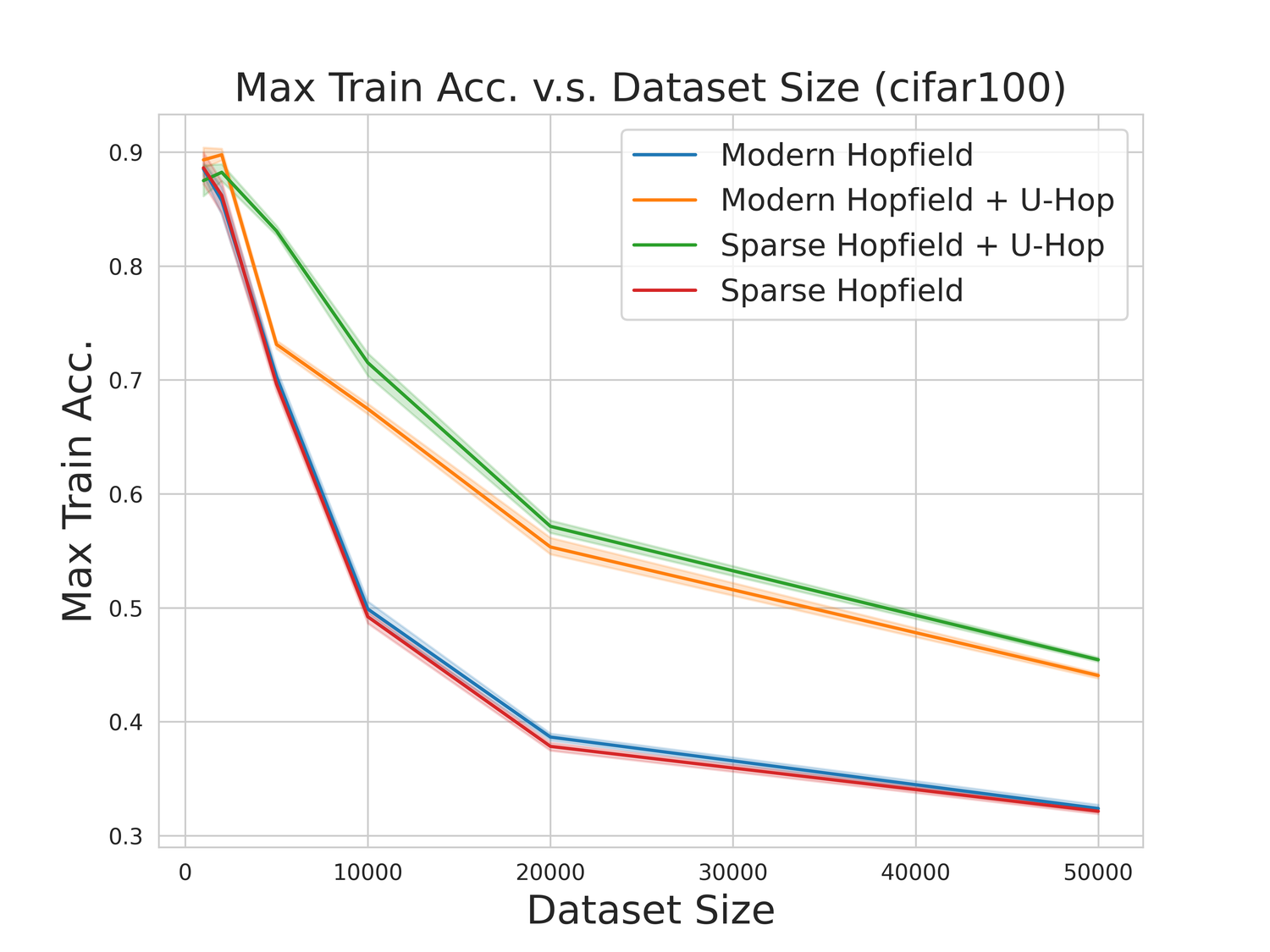}
    \caption{
    \textbf{Max Training Accuracy with respect to Sample Size Increase on CIFAR100}
    }
    \label{fig:cifar100-max}
\end{figure*}

\begin{figure*}[h]
    \centering
    \includegraphics[width=\textwidth]{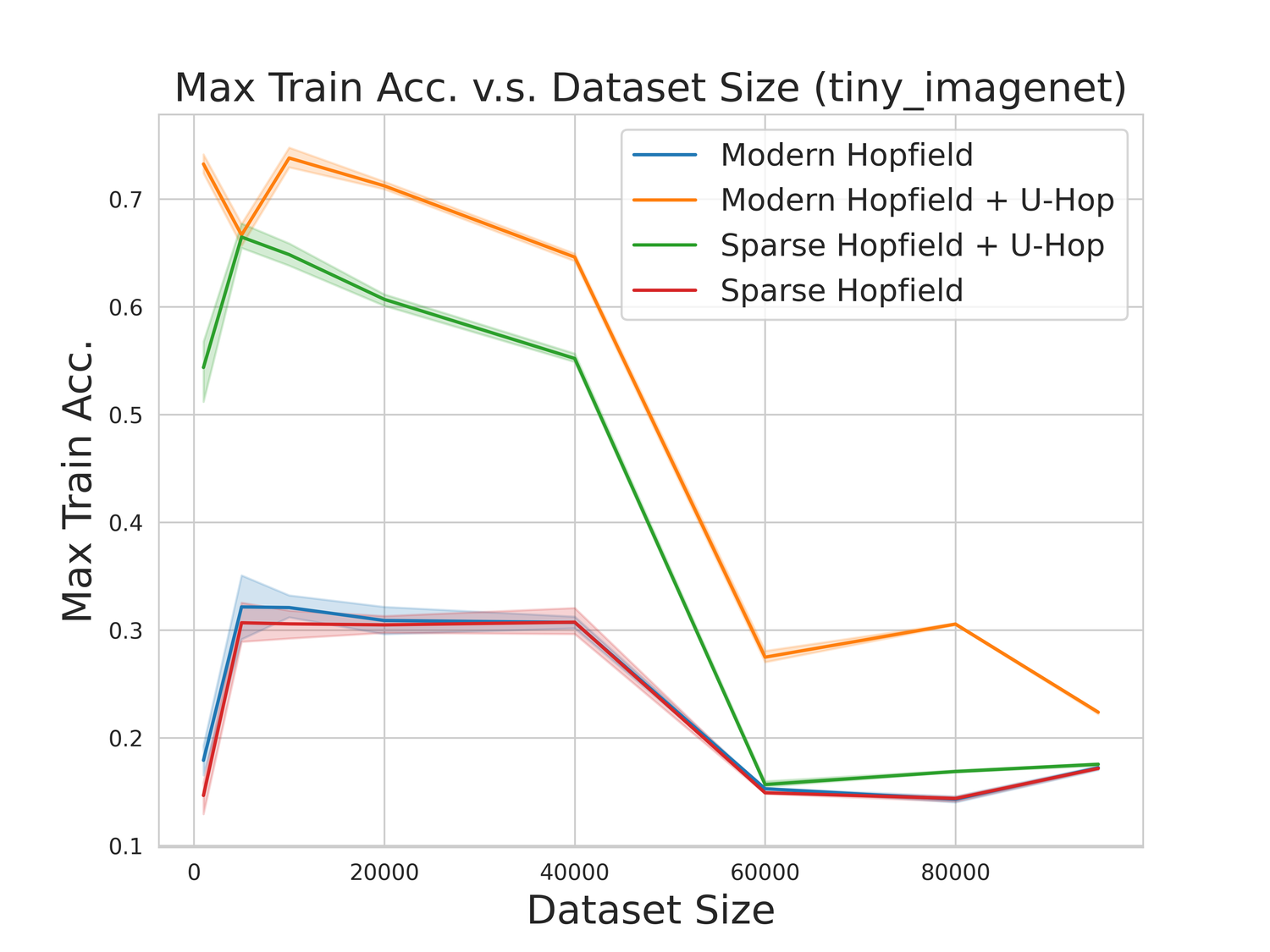}
    \caption{
    \textbf{Max Training Accuracy with respect to Sample Size Increase on Tiny ImageNet.}
    }
    \label{fig:cifar10-nax}
\end{figure*}

\clearpage

\subsection{Classification}\label{app:cls-extra}
Here we conduct empirical analysis on the correlation between dataset size and model convergence.
In general, it is more difficult to memorize all samples for larger dataset.
However, learning from more samples might lead to a better generalization performance, results in higher test accuracy.

\subsubsection{CIFAR10}\label{app:extra-exp-cifar10}
\paragraph{Results.}
The result demonstrates {\uhop} significantly improves model's performance on 3 aspects: 
\begin{itemize}
    \item (i) Generalization (test accuracy) 
    
    \item (ii) Convergence Speed 
    
    \item (iii) Memorization (training accuracy) 
\end{itemize}
The improvement also became more obvious when the dataset size increases.
This is reasonable as the standard $\mathtt{Hopfield}$ layer is powerful enough to memorize small sample size with and without {\uhop}.

\begin{figure*}[h]
    \centering
    \includegraphics[width=\textwidth]{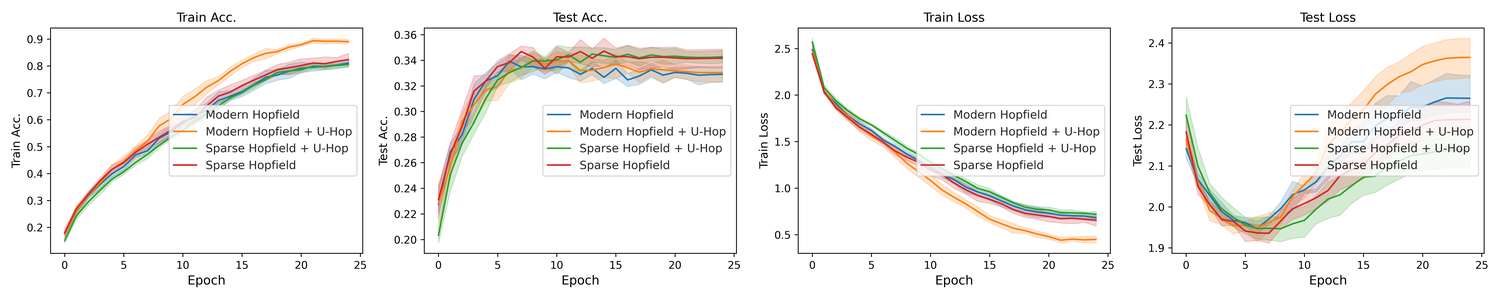}
    \caption{
    \textbf{CIFAR10 Convergence Comparison with Dataset Size=1000}
    Left to right: Training Accuracy, Test Accuracy, Training Loss and Test Loss.
    }
    \label{fig:cifar10-1000}
\end{figure*}

\begin{figure*}[h]
    \centering
    \includegraphics[width=\textwidth]{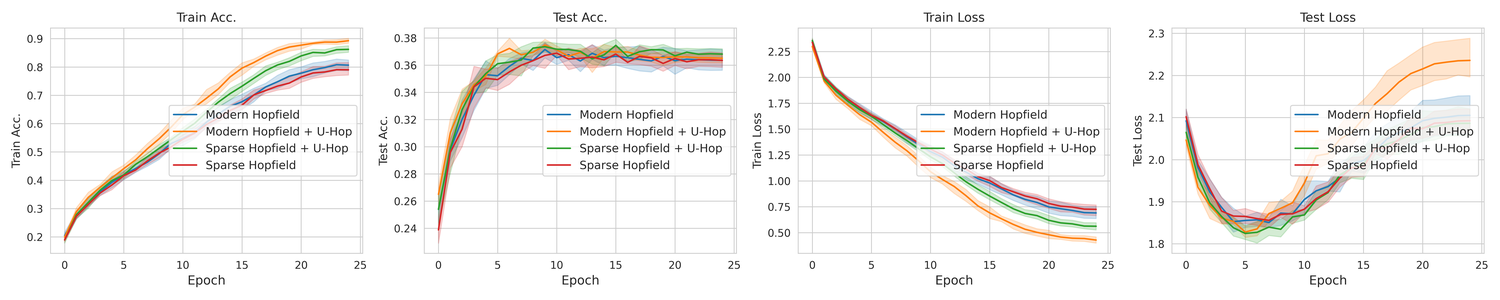}
    \caption{
    \textbf{CIFAR10 Convergence Comparison with Dataset Size=2000}
    Left to right: Training Accuracy, Test Accuracy, Training Loss and Test Loss.
    }
    \label{fig:cifar10-2000}
\end{figure*}

\begin{figure*}[h]
    \centering
    \includegraphics[width=\textwidth]{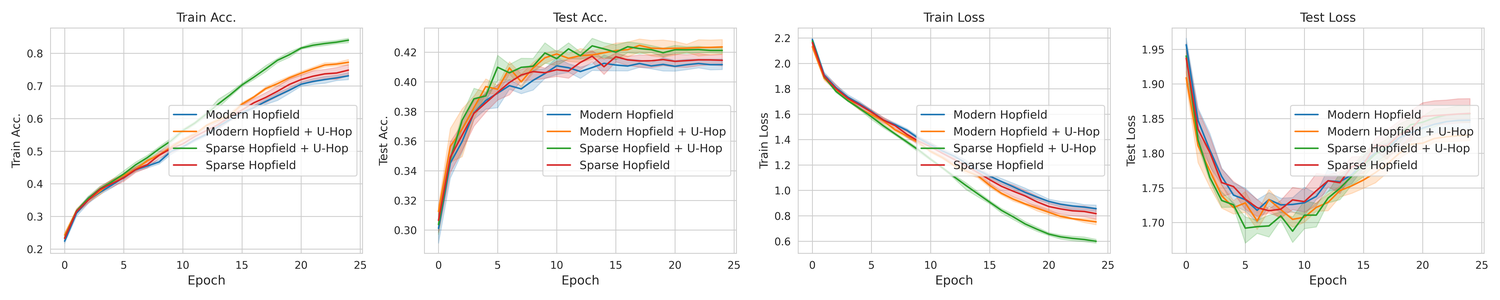}
    \caption{
    \textbf{CIFAR10 Convergence Comparison with Dataset Size=5000}    Left to right: Training Accuracy, Test Accuracy, Training Loss and Test Loss.
    }
    \label{fig:cifar10-5000}
\end{figure*}

\begin{figure*}[h]
    \centering
    \includegraphics[width=\textwidth]{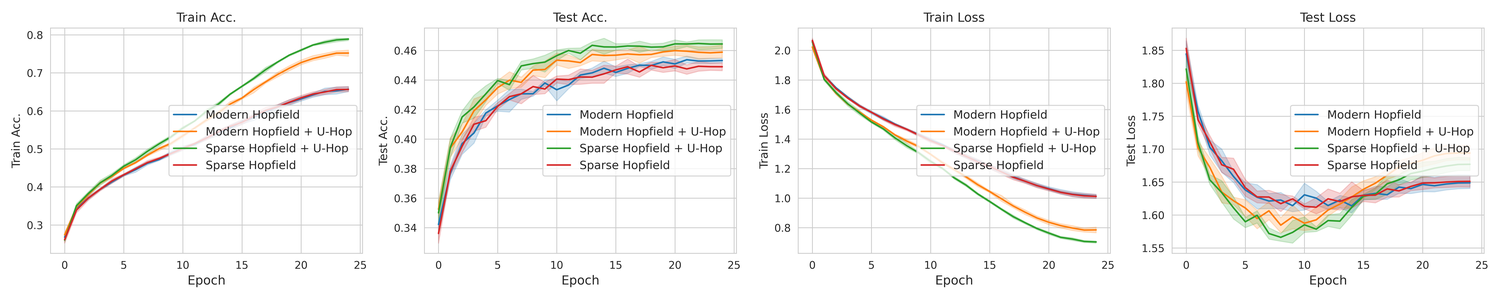}
    \caption{
    \textbf{CIFAR10 Convergence Comparison with Dataset Size=10000}    Left to right: Training Accuracy, Test Accuracy, Training Loss and Test Loss.
    }
    \label{fig:cifar10-10000}
\end{figure*}

\begin{figure*}[h]
    \centering
    \includegraphics[width=\textwidth]{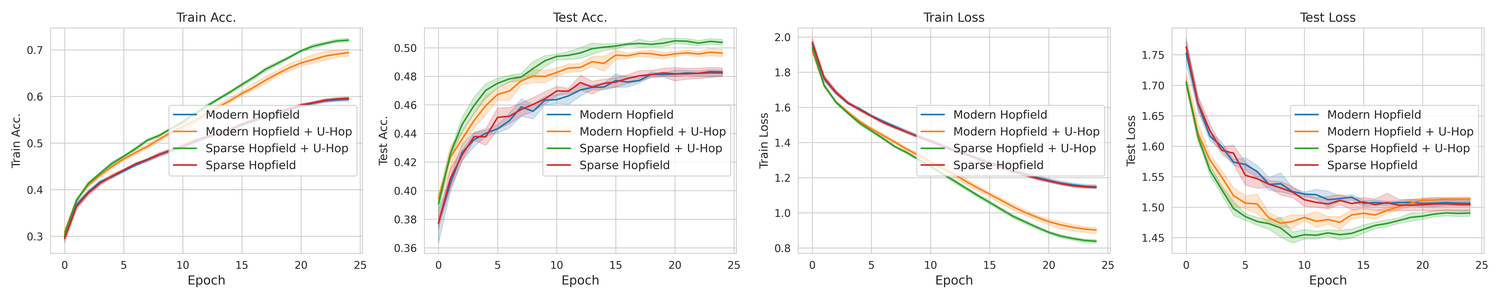}
    \caption{
    \textbf{CIFAR10 Convergence Comparison with Dataset Size=20000}    Left to right: Training Accuracy, Test Accuracy, Training Loss and Test Loss.
    }
    \label{fig:cifar10-20000}
\end{figure*}

\begin{figure*}[h]
    \centering
    \includegraphics[width=\textwidth]{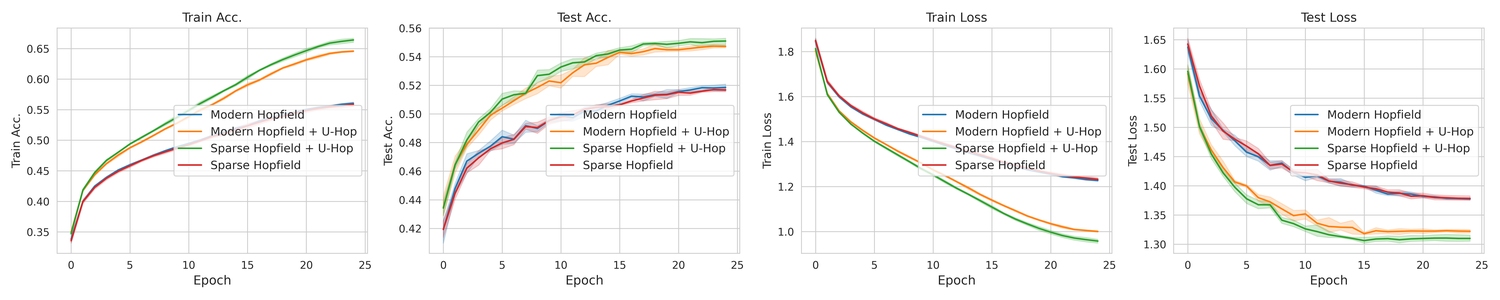}
    \caption{
    \textbf{CIFAR10 Convergence Comparison with Dataset Size=Full}    Left to right: Training Accuracy, Test Accuracy, Training Loss and Test Loss.
    }
    \label{fig:cifar10-50000}
\end{figure*}

\clearpage

\subsubsection{CIFAR100}\label{app:extra-exp-cifar100}

\paragraph{Results.}
The model behavior was similar comparing to what we observe from CIFAR10.
The performance improvement under {\uhop} became stronger with the increase of dataset size.
With {\uhop}, the model improves on both convergence speed, memorization capacity (training accuracy) and generalization power (test accuracy).

\begin{figure*}[h]
    \centering
    \includegraphics[width=\textwidth]{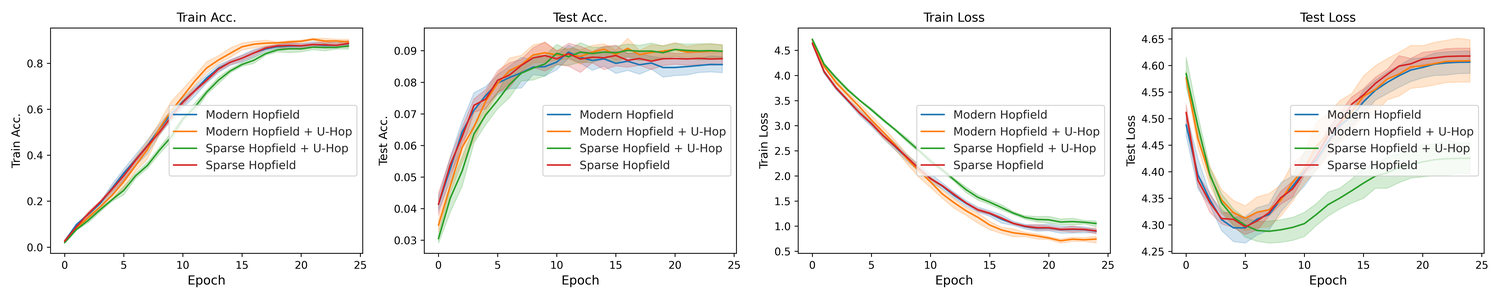}
    \caption{
    \textbf{CIFAR100 Convergence Comparison with Dataset Size=1000}    Left to right: Training Accuracy, Test Accuracy, Training Loss and Test Loss.
    }
    \label{fig:cifar100-1000}
\end{figure*}

\begin{figure*}[h]
    \centering
    \includegraphics[width=\textwidth]{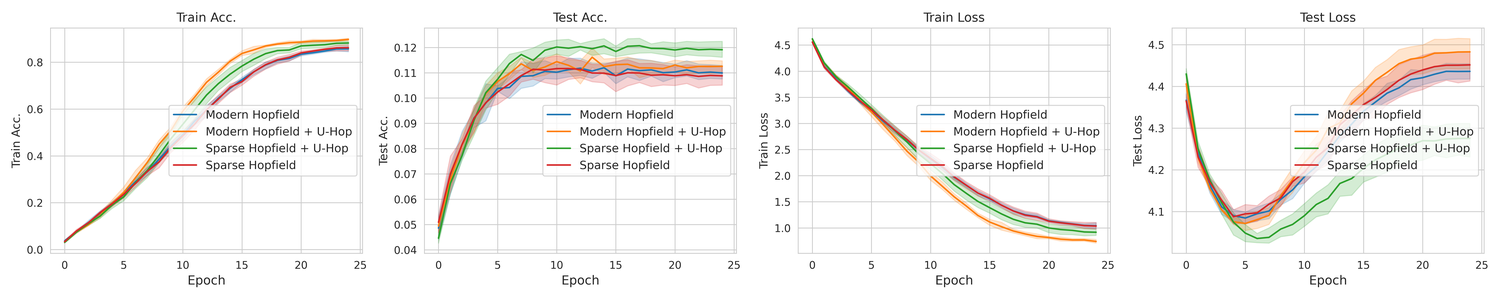}
    \caption{
    \textbf{CIFAR100 Convergence Comparison with Dataset Size=2000}    Left to right: Training Accuracy, Test Accuracy, Training Loss and Test Loss.
    }
    \label{fig:cifar100-2000}
\end{figure*}

\begin{figure*}[h]
    \centering
    \includegraphics[width=\textwidth]{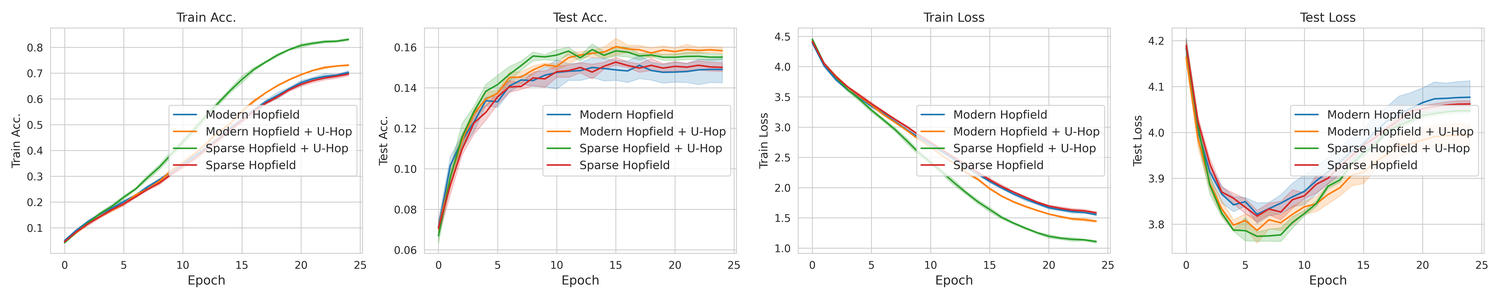}
    \caption{
    \textbf{CIFAR100 Convergence Comparison with Dataset Size=5000}    Left to right: Training Accuracy, Test Accuracy, Training Loss and Test Loss.
    }
    \label{fig:cifar100-5000}
\end{figure*}

\begin{figure*}[h]
    \centering
    \includegraphics[width=\textwidth]{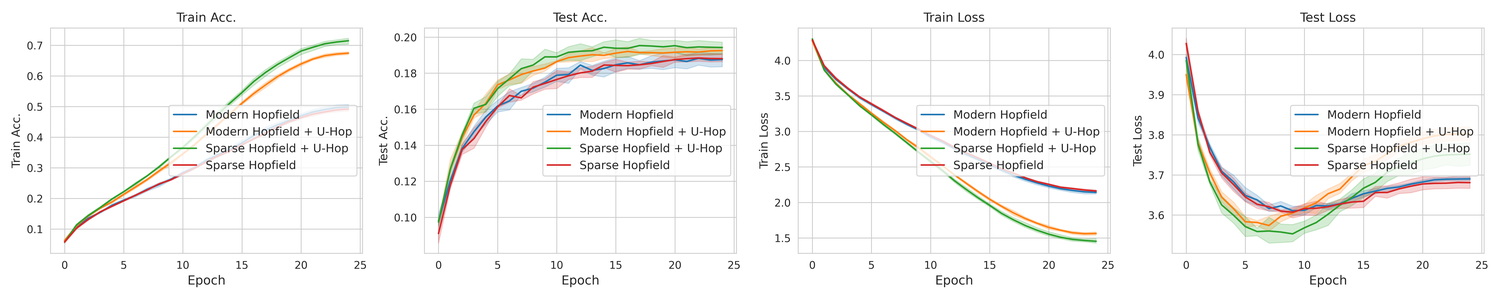}
    \caption{
    \textbf{CIFAR100 Convergence Comparison with Dataset Size=10000}    Left to right: Training Accuracy, Test Accuracy, Training Loss and Test Loss.
    }
    \label{fig:cifar100-10000}
\end{figure*}

\begin{figure*}[h]
    \centering
    \includegraphics[width=\textwidth]{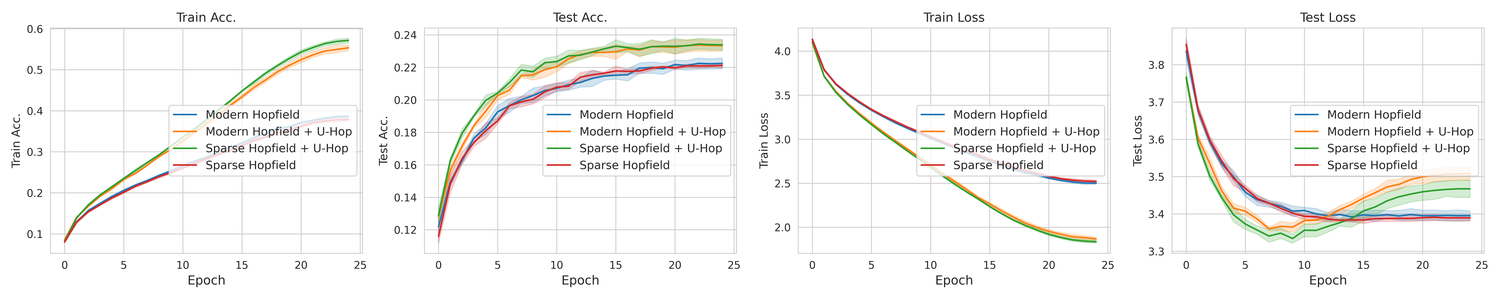}
    \caption{
    \textbf{CIFAR100 Convergence Comparison with Dataset Size=20000}    Left to right: Training Accuracy, Test Accuracy, Training Loss and Test Loss.
    }
    \label{fig:cifar100-20000}
\end{figure*}

\begin{figure*}[h]
    \centering
    \includegraphics[width=\textwidth]{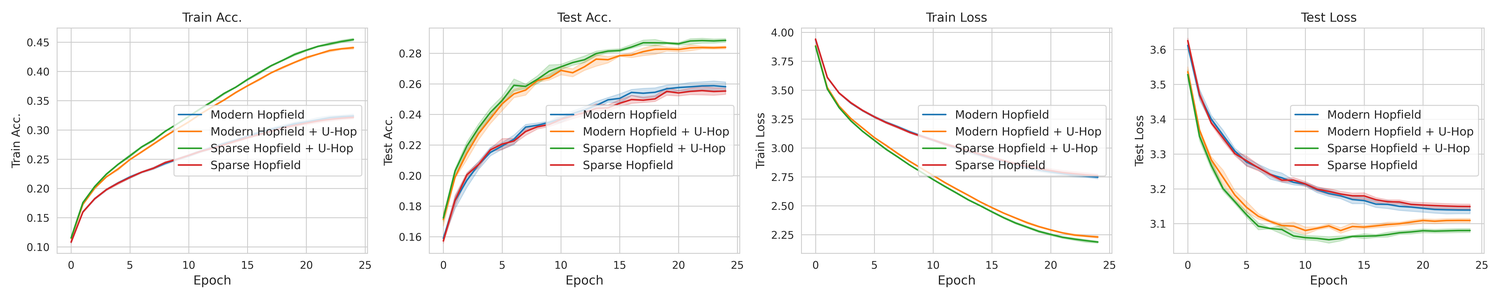}
    \caption{
    \textbf{CIFAR100 Convergence Comparison with Dataset Size=Full}    Left to right: Training Accuracy, Test Accuracy, Training Loss and Test Loss.
    }
    \label{fig:cifar100-50000}
\end{figure*}
\clearpage

\subsubsection{Tiny ImageNet}\label{app:extra-exp-tiny}
Models under {\uhop} continue to show strong performance against baselines on Tiny ImageNet dataset.
Notably, we use a 3 layer encoder for this dataset, which provides additional insights ensuring that {\uhop} works well under deep neural network architecture.

\begin{figure*}[h]
    \centering
    \includegraphics[width=\textwidth]{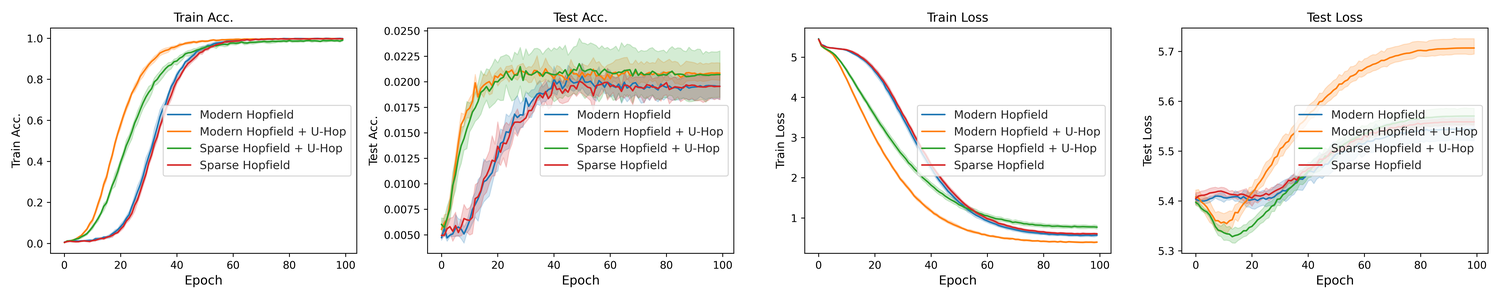}
    \caption{
    \textbf{Tiny ImageNet Convergence Comparison with Dataset Size=1000}    Left to right: Training Accuracy, Test Accuracy, Training Loss and Test Loss.
    }
    \label{fig:tiny-1000}
\end{figure*}

\begin{figure*}[h]
    \centering
    \includegraphics[width=\textwidth]{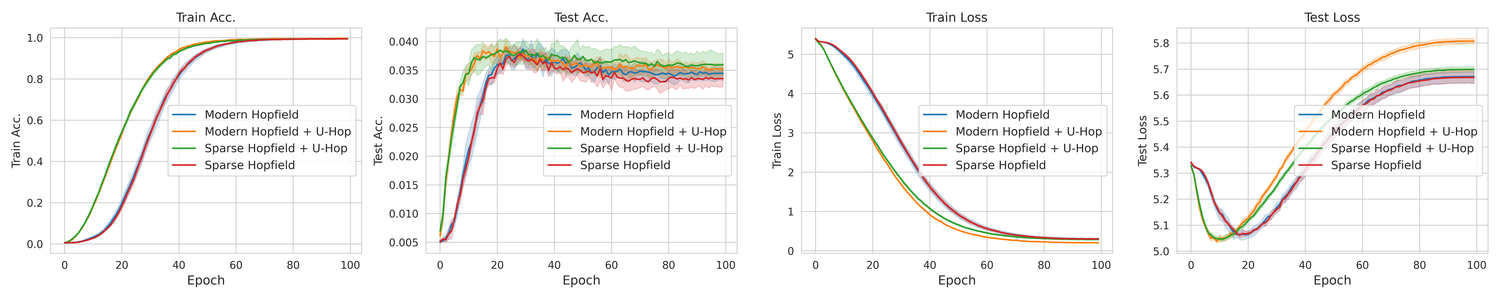}
    \caption{
    \textbf{Tiny ImageNet Convergence Comparison with Dataset Size=5000}    Left to right: Training Accuracy, Test Accuracy, Training Loss and Test Loss.
    }
    \label{fig:tiny-5000}
\end{figure*}

\begin{figure*}[h]
    \centering
    \includegraphics[width=\textwidth]{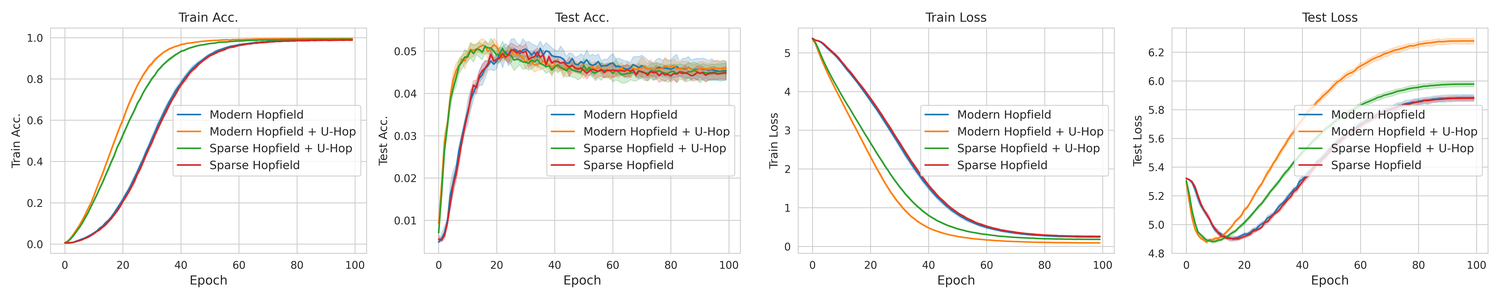}
    \caption{
    \textbf{Tiny ImageNet Convergence Comparison with Dataset Size=10000}    Left to right: Training Accuracy, Test Accuracy, Training Loss and Test Loss.
    }
    \label{fig:tiny-10000}
\end{figure*}

\begin{figure*}[h]
    \centering
    \includegraphics[width=\textwidth]{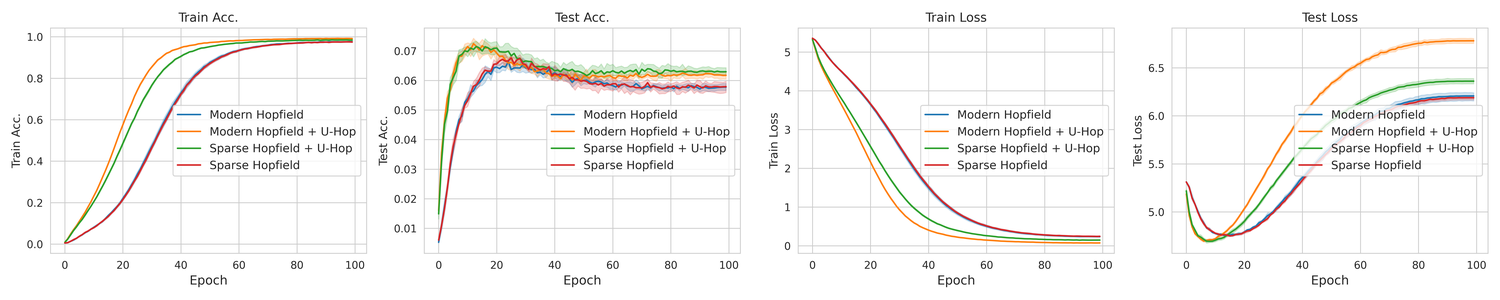}
    \caption{
    \textbf{Tiny ImageNet Convergence Comparison with Dataset Size=20000}    Left to right: Training Accuracy, Test Accuracy, Training Loss and Test Loss.
    }
    \label{fig:tiny-20000}
\end{figure*}
\begin{figure*}[h]
    \centering
    \includegraphics[width=\textwidth]{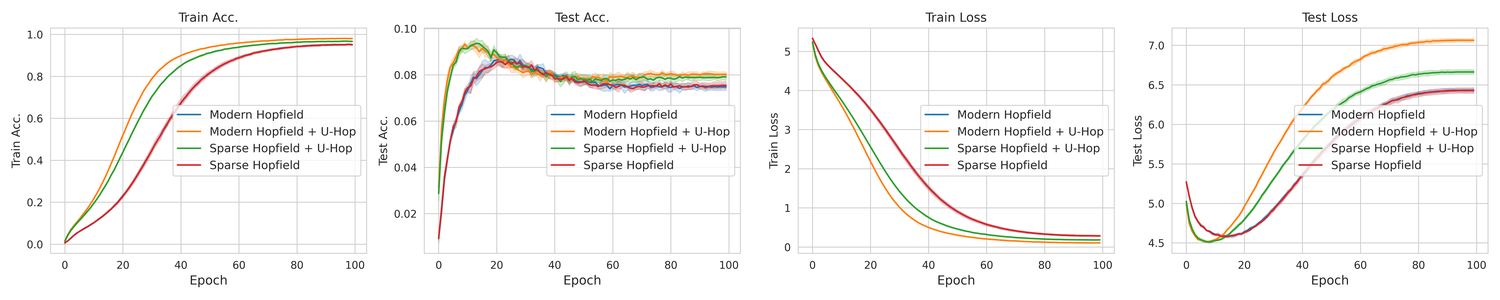}
    \caption{
    \textbf{Tiny ImageNet Convergence Comparison with Dataset Size=40000}    Left to right: Training Accuracy, Test Accuracy, Training Loss and Test Loss.
    }
    \label{fig:tiny-40000}
\end{figure*}

\begin{figure*}[h]
    \centering
    \includegraphics[width=\textwidth]{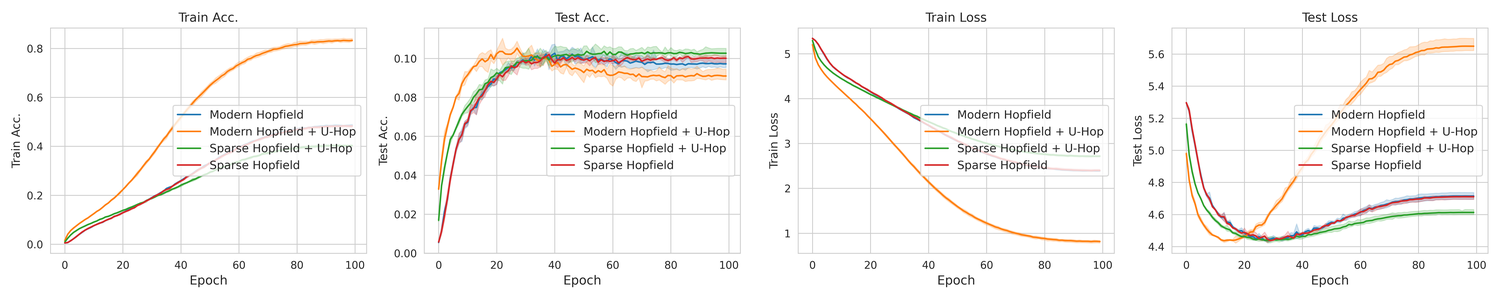}
    \caption{
    \textbf{Tiny ImageNet Convergence Comparison with Dataset Size=60000}    Left to right: Training Accuracy, Test Accuracy, Training Loss and Test Loss.
    }
    \label{fig:tiny-60000}
\end{figure*}

\begin{figure*}[h]
    \centering
    \includegraphics[width=\textwidth]{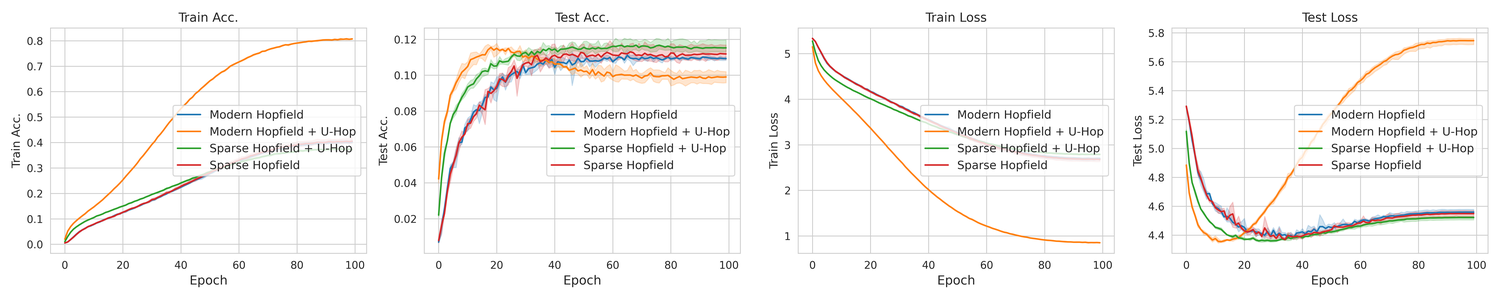}
    \caption{
    \textbf{Tiny ImageNet Convergence Comparison with Dataset Size=80000}    Left to right: Training Accuracy, Test Accuracy, Training Loss and Test Loss.
    }
    \label{fig:tiny-80000}
\end{figure*}

\begin{figure*}[h]
    \centering
    \includegraphics[width=\textwidth]{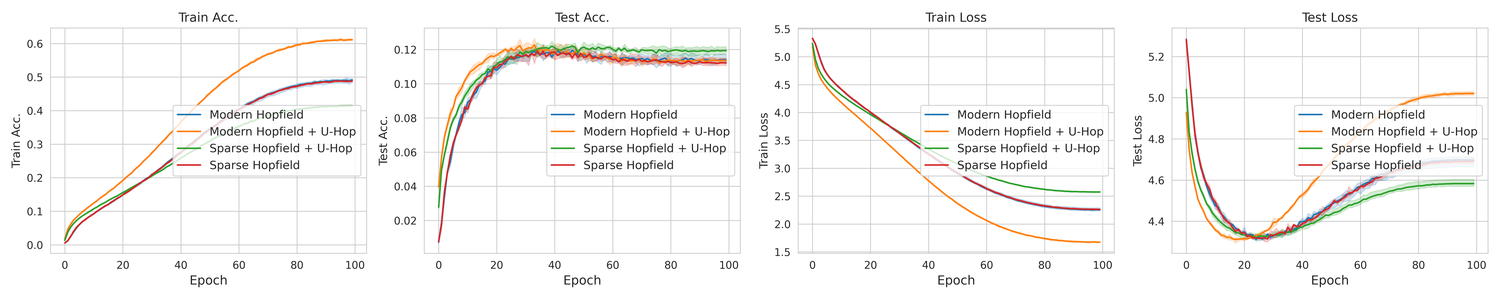}
    \caption{
    \textbf{Tiny ImageNet Convergence Comparison with Dataset Size=Full}    Left to right: Training Accuracy, Test Accuracy, Training Loss and Test Loss.
    }
    \label{fig:tiny-Full}
\end{figure*}

\clearpage

\subsection{Time Series Prediction}\label{sec:time-series}
Here we report the results of our time series prediction experiment.
From \cref{table:time-series}, we observe that {\uhop} obtains improvement in most datasets and prediction horizons.

\begin{table}[!h]
\vspace{-1.5em}
\centering
\caption{\small
\textbf{STanHop \cite{wu2023stanhop}: Multivariate Time Series Predictions.}
We compare \textbf{STanHop-Net} \cite{wu2023stanhop} with and without \uhop. 
We report the average Mean Square Error (MSE) and Mean Absolute Error (MAE) metrics with variance omitted as they are all $\le 2$\%.
We evaluated each dataset with different prediction horizons (shown in the second column).
We have the best results \textbf{bolded}.
}
\resizebox{.45\textwidth}{!}{    
\begin{tabular}{cccccccc}
\toprule
 \multicolumn{2}{c}{Models} &  
 \multicolumn{2}{c}{\footnotesize\textbf{STanHop-Net}} &
  \multicolumn{2}{c}{\footnotesize\textbf{STanHop-Net} + \uhop} \\
\midrule
 \multicolumn{2}{c}{Metric} &   MSE & MAE &  MSE & MAE  \\
\midrule
 \multirow{5}{1em}{\rot{ETTh1}} 
 & 96
& $0.395$ & $0.402$ & \cellcolor{LightCyan}$\bm{0.392}$ &\cellcolor{LightCyan} $\bm{0.400}$ \\ 
 & 192 
& $0.425$ & $0.432$ &\cellcolor{LightCyan}  $\bm{0.420}$ &\cellcolor{LightCyan} $\bm{0.428}$  \\ 
 & 336
& $0.495$ & $0.487$ & \cellcolor{LightCyan} $\bm{0.470}$ &\cellcolor{LightCyan} $\bm{0.473}$ \\ 
& 720 
& $0.631$ & $0.600$ &\cellcolor{LightCyan} $\bm{0.572}$&\cellcolor{LightCyan} $\bm{0.559}$ \\ 
\midrule
 \multirow{5}{1em}{\rot{ETTm1}} 
 & 96 
& $0.334$ & $0.366$ & 
\cellcolor{LightCyan}$\bm{0.333}$ &\cellcolor{LightCyan} $\bm{0.365}$ \\ 
 & 192 
& $\bm{0.351}$ & $\bm{0.380}$ 
&\cellcolor{LightCyan} $0.355$ &\cellcolor{LightCyan} $0.385$ \\ 
& 336 
& $\bm{0.391}$ & $\bm{0.393}$ & \cellcolor{LightCyan} $0.392$ & \cellcolor{LightCyan} $0.399$ \\ 
& 720 
& $0.436$ & $0.431$ 
&\cellcolor{LightCyan} $\bm{0.435}$  &\cellcolor{LightCyan} $\bm{0.423}$    \\ 
\midrule
\multirow{5}{1em}{\rot{WTH}}
& 96 
& $0.494$ & $0.505$ & 
\cellcolor{LightCyan}$\bm{0.483}$  &\cellcolor{LightCyan} $\bm{0.498}$ \\ 
& 192 
& $\bm{0.513}$ & $\bm{0.526}$ & 
\cellcolor{LightCyan}$0.528$ &\cellcolor{LightCyan} $0.536$ \\ 
& 336 
& $0.523$ & $0.539$ & 
\cellcolor{LightCyan}$0.523$ &\cellcolor{LightCyan} $0.539$ \\ 
& 720 
& $\bm{0.601}$ & $0.609$ 
&\cellcolor{LightCyan} $0.603$  & \cellcolor{LightCyan}$\bm{0.589}$  \\ 
\bottomrule
\end{tabular}
}
\label{table:time-series}
\vspace{-1.5em}
\end{table}

\newpage
\twocolumn

\def\arxivfont{\rm}
\bibliographystyle{plainnat}

\bibliography{refs}

\begin{thebibliography}{65}
\providecommand{\natexlab}[1]{#1}
\providecommand{\url}[1]{\texttt{#1}}
\expandafter\ifx\csname urlstyle\endcsname\relax
  \providecommand{\doi}[1]{doi: #1}\else
  \providecommand{\doi}{doi: \begingroup \urlstyle{rm}\Url}\fi

\bibitem[Alman and Song(2023)]{alman2023fast}
Josh Alman and Zhao Song.
\newblock Fast attention requires bounded entries.
\newblock In \emph{Thirty-seventh Conference on Neural Information Processing Systems (NeurIPS)}, 2023.
\newblock URL \url{https://openreview.net/forum?id=KOVWXcrFIK}.

\bibitem[Alman and Song(2024{\natexlab{a}})]{alman2024fine}
Josh Alman and Zhao Song.
\newblock The fine-grained complexity of gradient computation for training large language models.
\newblock \emph{arXiv preprint arXiv:2402.04497}, 2024{\natexlab{a}}.

\bibitem[Alman and Song(2024{\natexlab{b}})]{as23_tensor}
Josh Alman and Zhao Song.
\newblock How to capture higher-order correlations? generalizing matrix softmax attention to kronecker computation.
\newblock In \emph{The Twelfth International Conference on Learning Representations (ICLR)}, 2024{\natexlab{b}}.
\newblock URL \url{https://openreview.net/forum?id=v0zNCwwkaV}.

\bibitem[Auer et~al.(2024)Auer, Gauch, Klotz, and Hochreiter]{auer2023conformal}
Andreas Auer, Martin Gauch, Daniel Klotz, and Sepp Hochreiter.
\newblock Conformal prediction for time series with modern hopfield networks.
\newblock \emph{Advances in Neural Information Processing Systems (NeurIPS)}, 36, 2024.
\newblock URL \url{https://arxiv.org/abs/2303.12783}.

\bibitem[Bartunov et~al.(2019)Bartunov, Rae, Osindero, and Lillicrap]{bartunov2019meta}
Sergey Bartunov, Jack~W Rae, Simon Osindero, and Timothy~P Lillicrap.
\newblock Meta-learning deep energy-based memory models.
\newblock \emph{Eighth International Conference on Learning Representations (ICLR)}, 2019.
\newblock URL \url{https://arxiv.org/abs/1910.02720}.

\bibitem[Bhojanapalli et~al.(2020)Bhojanapalli, Yun, Rawat, Reddi, and Kumar]{bhojanapalli2020low}
Srinadh Bhojanapalli, Chulhee Yun, Ankit~Singh Rawat, Sashank Reddi, and Sanjiv Kumar.
\newblock Low-rank bottleneck in multi-head attention models.
\newblock In \emph{Thiry-Seventh International conference on machine learning(ICML)}, pages 864--873. PMLR, 2020.
\newblock URL \url{https://arxiv.org/abs/2002.07028}.

\bibitem[Bondarenko et~al.(2021)Bondarenko, Nagel, and Blankevoort]{bondarenko2021understanding}
Yelysei Bondarenko, Markus Nagel, and Tijmen Blankevoort.
\newblock Understanding and overcoming the challenges of efficient transformer quantization, 2021.
\newblock URL \url{https://arxiv.org/abs/2109.12948}.

\bibitem[Bondarenko et~al.(2023)Bondarenko, Nagel, and Blankevoort]{bondarenko2023quantizable}
Yelysei Bondarenko, Markus Nagel, and Tijmen Blankevoort.
\newblock Quantizable transformers: Removing outliers by helping attention heads do nothing.
\newblock \emph{Advances in Neural Information Processing Systems (NeurIPS)}, 36, 2023.
\newblock URL \url{https://arxiv.org/abs/2306.12929}.

\bibitem[Burns(2024)]{burns2024semantically}
Thomas~F Burns.
\newblock Semantically-correlated memories in a dense associative model.
\newblock In \emph{Forty-first International Conference on Machine Learning (ICML)}, 2024.
\newblock URL \url{https://arxiv.org/abs/2404.07123}.

\bibitem[Burns and Fukai(2023)]{burns2023simplicial}
Thomas~F Burns and Tomoki Fukai.
\newblock Simplicial hopfield networks.
\newblock In \emph{The Eleventh International Conference on Learning Representations (ICLR)}, 2023.
\newblock URL \url{https://openreview.net/forum?id=_QLsH8gatwx}.

\bibitem[Cabannes et~al.(2024{\natexlab{a}})Cabannes, Dohmatob, and Bietti]{cabannes2023scaling}
Vivien Cabannes, Elvis Dohmatob, and Alberto Bietti.
\newblock Scaling laws for associative memories.
\newblock In \emph{The Twelfth International Conference on Learning Representations (ICLR)}, 2024{\natexlab{a}}.
\newblock URL \url{https://openreview.net/forum?id=Tzh6xAJSll}.

\bibitem[Cabannes et~al.(2024{\natexlab{b}})Cabannes, Simsek, and Bietti]{cabannes2024learning}
Vivien Cabannes, Berfin Simsek, and Alberto Bietti.
\newblock Learning associative memories with gradient descent.
\newblock \emph{arXiv preprint arXiv:2402.18724}, 2024{\natexlab{b}}.
\newblock URL \url{https://arxiv.org/abs/2402.18724}.

\bibitem[Chen et~al.(2021{\natexlab{a}})Chen, Dao, Winsor, Song, Rudra, and R{\'e}]{chen2021scatterbrain}
Beidi Chen, Tri Dao, Eric Winsor, Zhao Song, Atri Rudra, and Christopher R{\'e}.
\newblock Scatterbrain: Unifying sparse and low-rank attention.
\newblock \emph{Advances in Neural Information Processing Systems (NeurIPS)}, 34:\penalty0 17413--17426, 2021{\natexlab{a}}.
\newblock URL \url{https://arxiv.org/abs/2110.15343}.

\bibitem[Chen et~al.(2020)Chen, Kornblith, Norouzi, and Hinton]{chen2020simple}
Ting Chen, Simon Kornblith, Mohammad Norouzi, and Geoffrey Hinton.
\newblock A simple framework for contrastive learning of visual representations.
\newblock In \emph{Thirty-seventh International conference on machine learning (ICML)}, pages 1597--1607. PMLR, 2020.
\newblock URL \url{https://arxiv.org/abs/2002.05709}.

\bibitem[Chen et~al.(2021{\natexlab{b}})Chen, Zeng, Ji, and Yang]{chen2021skyformer}
Yifan Chen, Qi~Zeng, Heng Ji, and Yun Yang.
\newblock Skyformer: Remodel self-attention with gaussian kernel and nystr$\backslash$" om method.
\newblock \emph{Advances in Neural Information Processing Systems (NeurIPS)}, 34:\penalty0 2122--2135, 2021{\natexlab{b}}.
\newblock URL \url{https://arxiv.org/abs/2111.00035}.

\bibitem[Correia et~al.(2019)Correia, Niculae, and Martins]{correia2019adaptively}
Gon{\c{c}}alo~M Correia, Vlad Niculae, and Andr{\'e}~FT Martins.
\newblock Adaptively sparse transformers.
\newblock \emph{arXiv preprint arXiv:1909.00015}, 2019.
\newblock URL \url{https://arxiv.org/abs/1909.00015}.

\bibitem[Demircigil et~al.(2017)Demircigil, Heusel, L{\"o}we, Upgang, and Vermet]{demircigil2017model}
Mete Demircigil, Judith Heusel, Matthias L{\"o}we, Sven Upgang, and Franck Vermet.
\newblock On a model of associative memory with huge storage capacity.
\newblock \emph{Journal of Statistical Physics}, 168:\penalty0 288--299, 2017.
\newblock URL \url{https://arxiv.org/abs/1702.01929}.

\bibitem[Dosovitskiy et~al.(2020)Dosovitskiy, Beyer, Kolesnikov, Weissenborn, Zhai, Unterthiner, Dehghani, Minderer, Heigold, Gelly, et~al.]{dosovitskiy2020image}
Alexey Dosovitskiy, Lucas Beyer, Alexander Kolesnikov, Dirk Weissenborn, Xiaohua Zhai, Thomas Unterthiner, Mostafa Dehghani, Matthias Minderer, Georg Heigold, Sylvain Gelly, et~al.
\newblock An image is worth 16x16 words: Transformers for image recognition at scale.
\newblock \emph{Eighth International Conference on Learning Representations (ICLR)}, 2020.
\newblock URL \url{https://arxiv.org/abs/2010.11929}.

\bibitem[F{\"u}rst et~al.(2022)F{\"u}rst, Rumetshofer, Lehner, Tran, Tang, Ramsauer, Kreil, Kopp, Klambauer, Bitto, et~al.]{furst2022cloob}
Andreas F{\"u}rst, Elisabeth Rumetshofer, Johannes Lehner, Viet~T Tran, Fei Tang, Hubert Ramsauer, David Kreil, Michael Kopp, G{\"u}nter Klambauer, Angela Bitto, et~al.
\newblock Cloob: Modern hopfield networks with infoloob outperform clip.
\newblock \emph{Advances in neural information processing systems (NeurIPS)}, 35:\penalty0 20450--20468, 2022.
\newblock URL \url{https://arxiv.org/abs/2110.11316}.

\bibitem[Gao et~al.(2023)Gao, Song, Wang, and Yin]{gao2023fast}
Yeqi Gao, Zhao Song, Weixin Wang, and Junze Yin.
\newblock A fast optimization view: Reformulating single layer attention in llm based on tensor and svm trick, and solving it in matrix multiplication time.
\newblock \emph{arXiv preprint arXiv:2309.07418}, 2023.

\bibitem[Gu et~al.(2024{\natexlab{a}})Gu, Liang, Liu, Shi, Song, and Yin]{gu2024conv}
Jiuxiang Gu, Yingyu Liang, Heshan Liu, Zhenmei Shi, Zhao Song, and Junze Yin.
\newblock Conv-basis: A new paradigm for efficient attention inference and gradient computation in transformers.
\newblock \emph{arXiv preprint arXiv:2405.05219}, 2024{\natexlab{a}}.
\newblock URL \url{https://arxiv.org/abs/2405.05219}.

\bibitem[Gu et~al.(2024{\natexlab{b}})Gu, Liang, Shi, Song, and Zhou]{gu2024tensor}
Jiuxiang Gu, Yingyu Liang, Zhenmei Shi, Zhao Song, and Yufa Zhou.
\newblock Tensor attention training: Provably efficient learning of higher-order transformers.
\newblock \emph{arXiv preprint arXiv:2405.16411}, 2024{\natexlab{b}}.
\newblock URL \url{https://arxiv.org/abs/2405.16411}.

\bibitem[Hofmann et~al.(2024)Hofmann, Schmid, Lehner, Klotz, and Hochreiter]{hofmann2024energy}
Claus Hofmann, Simon Schmid, Bernhard Lehner, Daniel Klotz, and Sepp Hochreiter.
\newblock Energy-based hopfield boosting for out-of-distribution detection.
\newblock \emph{arXiv preprint arXiv:2405.08766}, 2024.
\newblock URL \url{https://arxiv.org/abs/2405.08766}.

\bibitem[Hopfield(1982)]{hopfield1982neural}
John~J Hopfield.
\newblock Neural networks and physical systems with emergent collective computational abilities.
\newblock \emph{Proceedings of the national academy of sciences}, 79\penalty0 (8):\penalty0 2554--2558, 1982.

\bibitem[Hopfield(1984)]{hopfield1984neurons}
John~J Hopfield.
\newblock Neurons with graded response have collective computational properties like those of two-state neurons.
\newblock \emph{Proceedings of the national academy of sciences}, 81\penalty0 (10):\penalty0 3088--3092, 1984.

\bibitem[Hu et~al.(2023)Hu, Yang, Wu, Xu, Chen, and Liu]{hu2023SparseHopfield}
Jerry Yao-Chieh Hu, Donglin Yang, Dennis Wu, Chenwei Xu, Bo-Yu Chen, and Han Liu.
\newblock On sparse modern hopfield model.
\newblock In \emph{Thirty-seventh Conference on Neural Information Processing Systems (NeurIPS)}, 2023.
\newblock URL \url{https://arxiv.org/abs/2309.12673}.

\bibitem[Hu et~al.(2024{\natexlab{a}})Hu, Chang, Luo, Chen, Li, Wang, and Liu]{hu2024outlier}
Jerry Yao-Chieh Hu, Pei-Hsuan Chang, Robin Luo, Hong-Yu Chen, Weijian Li, Wei-Po Wang, and Han Liu.
\newblock Outlier-efficient hopfield layers for large transformer-based models.
\newblock In \emph{Forty-first International Conference on Machine Learning (ICML)}, 2024{\natexlab{a}}.
\newblock URL \url{https://arxiv.org/abs/2404.03828}.

\bibitem[Hu et~al.(2024{\natexlab{b}})Hu, Chen, Wu, Ruan, and Liu]{hu2024nonparametric}
Jerry Yao-Chieh Hu, Bo-Yu Chen, Dennis Wu, Feng Ruan, and Han Liu.
\newblock Nonparametric modern hopfield models.
\newblock \emph{arXiv preprint arXiv:2404.03900}, 2024{\natexlab{b}}.
\newblock URL \url{https://arxiv.org/abs/2404.03900}.

\bibitem[Hu et~al.(2024{\natexlab{c}})Hu, Lin, Song, and Liu]{hu2024computational}
Jerry Yao-Chieh Hu, Thomas Lin, Zhao Song, and Han Liu.
\newblock On computational limits of modern hopfield models: A fine-grained complexity analysis.
\newblock In \emph{Forty-first International Conference on Machine Learning (ICML)}, 2024{\natexlab{c}}.
\newblock URL \url{https://arxiv.org/abs/2402.04520}.

\bibitem[Hu et~al.(2024{\natexlab{d}})Hu, Wu, and Liu]{hwl24}
Jerry Yao-Chieh Hu, Dennis Wu, and Han Liu.
\newblock Provably optimal memory capacity for modern hopfield models: Transformer-compatible dense associative memories as spherical codes.
\newblock In \emph{Thirty-eighth Conference on Neural Information Processing Systems (NeurIPS)}, 2024{\natexlab{d}}.

\bibitem[Iatropoulos et~al.(2022)Iatropoulos, Brea, and Gerstner]{iatropoulos2022kernel}
Georgios Iatropoulos, Johanni Brea, and Wulfram Gerstner.
\newblock Kernel memory networks: A unifying framework for memory modeling.
\newblock \emph{Advances in Neural Information Processing Systems (NeurIPS)}, 35:\penalty0 35326--35338, 2022.
\newblock URL \url{https://arxiv.org/abs/2208.09416}.

\bibitem[Kanerva(1988)]{kanerva1988sparse}
Pentti Kanerva.
\newblock \emph{Sparse distributed memory}.
\newblock MIT press, 1988.

\bibitem[Kitaev et~al.(2020)Kitaev, Kaiser, and Levskaya]{kitaev2020reformer}
Nikita Kitaev, {\L}ukasz Kaiser, and Anselm Levskaya.
\newblock Reformer: The efficient transformer.
\newblock \emph{Ninth International Conference on Learning Repre- sentations (ICLR)}, 2020.
\newblock URL \url{https://arxiv.org/abs/2001.04451}.

\bibitem[Kozachkov et~al.(2023)Kozachkov, Kastanenka, and Krotov]{kozachkov2023building}
Leo Kozachkov, Ksenia~V Kastanenka, and Dmitry Krotov.
\newblock Building transformers from neurons and astrocytes.
\newblock \emph{Proceedings of the National Academy of Sciences}, 120\penalty0 (34):\penalty0 e2219150120, 2023.

\bibitem[Krizhevsky et~al.(2009)Krizhevsky, Hinton, et~al.]{krizhevsky2009learning}
Alex Krizhevsky, Geoffrey Hinton, et~al.
\newblock Learning multiple layers of features from tiny images.
\newblock 2009.

\bibitem[Krotov and Hopfield(2016)]{krotov2016dense}
Dmitry Krotov and John~J Hopfield.
\newblock Dense associative memory for pattern recognition.
\newblock \emph{Advances in neural information processing systems (NeurIPS)}, 29, 2016.
\newblock URL \url{https://arxiv.org/abs/1606.01164}.

\bibitem[Krotov and Hopfield(2021)]{krotov2020large}
Dmitry Krotov and John~J. Hopfield.
\newblock Large associative memory problem in neurobiology and machine learning.
\newblock In \emph{9th International Conference on Learning Representations (ICLR)}, 2021.
\newblock URL \url{https://arxiv.org/abs/2008.06996}.

\bibitem[Le and Yang(2015)]{le2015tiny}
Ya~Le and Xuan Yang.
\newblock Tiny imagenet visual recognition challenge.
\newblock \emph{CS 231N}, 7\penalty0 (7):\penalty0 3, 2015.

\bibitem[LeCun et~al.(1998)LeCun, Bottou, Bengio, and Haffner]{lecun1998gradient}
Yann LeCun, L{\'e}on Bottou, Yoshua Bengio, and Patrick Haffner.
\newblock Gradient-based learning applied to document recognition.
\newblock \emph{Proceedings of the IEEE}, 86\penalty0 (11):\penalty0 2278--2324, 1998.

\bibitem[Martins and Astudillo(2016)]{martins2016softmax}
Andre Martins and Ramon Astudillo.
\newblock From softmax to sparsemax: A sparse model of attention and multi-label classification.
\newblock In \emph{Thirty-third International conference on machine learning (ICML)}, pages 1614--1623. PMLR, 2016.
\newblock URL \url{https://arxiv.org/abs/1602.02068}.

\bibitem[Martins et~al.(2023)Martins, Niculae, and McNamee]{martins2023sparse}
Andre Martins, Vlad Niculae, and Daniel~C McNamee.
\newblock Sparse modern hopfield networks.
\newblock In \emph{Associative Memory {\&} Hopfield Networks in 2023}, 2023.
\newblock URL \url{https://openreview.net/forum?id=zwqlV7HoaT}.

\bibitem[Millidge et~al.(2022)Millidge, Salvatori, Song, Lukasiewicz, and Bogacz]{millidge2022universal}
Beren Millidge, Tommaso Salvatori, Yuhang Song, Thomas Lukasiewicz, and Rafal Bogacz.
\newblock Universal hopfield networks: A general framework for single-shot associative memory models.
\newblock In \emph{Thirty-ninth International Conference on Machine Learning (ICML)}, pages 15561--15583. PMLR, 2022.
\newblock URL \url{https://arxiv.org/abs/2202.04557}.

\bibitem[Negri et~al.(2023)Negri, Lauditi, Perugini, Lucibello, and Malatesta]{negri2023storage}
Matteo Negri, Clarissa Lauditi, Gabriele Perugini, Carlo Lucibello, and Enrico Malatesta.
\newblock Storage and learning phase transitions in the random-features hopfield model.
\newblock \emph{Physical Review Letters}, 131\penalty0 (25):\penalty0 257301, 2023.
\newblock URL \url{https://arxiv.org/abs/2303.16880}.

\bibitem[Pan et~al.(2024)Pan, Luo, Li, and Liu]{pan2024conv}
Zhenyu Pan, Haozheng Luo, Manling Li, and Han Liu.
\newblock Conv-coa: Improving open-domain question answering in large language models via conversational chain-of-action.
\newblock \emph{arXiv preprint arXiv:2405.17822}, 2024.
\newblock URL \url{https://arxiv.org/abs/2405.17822}.

\bibitem[Ramsauer et~al.(2020)Ramsauer, Sch{\"a}fl, Lehner, Seidl, Widrich, Adler, Gruber, Holzleitner, Pavlovi{\'c}, Sandve, et~al.]{ramsauer2020hopfield}
Hubert Ramsauer, Bernhard Sch{\"a}fl, Johannes Lehner, Philipp Seidl, Michael Widrich, Thomas Adler, Lukas Gruber, Markus Holzleitner, Milena Pavlovi{\'c}, Geir~Kjetil Sandve, et~al.
\newblock Hopfield networks is all you need.
\newblock \emph{arXiv preprint arXiv:2008.02217}, 2020.
\newblock URL \url{https://arxiv.org/abs/2008.02217}.

\bibitem[Reneau et~al.(2023)Reneau, Hu, Xu, Li, Gilani, and Liu]{reneau2023feature}
Alex Reneau, Jerry Yao-Chieh Hu, Chenwei Xu, Weijian Li, Ammar Gilani, and Han Liu.
\newblock Feature programming for multivariate time series prediction.
\newblock In \emph{Proceedings of the 40th International Conference on Machine Learning (ICML)}, volume 202 of \emph{Proceedings of Machine Learning Research}, pages 29009--29029. PMLR, 23--29 Jul 2023.
\newblock URL \url{https://arxiv.org/abs/2306.06252}.

\bibitem[Salvatori et~al.(2021)Salvatori, Song, Hong, Sha, Frieder, Xu, Bogacz, and Lukasiewicz]{salvatori2021associative}
Tommaso Salvatori, Yuhang Song, Yujian Hong, Lei Sha, Simon Frieder, Zhenghua Xu, Rafal Bogacz, and Thomas Lukasiewicz.
\newblock Associative memories via predictive coding.
\newblock \emph{Advances in Neural Information Processing Systems (NeurIPS)}, 34:\penalty0 3874--3886, 2021.
\newblock URL \url{https://arxiv.org/abs/2109.08063}.

\bibitem[Saunshi et~al.(2022)Saunshi, Ash, Goel, Misra, Zhang, Arora, Kakade, and Krishnamurthy]{saunshi2022understanding}
Nikunj Saunshi, Jordan Ash, Surbhi Goel, Dipendra Misra, Cyril Zhang, Sanjeev Arora, Sham Kakade, and Akshay Krishnamurthy.
\newblock Understanding contrastive learning requires incorporating inductive biases.
\newblock In \emph{Thirty-ninth International Conference on Machine Learning (ICML)}, pages 19250--19286. PMLR, 2022.
\newblock URL \url{https://arxiv.org/abs/2202.14037}.

\bibitem[Schaeffer et~al.(2024)Schaeffer, Zahedi, Khona, Pai, Truong, Du, Ostrow, Chandra, Carranza, Fiete, et~al.]{schaeffer2024bridging}
Rylan Schaeffer, Nika Zahedi, Mikail Khona, Dhruv Pai, Sang Truong, Yilun Du, Mitchell Ostrow, Sarthak Chandra, Andres Carranza, Ila~Rani Fiete, et~al.
\newblock Bridging associative memory and probabilistic modeling.
\newblock \emph{arXiv preprint arXiv:2402.10202}, 2024.
\newblock URL \url{https://arxiv.org/abs/2402.10202}.

\bibitem[Scholkopf and Smola(2018)]{scholkopf2018learning}
Bernhard Scholkopf and Alexander~J Smola.
\newblock \emph{Learning with kernels: support vector machines, regularization, optimization, and beyond}.
\newblock MIT press, 2018.

\bibitem[Song et~al.(2021)Song, Jung, Kim, and Moon]{song2021implicit}
Kyungwoo Song, Yohan Jung, Dongjun Kim, and Il-Chul Moon.
\newblock Implicit kernel attention.
\newblock In \emph{Proceedings of the AAAI Conference on Artificial Intelligence}, volume~35, pages 9713--9721, 2021.
\newblock URL \url{https://arxiv.org/abs/2006.06147}.

\bibitem[Sriperumbudur and Lanckriet(2009)]{sriperumbudur2009convergence}
Bharath~K Sriperumbudur and Gert~RG Lanckriet.
\newblock On the convergence of the concave-convex procedure.
\newblock In \emph{Advances in neural information processing systems (NeurIPS)}, volume~9, pages 1759--1767, 2009.
\newblock URL \url{https://papers.nips.cc/paper_files/paper/2009/file/8b5040a8a5baf3e0e67386c2e3a9b903-Paper.pdf}.

\bibitem[Sun et~al.(2024)Sun, Chen, Kolter, and Liu]{sun2024massive}
Mingjie Sun, Xinlei Chen, J~Zico Kolter, and Zhuang Liu.
\newblock Massive activations in large language models.
\newblock \emph{arXiv preprint arXiv:2402.17762}, 2024.
\newblock URL \url{https://arxiv.org/abs/2402.17762}.

\bibitem[Tyulmankov et~al.(2021)Tyulmankov, Fang, Vadaparty, and Yang]{tyulmankov2021biological}
Danil Tyulmankov, Ching Fang, Annapurna Vadaparty, and Guangyu~Robert Yang.
\newblock Biological learning in key-value memory networks.
\newblock \emph{Advances in Neural Information Processing Systems (NeurIPS)}, 34:\penalty0 22247--22258, 2021.
\newblock URL \url{https://arxiv.org/abs/2110.13976}.

\bibitem[Vaswani et~al.(2017)Vaswani, Shazeer, Parmar, Uszkoreit, Jones, Gomez, Kaiser, and Polosukhin]{vaswani2017attention}
Ashish Vaswani, Noam Shazeer, Niki Parmar, Jakob Uszkoreit, Llion Jones, Aidan~N Gomez, {\L}ukasz Kaiser, and Illia Polosukhin.
\newblock Attention is all you need.
\newblock \emph{Advances in neural information processing systems (NeurIPS)}, 30, 2017.
\newblock URL \url{https://arxiv.org/abs/1706.03762}.

\bibitem[Wang and Isola(2020)]{wang2020understanding}
Tongzhou Wang and Phillip Isola.
\newblock Understanding contrastive representation learning through alignment and uniformity on the hypersphere.
\newblock In \emph{39th International Conference on Machine Learning (ICML)}, pages 9929--9939. PMLR, 2020.

\bibitem[Widrich et~al.(2020)Widrich, Sch{\"a}fl, Pavlovi{\'c}, Ramsauer, Gruber, Holzleitner, Brandstetter, Sandve, Greiff, Hochreiter, et~al.]{widrich2020modern}
Michael Widrich, Bernhard Sch{\"a}fl, Milena Pavlovi{\'c}, Hubert Ramsauer, Lukas Gruber, Markus Holzleitner, Johannes Brandstetter, Geir~Kjetil Sandve, Victor Greiff, Sepp Hochreiter, et~al.
\newblock Modern hopfield networks and attention for immune repertoire classification.
\newblock \emph{Advances in Neural Information Processing Systems (NeurIPS)}, 33:\penalty0 18832--18845, 2020.
\newblock URL \url{https://arxiv.org/abs/2007.13505}.

\bibitem[Willshaw et~al.(1969)Willshaw, Buneman, and Longuet-Higgins]{willshaw1969non}
David~J Willshaw, O~Peter Buneman, and Hugh~Christopher Longuet-Higgins.
\newblock Non-holographic associative memory.
\newblock \emph{Nature}, 222\penalty0 (5197):\penalty0 960--962, 1969.

\bibitem[Wu et~al.(2024)Wu, Hu, Li, Chen, and Liu]{wu2023stanhop}
Dennis Wu, Jerry Yao-Chieh Hu, Weijian Li, Bo-Yu Chen, and Han Liu.
\newblock {ST}anhop: Sparse tandem hopfield model for memory-enhanced time series prediction.
\newblock In \emph{The Twelfth International Conference on Learning Representations (ICLR)}, 2024.
\newblock URL \url{https://arxiv.org/abs/2312.17346}.

\bibitem[Xu et~al.(2024)Xu, Huang, Hu, Li, Gilani, Goan, and Liu]{xu2024bishop}
Chenwei Xu, Yu-Chao Huang, Jerry Yao-Chieh Hu, Weijian Li, Ammar Gilani, Hsi-Sheng Goan, and Han Liu.
\newblock Bishop: Bi-directional cellular learning for tabular data with generalized sparse modern hopfield model.
\newblock In \emph{Forty-first International Conference on Machine Learning (ICML)}, 2024.
\newblock URL \url{https://arxiv.org/abs/2404.03830}.

\bibitem[Yampolskaya and Mehta(2023)]{yampolskaya2023controlling}
Maria Yampolskaya and Pankaj Mehta.
\newblock Controlling the bifurcations of attractors in modern hopfield networks.
\newblock In \emph{Associative Memory $\{$$\backslash$\&$\}$ Hopfield Networks in 2023}, 2023.

\bibitem[Yoo and Wood(2022)]{yoo2022bayespcn}
Jinsoo Yoo and Frank Wood.
\newblock Bayespcn: A continually learnable predictive coding associative memory.
\newblock \emph{Advances in Neural Information Processing Systems (NeurIPS)}, 35:\penalty0 29903--29914, 2022.
\newblock URL \url{https://arxiv.org/abs/2205.09930}.

\bibitem[Yuille and Rangarajan(2001)]{yuille2001concave}
Alan~L Yuille and Anand Rangarajan.
\newblock The concave-convex procedure (cccp).
\newblock \emph{Advances in neural information processing systems (NeurIPS)}, 14, 2001.

\bibitem[Zangwill(1969)]{zangwill1969nonlinear}
Willard~I Zangwill.
\newblock \emph{Nonlinear programming: a unified approach}, volume~52.
\newblock Prentice-Hall Englewood Cliffs, NJ, 1969.

\bibitem[Zhou et~al.(2021)Zhou, Zhang, Peng, Zhang, Li, Xiong, and Zhang]{zhou2021informer}
Haoyi Zhou, Shanghang Zhang, Jieqi Peng, Shuai Zhang, Jianxin Li, Hui Xiong, and Wancai Zhang.
\newblock Informer: Beyond efficient transformer for long sequence time-series forecasting.
\newblock In \emph{Proceedings of the AAAI conference on artificial intelligence}, volume~35, pages 11106--11115, 2021.
\newblock URL \url{https://arxiv.org/abs/2012.07436}.

\end{thebibliography}

\end{document}